\documentclass{article}

\PassOptionsToPackage{numbers,compress}{natbib}
\usepackage[main, final]{neurips_2026}

\usepackage[utf8]{inputenc}
\usepackage[T1]{fontenc}
\usepackage[colorlinks,citecolor=blue,urlcolor=magenta]{hyperref}
\usepackage{url}
\usepackage{booktabs}
\usepackage{amsfonts}
\usepackage{nicefrac}
\usepackage{microtype}
\usepackage[table]{xcolor}
\usepackage{amsmath}
\usepackage{graphicx}
\usepackage{multirow}
\usepackage{enumitem}
\usepackage{algorithm}
\usepackage{algpseudocode}
\usepackage{tikz}
\usepackage{xspace}
\usepackage{caption}
\usepackage{subcaption}
\usepackage{float}
\usepackage{threeparttable}
\usepackage{wrapfig}
\usepackage{tabularx}
\usepackage{array}
\usepackage{ragged2e}

\definecolor{modelcolor}{HTML}{F0F0FF}

\newcommand{\xhdr}[1]{\vspace{0.5mm}\noindent{{\bf #1.}}}
\newcommand{\name}{\textsc{AutoScientists}\xspace}
\newif\ifshowrevision \showrevisiontrue
\definecolor{revisioncolor}{HTML}{000000}
\DeclareRobustCommand{\revision}[1]{\ifshowrevision{\color{revisioncolor}#1}\else#1\fi}
\newcolumntype{Y}{>{\RaggedRight\arraybackslash}X}
\newcolumntype{L}[1]{>{\RaggedRight\arraybackslash}p{#1}}

\title{
\name: Self-Organizing Agent Teams \\ for Long-Running Scientific Experimentation
}

\author{%
  Shanghua Gao\footnotemark[1] \\
  Harvard University \\
  \texttt{shanghua\_gao@hms.harvard.edu}
   \\
  \And
  Ada Fang\thanks{Equal contribution.} \\
  Harvard University \\
  \texttt{ada\_fang@g.harvard.edu}\\
  \AND
  Marinka Zitnik \\
  Harvard University \\
  \texttt{marinka@hms.harvard.edu} \\
}

\begin{document}

\maketitle

\vspace{-2em}
\begin{center}
\name website: \url{https://autoscientists.openscientist.ai} \\
\name code: \url{https://github.com/mims-harvard/AutoScientists}
\end{center}

\begin{abstract}
Scientific research proceeds through iterative cycles of hypothesis generation, experiment design, execution, and revision.
AI agents can automate parts of this process, but existing approaches typically follow a single research trajectory or coordinate through a central planner with fixed objectives. As a result, they struggle to sustain parallel exploration, adapt as experimental evidence changes, or preserve knowledge of failed directions over long-running experiments.
We introduce \name, a decentralized team of AI agents for long-running computational scientific experimentation.
Agents interpret a shared experimental state, self-organize into teams around promising hypotheses, critique proposals before using experimental compute, and share successes and failures to reduce redundant exploration.
Under matched experimental budgets, \name improves over prior AI agents across biomedical machine learning, language model training optimization, and protein fitness prediction.
On BioML-Bench, spanning biomedical imaging, protein engineering, single-cell omics, and drug discovery, \name achieves a mean leaderboard percentile of 74.4\% across 24 tasks, improving over the strongest AI agent by +8.33\% in the absolute leaderboard percentile points.
On GPT training optimization, \name reaches a target validation bits-per-byte 1.9$\times$ faster than Autoresearch and continues discovering improvements from a  starting champion where the single-agent approach finds none (7 vs.\ 0 accepted improvements).
On ProteinGym fitness prediction, \name discovers a method for ACE2-Spike binding that improves over the current state-of-the-art model by +12.5\% in Spearman correlation. Applied without modification across all 217 ProteinGym assays, the same method improves over the prior state of the art by +6.5\% (Spearman correlation).

\end{abstract}

\section{Introduction}

AI agents for science are beginning to move beyond answering questions and running predefined workflows toward proposing and executing research steps~\cite{gao2024empowering}, from protein engineering in biology to language model optimization in machine learning~\cite{miller2025bioml, karpathy2026autoresearch}. Agents can generate hypotheses, synthesize literature, design computational experiments, write and execute code, and refine models from experimental feedback~\cite{mitchener2025kosmos, jin2025stella, gao2025txagent, gottweis2025towards, penades2025ai, sui2026medea, huang2025biomni}. However, most current approaches remain limited to short horizon optimization or fixed pipelines. They typically follow a single reasoning thread or use a search space decomposition set at the start of the run. \revision{These designs break down} in long-running scientific experimentation, where research directions are not known in advance and change over time.

Existing AI agents can run experiments, but long-running science requires maintaining competing hypotheses, updating them as evidence changes, and using failures to redirect the search. Single-agent systems such as AIDE~\cite{jiang2025aide} and Autoresearch~\cite{karpathy2026autoresearch} iteratively refine proposals but follow a single search trajectory, limiting their ability to explore competing hypotheses in parallel. Multi-agent systems~\cite{qu2026coral, feng2026internagent, swanson2025virtual} distribute work across agents, but still coordinate through a central structure, in which a planner decomposes the problem, a search algorithm ranks proposals, or agents converge through discussion or voting~\cite{du2024improving, chen2024reconcile}. These approaches assume that the search space can be partitioned into stable directions at the start of the run. {\em In long-running experimentation, however, productive directions shift as evidence accumulates. Some hypotheses stop yielding improvements, failed directions must be tracked to avoid repeated exploration, and new hypotheses often emerge only after earlier experiments are analyzed.}

\begin{figure}[t]
\centering
\includegraphics[width=\linewidth]{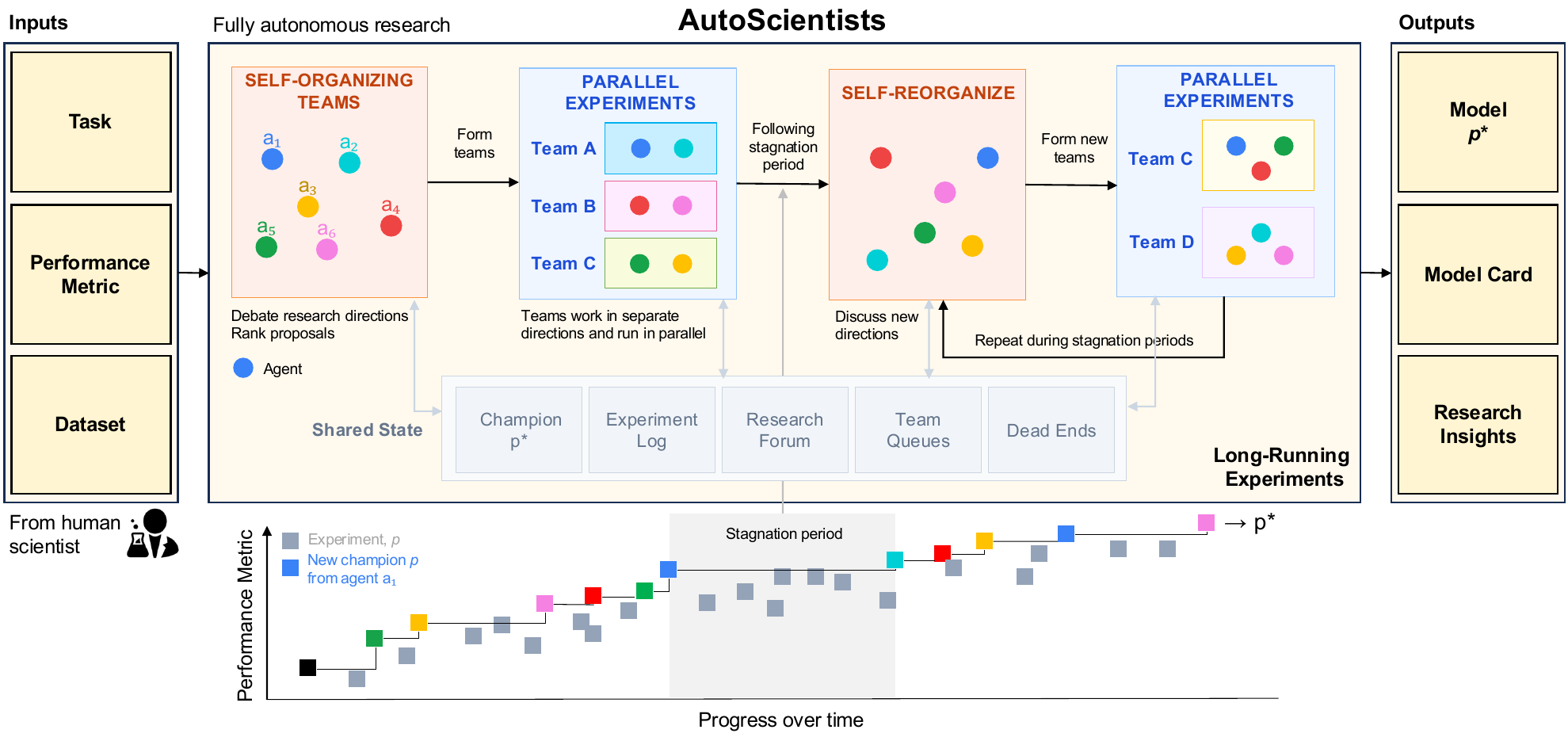}
\caption{\textbf{Self-organizing agent teams for long-running experimentation.} Overview of \name. Agents identify promising research directions, organize into teams, and execute experiments in parallel.}
\vspace{-7mm}
\label{fig:overview}
\end{figure}

\xhdr{Present Work}
We introduce \name, a self-organizing agent team for long-running scientific experimentation that coordinates without a central orchestrator agent (Figure~\ref{fig:overview}). Rather than receiving assignments from a planner, agents act on a shared state that records proposals, experiments, results, failures, and the current champion. Teams form dynamically through agent interaction rather than \revision{a} user specified decomposition. Agents post experiment proposals to a shared forum, where peers critique them before execution, filtering weak ideas before compute is committed. As results accumulate, agents reorganize around productive directions, retire exhausted directions, and share successes and failures across teams to reduce redundant exploration. We apply \name to research tasks spanning imaging, drug discovery, single-cell omics, protein engineering, protein fitness prediction, and language model training optimization.

Across benchmarks, \name improves over existing AI agents. On BioML-Bench~\cite{miller2025bioml}, \name achieves the highest average leaderboard percentile among the evaluated agents, reaching $74.40\%$ across 24 biomedical ML tasks compared with $66.07\%$ for Autoresearch under the same task interface, model backend, and hardware budget. The performance improvements are largest in drug discovery, where \name improves from $46.16\%$ to $64.52\%$. On GPT nanochat training optimization~\cite{karpathy2026autoresearch}, \name reaches the same intermediate validation loss in 34 experiments that Autoresearch reaches in 65 experiments, and when continuing from an \name champion reaches a validation bits-per-byte (bpb) of 0.9730 while Autoresearch finds no accepted improvements over 100 experiments. On ProteinGym supervised substitution fitness prediction~\cite{proteingym}, \name starts from Kermut and discovers a Kermut extension that improves ACE2--Spike binding Spearman's $\rho$ from $0.747$ to $0.840$. Furthermore, the frozen recipe transfers across the full 217-assay ProteinGym supervised substitution benchmark, improving the official average Spearman's $\rho$ from $0.657$ to $0.700$. Below we summarize our contributions.

\begin{itemize}[leftmargin=20pt, itemsep=2pt, parsep=0pt]
    \item \textbf{A self-organizing agent team for long-horizon scientific experimentation}. Unlike prior systems that rely on central coordinators, consensus-based discussion, or fixed decompositions of the search space, \name allows agents to independently interpret a shared experimental state and decide which hypotheses to pursue. Agents post proposals to a shared forum where peers critique and filter them before experiments run, allowing teams and experimental directions to emerge through interaction rather than external assignment.

    \item \textbf{State-of-the-art performance across scientific domains, with sustained improvement during long-running experimental search.} \name improves over prior agents on biomedical ML, protein fitness prediction, and language model training optimization, and continues identifying productive modifications after single-agent baselines stop improving.
\end{itemize}

\section{Related Work}

\xhdr{AI Agents for Scientific Research}
AI agents are being developed to automate scientific workflows, including literature review, hypothesis generation, tool use, code execution, experimental design, benchmarking, and manuscript drafting \cite{gridach2025agenticaiscientificdiscovery, liu2025visionautoresearchllm, schmidgall2025agent, bragg2026astabenchrigorousbenchmarkingai, yan2025lmrbenchevaluatingllmagents}. Biomedical agents combine multi-step reasoning with biomedical tools, literature grounding, omics analysis, code execution, and evidence reconciliation \cite{huang2025biomni, gao2025txagent, sui2026medea, bu2026empowering, ghareeb2025robin, liu2025genomas}. Other systems push toward longer horizon discovery through repeated cycles of literature search, hypothesis generation, debate, refinement, tool integration, optimization, equation discovery, self-directed exploration, and skill accumulation \cite{mitchener2025kosmos, gottweis2025towards, pu2025piflow, ding2025scitoolagent, han2025chembomas, xia2025sr, wu2026selfaiselfdirectedframeworklonghorizon, lyu2026evoscientistmultiagentevolvingai, qu2026coral, huang2025cascade}. Several scientific agent systems rely on role specialized architectures, such as PI--scientist--critic organizations or Manager--Developer--Critic--Tool Creation pipelines \cite{swanson2025virtual, jin2025stella}. In AI research, related systems have also generated research papers or evolved algorithms through iterative code modification and experimentation \cite{lu2026towards, karpathy2026autoresearch, novikov2025alphaevolve}. Our system differs in that agents collectively determine research directions through discussion and coordinate through shared forums rather than fixed pipelines or a central orchestrator that directs others. Unlike debate frameworks that use discussion to converge on a shared hypothesis~\cite{du2024improving, chen2024reconcile}, \name uses discussion to filter out weak proposals before any experiment runs, while allowing agents to continue pursuing different research directions in parallel.

\xhdr{Coordination of Multi-Agent Systems}
Beyond scientific applications, multi-agent performance depends on collaboration structure and agent composition \cite{zhuge2024language, huang2025on, berdoz2026can, tran2025multiagentcollaborationmechanismssurvey, yang2026understanding, kim2025towards}. Interaction is not automatically beneficial. For example, multi-agent systems have underperformed their best individual member on \revision{some} tasks \cite{pappu2026multi, kim2025towards}, and recent benchmarks analyze how collaboration and competition affect collective performance \cite{zhu2025multiagentbenchevaluatingcollaborationcompetition}. These findings motivate our ablation studies and comparison to single-agent baselines like Autoresearch. \revision{Performance depends on how well the coordination structure matches the task rather than on agent count alone~\cite{kim2025towards}, and multi-agent failures often stem from system design, inter-agent misalignment, and weak task verification~\cite{cemri2025multi}. In \name, the shared state, proposal critique, dead-end registry, noise-aware validation, and team reorganization target these limitations.} Human scientific teams provide a complementary perspective as they benefit from diversity and flatter structures, but excessive diversity can introduce coordination costs \cite{cummings2005collaborative, wuchty2007increasing, xu2022flat, hall2018science}. Recent work also emphasizes context management, memory, and reusable skills for sustained collaboration \cite{qu2026coral, zhu2026toward, huang2025cascade}. Our system draws on these findings by organizing agents as teams focused on complementary research directions and \revision{using} shared forums to support conference style knowledge sharing and collective intelligence \cite{riedl2025emergent}. \revision{Shared state coordination dates back to blackboard architectures such as Hearsay-II~\cite{erman1980hearsay, nii1986blackboard}. Recent LLM blackboard systems decentralize which agent contributes next to a shared reasoning thread~\cite{han2025blackboard}, which agents answer a centrally posted request~\cite{salemi2025blackboard}, or which agent claims a predefined work unit~\cite{pugachev2025codecrdt}. \name instead decentralizes how research is structured, with agents that create, continue, and reorganize persistent experimental threads and their teams in response to experimental evidence.}

\section{\name: Long-Running Self-Organizing Agent Teams}\label{sec:multi-agent-defn}

\revision{We formalize long-running scientific experimentation as an iterative search process and then introduce \name.} We first define the optimization setting and then describe how agents organize into teams, propose and execute experiments, exchange experimental evidence through a shared state, and reorganize as search trajectories evolve over time.

\subsection{Problem Formulation}

We are given a task description, optionally accompanied by an initial program (e.g., a training script) $p_0$, together with a dataset $\mathcal{D}$ and an evaluation metric $\ell$.

The dataset $\mathcal{D}$ consists of a training set $\mathcal{D}_{\text{train}}$ and an evaluation protocol. The evaluation protocol may take one of \revision{two forms, either}
a validation set $\mathcal{D}_{\text{val}}$ \revision{or} a cross-validation (CV) scheme over $\mathcal{D}_{\text{train}}$.
We denote by $\ell_{\text{eval}}(p; \mathcal{D})$ the evaluation metric computed under this protocol.

A system of $n$ long-running LLM agents $\mathcal{A} = \{a_1, \ldots, a_n\}$ iteratively proposes and generates new programs. Long-running agents $a_i$ persist over the course of the search process, maintaining internal state and updating their behavior based on accumulated experience. This contrasts with one-shot agents that generate a solution in a single forward pass. Each proposed program $p$ is trained on $\mathcal{D}_{\text{train}}$ and evaluated using $\ell_{\text{eval}}$. The goal is to identify a program
\[
p^* = \arg\max_{p \in \mathcal{P}} \ \ell_{\text{eval}}(p; \mathcal{D}),
\]
where $\mathcal{P}$ denotes the space of programs explored by the agents during the search process, optionally initialized from $p_0$. We assume without loss of generality that $\ell$ is oriented so that higher values correspond to better performance (e.g., by negating metrics that are typically minimized such as loss).

At the end of the search process, performance is reported using $\ell_{\text{test}}$ if a held-out test set is available, \revision{and using $\ell_{\text{eval}}$ (e.g., validation or CV performance) otherwise.}

\subsection{\name Approach}

\xhdr{Overview}
\name deploys $n$ long-running AI agents that maintain state across the run, adapt their search strategy, \revision{reorganize} into teams, and update their search behavior from accumulated evidence (Figure~\ref{fig:overview}). The system alternates between two phases. In the \emph{discussion} phase, agents analyze the task, propose experimental directions, and organize into teams. In the \emph{execution} phase, teams run parallel experiments and write results back to the shared state $\mathcal{S}$. When performance on $\ell_\text{eval}$ stagnates, agents reopen discussion and may reorganize teams around different directions. This cycle continues for the duration of the run and is coordinated through $\mathcal{S}$ rather than a central planner agent.

\xhdr{Discussion and Self-Organization}
Agents identify and revise research directions through discussion phases, without a predefined partition of the search space. \name initializes with no teams and no predefined directions. At the start of each discussion phase, all agents read the task specification, the current champion $p^*$, and prior posts on the shared forum $\mathcal{F}$.
Discussion proceeds over multiple rounds. Early rounds focus on proposing and evaluating candidate directions, where agents independently analyze $p^*$, propose modifications, critique competing proposals, and identify gaps in the search space. Later rounds organize agents into $K$ teams $\{\mathcal{T}_1, \ldots, \mathcal{T}_K\}$, where each team is assigned one research direction. The final agent in the discussion round consolidates the proposals into a roster
$
R = \{(\mathcal{T}_k, \text{direction}_k, \text{members}_k)\}_{k=1}^K
$
and writes it to $\mathcal{S}$. Subsequent agents adopt the roster on their next heartbeat.

The roster changes as evidence accumulates. When a team stops producing improvements, agents trigger a new discussion phase and review results across all teams. Through the shared research forum, agents can propose to create, merge, split, or rebalance teams, with changes requiring endorsement from affected teams before taking effect. This allows \name to redirect effort during the run, so exhausted directions can be retired, and newly emerging hypotheses can form new teams.

\xhdr{Long-Running Parallel Experiments}
Each team $\mathcal{T}_k$ operates a continuous \revision{propose and execute} loop.
Every agent runs a heartbeat cycle in which it reads the shared state $\mathcal{S}$, acts according to its role, writes results back to $\mathcal{S}$, and repeats.
Agents persist across cycles with their own identity and memory files, accumulating knowledge over the duration of the run.
Two specialized roles collaborate in each team.

\xhdr{(1) Analyst Agents}
Analysts maintain the team's search knowledge and propose experiments.
\revision{In each} heartbeat cycle, an analyst reads the experiment log $\mathcal{L}$, audits which research directions have never been tested, and posts proposals to the team queue $Q_k$.
Proposals are ranked by observed effect sizes from $\mathcal{L}$, where underexplored research directions are prioritized, and research directions with consistently small effects are deprioritized (details in Appendix~\ref{app:analyst_details}).
After a champion update, the analyst \revision{identifies the features responsible for the improvement and proposes variants that share them.}

\xhdr{(2) Experiment Agents}
Experiment agents claim experiments from the team queue $Q_k$, apply the code change to $p^*$, train, and record the outcome \revision{in} $\mathcal{L}$ and $\mathcal{F}$.
Since the evaluation metric $\ell$ may be stochastic (e.g., \revision{variation across training seeds}), improvements within the empirically measured noise band are confirmed on a second seed before promotion to $p^*$ (details in Appendix~\ref{app:noise_gate}).
All results, including failures, are visible to every agent across all teams.

Teams execute in parallel for the full duration.
As experiments accumulate, teams track failed experiments in a dead-end registry $\mathcal{D}_k$ to avoid repeating unproductive directions, and rank \revision{their queues} $Q_k$ by observed effect sizes from $\mathcal{L}$ so that underexplored directions are tried first.
When a team's recent experiments consistently fail to improve $p^*$ (e.g., no improvement in the last 10 experiments), agents return to discussion and may reorganize into new teams around more productive directions.

\xhdr{Shared State}
The system maintains a shared state accessible to all agents, consisting of a champion $p^*$ tracking the current best model with full hyperparameters and reproduction instructions; an experiment log $\mathcal{L}$ of every completed experiment with outcome, metric delta, and training diagnostics; a shared forum $\mathcal{F}$ of structured posts where proposals are debated, results announced, and mechanistic analyses shared; and team local state (per team experiment queues $Q_k$, dead-end registries $\mathcal{D}_k$, and hypothesis documents) that is readable cross-team. Details are in Appendix~\ref{app:implementation}.

\begin{wrapfigure}{r}{0.5\linewidth}
\vspace{-4mm}
  \centering
  \includegraphics[width=\linewidth]{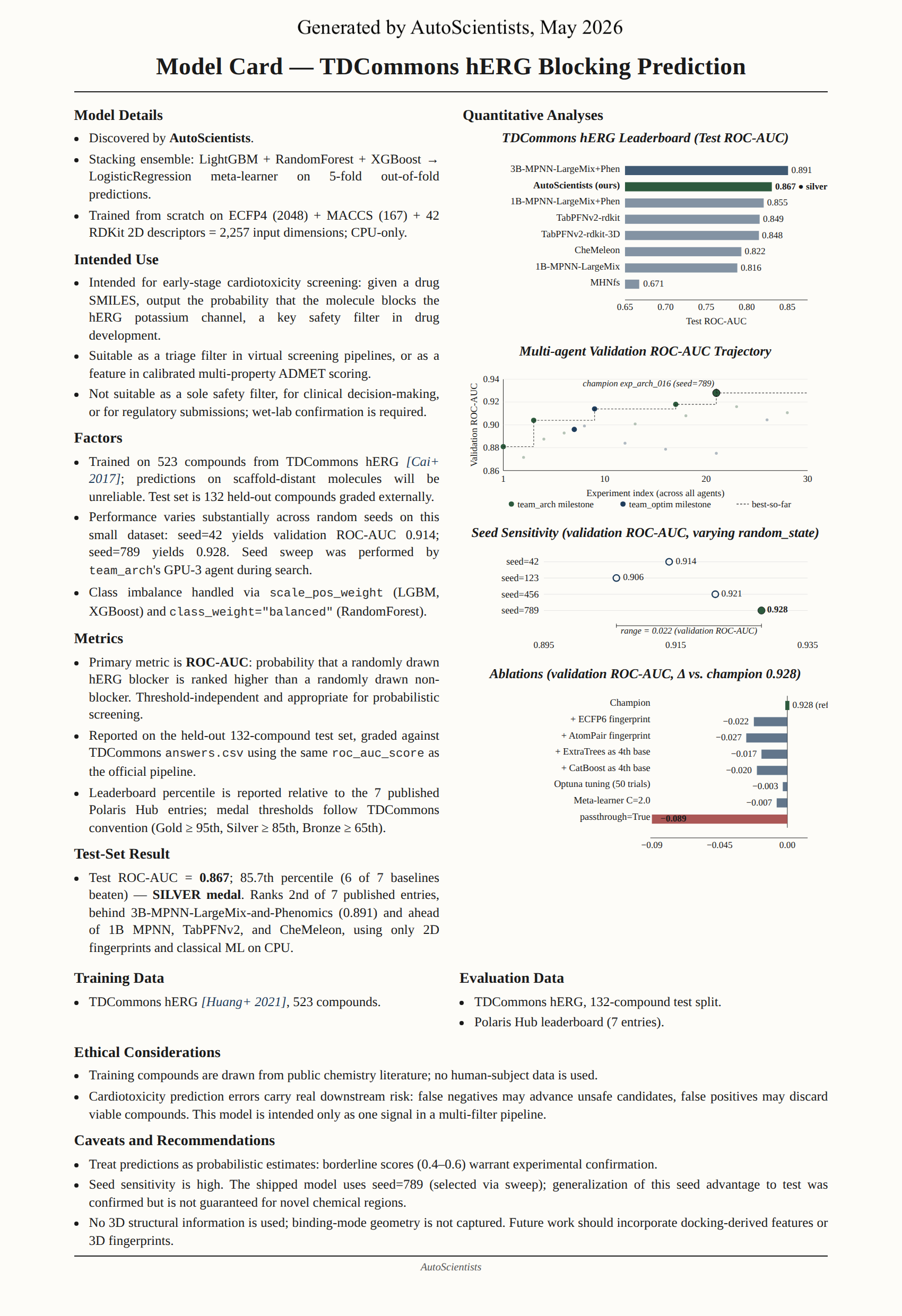}
 \caption{\textbf{Model card produced by \name.} TDC hERG Blocking Prediction model discovered by \name.}\label{fig:model-card}
\end{wrapfigure}
\xhdr{Output}
\name outputs the final champion model $p^*$ together with a model card and a research findings report derived from the agents' experimental process. \name produces the main technical components of a model card~\cite{mitchell2019model}. Model architecture, hyperparameters, and training procedures are recorded in the reproducible champion training script. Training and evaluation datasets are inherited from the shared task specification, and quantitative performance metrics are stored in the champion record. Figure~\ref{fig:model-card} shows a model card for a hERG prediction model discovered by \name on BioML-Bench.

In addition to the final model, \name records the experimental search process that produced it. Dead-end registries store failed experimental directions together with the tested research direction, performance change, and rejection reason. Analyst agents document the mechanisms underlying successful modifications and propose related follow-up directions. Combined with the full experiment log, these artifacts provide a record of how hypotheses evolved during the run, which directions were abandoned, and how the final model emerged from accumulated experimental evidence.
Appendix~\ref{app:autoscientists_output} presents the complete set of artifacts produced by \name on the GPT nanochat task.

\section{Experiments}

\subsection{Implementation Details}
All agents in \name use the same base model, \revision{the} Claude Code coding agent~\cite{anthropic_claude_code} with the base LLM Claude Sonnet 4.6~\cite{anthropic_claude_sonnet_46}. We use the same model backend for \name and the Autoresearch baseline. \revision{To test whether results depend on this backend, we also run \name with the OpenAI Codex coding agent on two tasks (Appendix~\ref{app:repeated_runs}).} Each agent is repeatedly invoked by a deterministic monitor process in a heartbeat loop. \name was given access to H100 GPUs for running experiments. \revision{Details for reproducing the experimental results are in} \revision{the \hyperref[app:reproduce]{Reproducibility Statement}}. Unless specified otherwise, the \name team is composed of 3 analyst agents and 6 experiment agents.

\subsection{End-to-End Biomedical Machine Learning with \name}
\label{sec:biomlbench}

\xhdr{Setup}
We evaluate \name on BioML-Bench, a benchmark of 24 end-to-end biomedical machine learning tasks spanning biomedical imaging (4), drug discovery (9), protein engineering (6), and single-cell omics (5)~\cite{miller2025bioml}. Each task provides a natural language task description, training data, test inputs, and an example submission format. For each of the four task types, an LLM generated list of general model paradigms is included in the \name agent's discussion prompts to encourage diverse research directions. \name develops models using the task description, training data, and development time validation feedback. Hidden test labels and private grader files are kept outside the agent workspace and are accessed only by the external evaluator\revision{, whereas protein engineering tasks are scored with the benchmark's cross-validation protocol (Appendix~\ref{app:biomlbench})}.
Following BioML-Bench, we report leaderboard percentile relative to public human submissions, whether the submission exceeds the public leaderboard median, whether it receives any medal, and completion rate. We compare against \revision{the BioML-Bench baselines} Reference, MLAgentBench~\cite{huang2023mlagentbench}, AIDE~\cite{jiang2025aide}, STELLA~\cite{jin2025stella}, and Biomni~\cite{huang2025biomni}. We additionally adapt Autoresearch~\cite{karpathy2026autoresearch}, implemented with the same coding agent backend, to the BioML-Bench \revision{tasks}. \revision{All agents receive the same experimental compute as \name.
}
Experimental compute settings and task-specific setup details are provided in Appendix~\ref{app:biomlbench}.

\begin{figure}[h]
    \centering
    \includegraphics[width=0.85\linewidth]{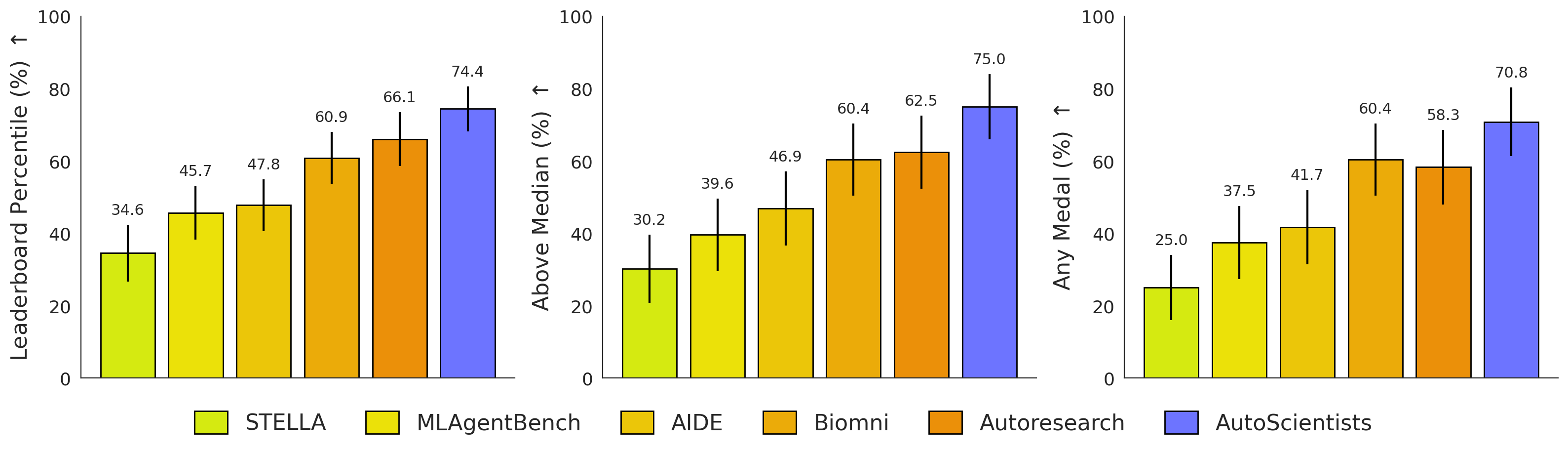}
    \caption{\textbf{\name improves performance across BioML-Bench tasks.} Performance on 24 biomedical tasks measured by leaderboard percentile (\textbf{left}), proportion above the public leaderboard median (\textbf{middle}), and proportion awarded a medal (\textbf{right}). Error bars show standard error of the mean. \revision{All agents use the same experimental compute allocation (Appendix~\ref{app:biomlbench}).} Additional results are reported in Table~\ref{tab:biomlbench-full}.}
    \label{fig:biomlbench}
    \vspace{-2mm}
\end{figure}

\begin{figure}[h]
    \centering
    \includegraphics[width=\linewidth]{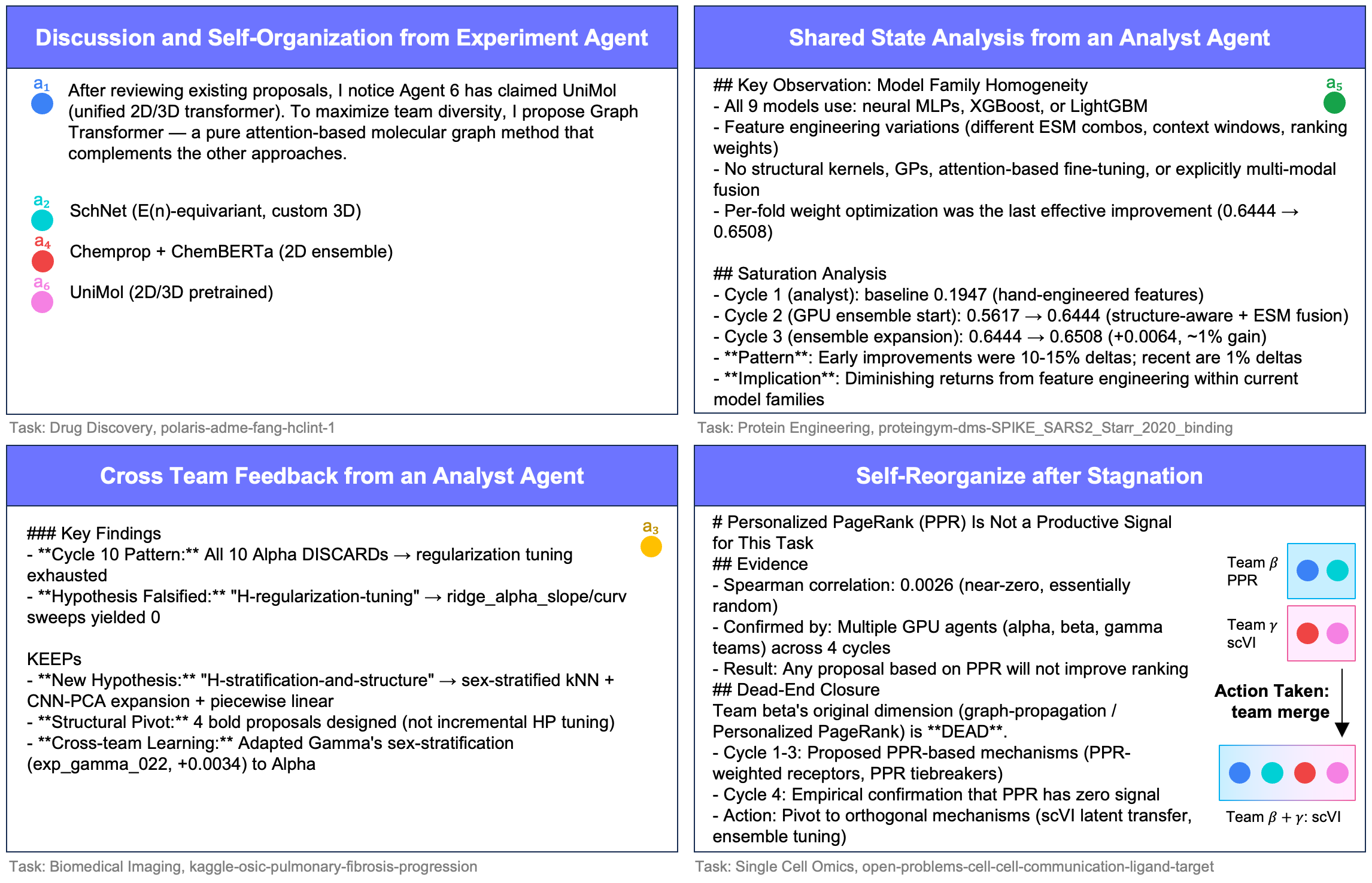}
    \caption{\textbf{Emergent coordination during long-running experimental search.} Illustrations of \name agent team interactions in long-running research experiments, featuring representative quotes from the agents.}
    \label{fig:biomlbench-case-studies}
    \vspace{-3mm}
\end{figure}

\xhdr{Results} We report aggregate performance in Fig.~\ref{fig:biomlbench} and domain-level performance in Table~\ref{tab:biomlbench}. Overall, \name achieves the highest mean (SE) leaderboard percentile among the evaluated systems, with $74.40\,(6.20)\%$ compared with $66.07\,(7.38)\%$ for Autoresearch, \revision{an improvement} of $+8.33$ leaderboard percentile points.
\name completes all 24 tasks. \name shows the \revision{largest improvement} in drug discovery, reaching $64.52\,(8.37)\%$ mean (SE) leaderboard percentile compared with $47.91\,(10.77)\%$ for Biomni.
Protein engineering is the strongest domain in absolute leaderboard percentile, but it is largely saturated\revision{,} with both \name and Autoresearch obtaining $96.97\,(3.03)\%$, \revision{and \name has a better mean rank of 1.50.}
Sec.~\ref{sec:proteingym} \revision{instead evaluates whether} \name can discover a single method that transfers across the full ProteinGym supervised substitution benchmark.
Biomedical imaging remains the most challenging domain\revision{, where each task requires training larger image models.}
We summarize \revision{the final \name approaches ($p^*$)} in Appendix~\ref{app:biomlbench_approaches}. To complement the quantitative results, we inspected the shared state and agent logs of \name to determine whether deliberation changed the experiments selected for execution. Fig.~\ref{fig:biomlbench-case-studies} shows representative examples in which agents diversified away from redundant proposals, identified saturated search directions, transferred hypotheses across teams, and retired dead-end directions after stagnation, supporting the interpretation that \name improves experiment selection under a fixed experiment compute budget.

\begin{table}[t]
\centering
\caption{BioML-Bench comparison of models under matched per domain experimental compute budgets, showing Biomni, Autoresearch, and \name. Values are mean (SE) and rank among all models.
}\label{tab:biomlbench}
\resizebox{\linewidth}{!}{
\scriptsize
\setlength{\tabcolsep}{3.2pt}
\renewcommand{\arraystretch}{0.95}
\begin{tabular}{lcccc>{\columncolor{modelcolor}}c>{\columncolor{modelcolor}}c}
\toprule
\multirow{2}{*}{\textbf{Domain}}
& \multicolumn{2}{c}{\textbf{Biomni}}
& \multicolumn{2}{c}{\textbf{Autoresearch}}
& \multicolumn{2}{c}{\cellcolor{modelcolor}\textbf{\name}} \\
\cmidrule(lr){2-3}\cmidrule(lr){4-5}\cmidrule(lr){6-7}
& \textbf{Leaderboard $\uparrow$} & \textbf{Rank $\downarrow$}
& \textbf{Leaderboard $\uparrow$} & \textbf{Rank $\downarrow$}
& \textbf{Leaderboard $\uparrow$} & \textbf{Rank $\downarrow$} \\
\midrule

Biomedical Imaging $(n=4)$
& $19.04\,(10.83)$ & 4.75
& $39.60\,(21.75)$ & 2.25
& $\mathbf{45.75}\,(\mathbf{22.18})$ & $\mathbf{1.50}$ \\

Drug Discovery $(n=9)$
& $47.91\,(10.77)$ & 3.11
& $46.16\,(10.59)$ & 2.89
& $\mathbf{64.52}\,(\mathbf{8.37})$ & $\mathbf{2.44}$ \\

Protein Engineering $(n=6)$
& $93.94\,(3.83)$ & 2.50
& $\mathbf{96.97}\,(\mathbf{3.03})$ & 2.00
& $\mathbf{96.97}\,(\mathbf{3.03})$ & $\mathbf{1.50}$ \\

Single Cell Omics $(n=5)$
& $78.00\,(10.20)$ & 3.60
& $86.00\,(9.80)$ & 2.00
& $\mathbf{88.00}\,(\mathbf{9.70})$ & $\mathbf{1.60}$ \\

\bottomrule
\end{tabular}
}
\vspace{-1mm}

{\footnotesize Full metrics are reported in Table~\ref{tab:biomlbench-full}.}
\end{table}

\subsection{GPT Training Optimization with \name}

We next evaluate whether \name generalizes beyond biomedical tasks by applying it to GPT nanochat training optimization, the language model training benchmark introduced by Autoresearch~\cite{karpathy2026autoresearch}. In this task, each experiment modifies a training program and is evaluated by the resulting validation bits-per-byte, so progress depends on selecting useful changes under a fixed experimental budget. This setting tests whether \name can coordinate search over interacting choices in architecture, optimization, and training schedule.

\xhdr{Setup}
Each experiment is a single 5-minute GPT training run on one H100 GPU, evaluated by validation bits-per-byte (\texttt{val\_bpb}, lower is better). We compare \name against single-agent Autoresearch \cite{karpathy2026autoresearch} on the same code repository. We evaluate two regimes, (i)~\emph{From Autoresearch baseline}, where both systems start from the same nanochat code at \texttt{val\_bpb} $=0.998$ and search for improvements, \revision{and} (ii)~\emph{From an \name champion}, where both systems start from the champion $p^*$ obtained after 50 prior \name experiments (\texttt{val\_bpb} $=0.9777$) and are given identical access to the negative knowledge file \texttt{EXPLORED.md} listing previously dead-ended directions. We report per experiment trajectories rather than wall clock so that the comparison isolates orchestration from hardware allocation. \revision{Since \texttt{val\_bpb} on the fixed held-out shard is queried throughout the search, it is a development validation objective.} Both regimes are compared in Appendix~\ref{app:singleagent_compare}.

\begin{figure}[t]
    \centering
    \begin{subfigure}[t]{0.49\linewidth}
        \centering
        \includegraphics[width=\linewidth]{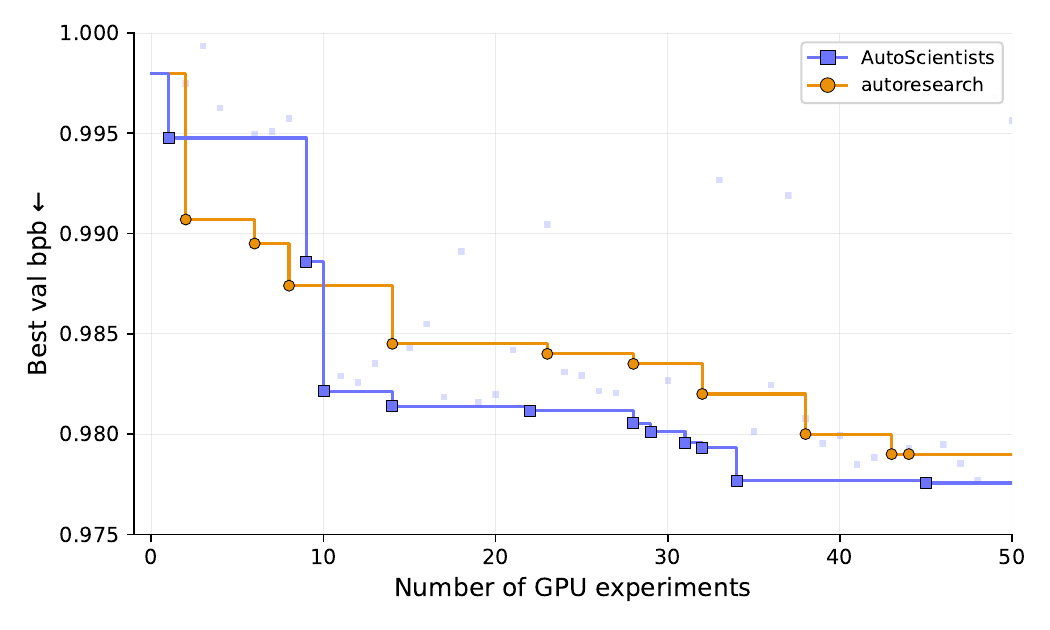}
        \subcaption{From Autoresearch baseline}
    \end{subfigure}
    \hfill
    \begin{subfigure}[t]{0.49\linewidth}
        \centering
        \includegraphics[width=\linewidth]{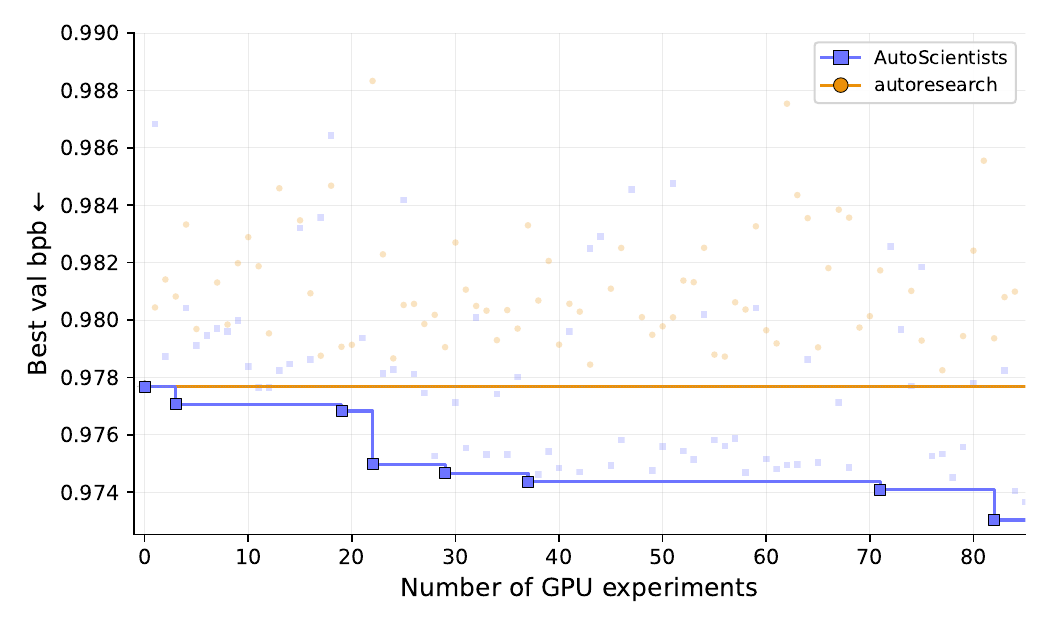}
        \subcaption{From an \name champion}
    \end{subfigure}
    \caption{\textbf{\name sustains improvement during long-running GPT training optimization.} GPT nanochat training optimization with \name and Autoresearch \cite{karpathy2026autoresearch}. \textbf{(a)~From Autoresearch baseline} (\texttt{val\_bpb} $=0.998$), \name reaches \texttt{val\_bpb} $\approx 0.978$ in 34 experiments vs.\ 65 for Autoresearch, a 1.9$\times$ speedup at the matched loss. \textbf{(b)~From an \name champion} obtained after 50 prior \name experiments (\texttt{val\_bpb} $=0.9777$) with the same dead-end registry, \name accepts 7 out of 93 experiments to reach \texttt{val\_bpb} $=0.9730$, whereas Autoresearch accepts 0 out of 100.
    }
    \vspace{-5mm}
    \label{fig:gpt_compare}
\end{figure}

\xhdr{From Autoresearch Baseline}
In the from baseline regime (Fig.~\ref{fig:gpt_compare}a), \name reaches \texttt{val\_bpb} $\approx 0.978$ in 34 experiments, while single-agent Autoresearch reaches the same value only after 65 experiments. From experiment~10 onward, \name stays lower for the remainder of the comparison window. The improvement comes from \revision{parallel exploration by self-organized teams}\revision{. Agents} in this run formed three teams (architecture, schedule, and optimizer) that propose and run experiments concurrently, so multiple research directions advance within a single agent cycle, \revision{whereas the single-agent loop advances one research direction per experiment.}

\xhdr{From an \name Champion}
In the from champion regime (Fig.~\ref{fig:gpt_compare}b), both methods start from the same \name champion at \texttt{val\_bpb} $=0.9777$. \name accepts seven improvements over 93 experiments and reaches \texttt{val\_bpb} $=0.9730$. Single-agent Autoresearch accepts \emph{zero} improvements over 100 experiments, and its best attempt reaches \texttt{val\_bpb} $=0.9783$. The seven \name improvements span heterogeneous research directions (query-key normalization order, matrix initialization, value embedding gate width, final learning rate fraction, softcap value, compile autotuning, and a noise-floor recalibration of the starting champion) rather than concentrating on a single direction. The first improvement \name discovered, query-key normalization order, was never proposed by the single-agent loop in any of its 100 attempts, indicating that the \revision{improvement comes from a wider set of hypotheses rather than from more compute}. In contrast, the single-agent loop repeatedly perturbed \revision{settings that were already near local optima in} the starting champion's tuned configuration and produced only null results.

\subsection{Extending Kermut on ProteinGym Supervised Fitness Prediction}
\label{sec:proteingym}

Here we evaluate whether \name can improve a strong existing scientific codebase rather than solve a benchmark from scratch. This setting reflects how scientific research is typically conducted, where researchers begin from the strongest available method and iteratively search for modifications that improve it.

\xhdr{Setup}
We use Kermut \cite{kermut}, a Gaussian process method for supervised protein variant effect prediction, as the seed program for \name \revision{because} it is the best-performing baseline. We refer to the resulting discovered model as \name-Kermut.
During development, \name is allowed to modify the Kermut implementation in any way and is evaluated on a single development assay, ACE2--Spike binding \revision{(ProteinGym assay \texttt{SPIKE\_SARS2\_Starr\_2020\_binding}, hereafter SPIKE-SARS2)}. Agents optimize the objective
$
\ell_{\mathrm{eval}}
=
\frac{1}{3}
\sum_{s}
\rho_s ,
$
where $s \in \{\mathrm{random}, \mathrm{modulo}, \mathrm{contiguous}\}$ and $\rho_s$ is the out-of-fold Spearman correlation under split scheme $s$. We choose this assay because Kermut obtains a low mean Spearman's $\rho$, making it a challenging development target. \name is run for 10 cycles with no human intervention and access to one H100 GPU. One cycle corresponds to each experiment agent completing one queued experiment. After development, the discovered recipe is frozen and applied without further modification to the full ProteinGym supervised substitution benchmark of 217 DMS assays.

\xhdr{Results}
\revision{On the ACE2--Spike development assay, across three independent \name runs from the Kermut model, \name-Kermut reaches $0.838 \pm 0.004$, versus $0.809 \pm 0.006$ for three single-agent Autoresearch runs from the same Kermut model and number of experiments (Appendix~\ref{app:repeated_runs}). This Kermut seeded setting differs from the cold start BioML-Bench SPIKE-SARS2 task in Table~\ref{tab:ablations_teamsize}, so the two results are not comparable.} We analyze the best performing \name-Kermut champion, where the discovered predictor is a three-GP ensemble that combines Kermut's structure kernel with expanded zero-shot features, greedy diversity-based feature selection, and quantile warped targets. \name explores research directions beyond hyperparameter tuning of Kermut. Architectural details are provided in Appendix~\ref{app:proteingym} with ablations of \name-introduced components in Appendix~\ref{app:kermut-ablations}. We then evaluate the frozen \name-Kermut recipe across all 217 supervised substitution DMS assays in ProteinGym. As shown in Table~\ref{tab:proteingym}, \name-Kermut improves the official average Spearman's $\rho$ from 0.657 for Kermut to 0.700, an absolute \revision{improvement} of 0.043 and a 6.5\% relative improvement. The improvement in Spearman's $\rho$ is observed across all three CV schemes. \name-Kermut also outperforms the other supervised baselines considered. \revision{The discovered quantile warping and rank oriented model selection improve variant ordering but not calibrated regression, as MSE increases by 0.006.} Extending \name to optimize multi-objective leaderboards, including MSE, is an important direction for future work.
\vspace{-2mm}
\begin{table}[h]
\centering
\caption{
Performance on the ProteinGym supervised substitution benchmark,
comprising 217 DMS assays across UniProt/function
groups. Best results are in \textbf{bold} and second best in \textit{italic}.
Values are mean (SE).
}
\label{tab:proteingym}
\resizebox{\linewidth}{!}{
\begin{tabular}{llcccccccc}
\toprule
\textbf{Model type} & \textbf{Model name} & \multicolumn{4}{c}{\textbf{Spearman's $\mathbf{\rho}$} ($\uparrow$)} & \multicolumn{4}{c}{\textbf{MSE} ($\downarrow$)} \\
\cmidrule(lr){3-6} \cmidrule(lr){7-10}
& & Contig. & Mod. & Rand. & Avg. & Contig. & Mod. & Rand. & Avg. \\
\midrule
Embed. & ESM-1v Embeddings
& \begin{tabular}{c}0.479 \\ (0.015)\end{tabular}
& \begin{tabular}{c}0.514 \\ (0.014)\end{tabular}
& \begin{tabular}{c}0.614 \\ (0.019)\end{tabular}
& \begin{tabular}{c}0.535 \\ (0.009)\end{tabular}
& \begin{tabular}{c}0.914 \\ (0.053)\end{tabular}
& \begin{tabular}{c}0.848 \\ (0.071)\end{tabular}
& \begin{tabular}{c}0.603 \\ (0.055)\end{tabular}
& \begin{tabular}{c}0.789 \\ (0.035)\end{tabular} \\
& MSA Transformer Embeddings
& \begin{tabular}{c}0.513 \\ (0.019)\end{tabular}
& \begin{tabular}{c}0.562 \\ (0.016)\end{tabular}
& \begin{tabular}{c}0.670 \\ (0.012)\end{tabular}
& \begin{tabular}{c}0.581 \\ (0.009)\end{tabular}
& \begin{tabular}{c}0.860 \\ (0.055)\end{tabular}
& \begin{tabular}{c}0.783 \\ (0.066)\end{tabular}
& \begin{tabular}{c}0.529 \\ (0.030)\end{tabular}
& \begin{tabular}{c}0.724 \\ (0.030)\end{tabular} \\
& Tranception Embeddings
& \begin{tabular}{c}0.439 \\ (0.014)\end{tabular}
& \begin{tabular}{c}0.525 \\ (0.011)\end{tabular}
& \begin{tabular}{c}0.681 \\ (0.015)\end{tabular}
& \begin{tabular}{c}0.548 \\ (0.008)\end{tabular}
& \begin{tabular}{c}1.037 \\ (0.048)\end{tabular}
& \begin{tabular}{c}0.849 \\ (0.052)\end{tabular}
& \begin{tabular}{c}0.518 \\ (0.043)\end{tabular}
& \begin{tabular}{c}0.802 \\ (0.028)\end{tabular} \\
\midrule
NPT & ProteinNPT
& \begin{tabular}{c}0.529 \\ (0.018)\end{tabular}
& \begin{tabular}{c}0.588 \\ (0.013)\end{tabular}
& \begin{tabular}{c}0.741 \\ (0.015)\end{tabular}
& \begin{tabular}{c}0.619 \\ (0.009)\end{tabular}
& \begin{tabular}{c}0.856 \\ (0.051)\end{tabular}
& \begin{tabular}{c}0.765 \\ (0.056)\end{tabular}
& \begin{tabular}{c}0.441 \\ (0.046)\end{tabular}
& \begin{tabular}{c}0.687 \\ (0.029)\end{tabular} \\
\midrule
Kermut & Kermut
& \begin{tabular}{c}\textit{0.593} \\ \textit{(0.015)}\end{tabular}
& \begin{tabular}{c}\textit{0.633} \\ \textit{(0.012)}\end{tabular}
& \begin{tabular}{c}\textit{0.746} \\ \textit{(0.014)}\end{tabular}
& \begin{tabular}{c}\textit{0.657} \\ \textit{(0.008)}\end{tabular}
& \begin{tabular}{c}\textbf{0.737} \\ \textbf{(0.042)}\end{tabular}
& \begin{tabular}{c}\textit{0.666} \\ \textit{(0.050)}\end{tabular}
& \begin{tabular}{c}\textbf{0.413} \\ \textbf{(0.035)}\end{tabular}
& \begin{tabular}{c}\textbf{0.605} \\ \textbf{(0.024)}\end{tabular} \\
\rowcolor{modelcolor}
& \name
& \begin{tabular}{c}\textbf{0.635} \\ \textbf{(0.013)}\end{tabular}
& \begin{tabular}{c}\textbf{0.681} \\ \textbf{(0.011)}\end{tabular}
& \begin{tabular}{c}\textbf{0.783} \\ \textbf{(0.010)}\end{tabular}
& \begin{tabular}{c}\textbf{0.700} \\ \textbf{(0.007)}\end{tabular}
& \begin{tabular}{c}\textit{0.812} \\ \textit{(0.131)}\end{tabular}
& \begin{tabular}{c}\textbf{0.607} \\ \textbf{(0.040)}\end{tabular}
& \begin{tabular}{c}\textbf{0.413} \\ \textbf{(0.025)}\end{tabular}
& \begin{tabular}{c}\textit{0.611} \\ \textit{(0.047)}\end{tabular} \\
\bottomrule
\end{tabular}
\vspace{-4mm}
}
\end{table}

\subsection{Ablations of \name}
\label{sec:ablations}

We next test which components of \name are responsible for its performance. The ablations remove one component at a time while keeping the agent backend, task interface, starting program, and experimental budget fixed. \revision{This tests} whether \revision{improvements} come from analyst guided proposal generation, cross-agent feedback, team reorganization, or the shared experimental record.

\xhdr{Setup} We identify the contribution of four \name components on four tasks\revision{, TDC-hERG, Cell-Cell Communication, Human Plasma-Protein Binding, and GPT nanochat training optimization (Table~\ref{tab:ablations})}, holding the agent backend, task interface, total compute, and starting program fixed. The four ablations are (1)~\textbf{No analyst} removes the three analyst agents and reassigns their duties (proposal generation, knowledge file maintenance, hypothesis tracking) to the experiment agents; (2)~\textbf{No cross-agent feedback} disables comment threads on proposals and results, so agents cannot critique each other or share near-misses across teams; (3)~\textbf{No self-organization} fixes \revision{a predefined} team organization \revision{at the start of the run} and prevents agents from reorganizing teams across rounds\revision{, while keeping parallel agents, the shared state, and cross-agent feedback}; (4)~\textbf{Independent agents} removes both cross-agent feedback and the shared state (champion program, results log, dead-end registry, and accumulated knowledge), so each agent runs a solo loop maintaining only its own private state and cannot observe any other agent's results.

\xhdr{Results} The full \name wins every task on its primary metric, and the most damaging ablation differs by task. Removing the \emph{analyst} is most damaging on TDC-hERG (AUROC $0.867 \rightarrow 0.738$, leaderboard percentile $85.7 \rightarrow 14.3$). Removing \emph{cross-agent feedback} is most damaging on Human Plasma-Protein Binding (Pearson correlation $0.8729 \rightarrow 0.7144$, leaderboard percentile $80 \rightarrow 30$). Removing \emph{self-organization} is most damaging on GPT training optimization (\texttt{val\_bpb} $0.9777 \rightarrow 0.9833$). The \emph{independent agents} \revision{ablation} is most damaging on Cell-Cell Communication (Odds Ratio $0.924 \rightarrow 0.435$, the largest proportional drop in the table). No single mechanism dominates across all four tasks and each removed component produces an \name that is dominated on at least one task. We read this as evidence that the four components address complementary failure modes rather than redundantly contributing the same kind of \revision{improvement}. Analyst driven refinement matters when proposal quality is the bottleneck, cross-agent feedback matters when individual agents observe only a partial signal, self-organization matters when the productive search direction shifts during the run, and a shared experimental record matters when isolated agents would otherwise duplicate work or converge to incompatible local optima.
\revision{
In addition to removing \textit{self-organization}, we further explore how team organization affects \name performance on GPT training optimization in Appendix~\ref{app:frozen_teams}. \revision{Letting agents form initial teams that then stay fixed reaches} a \texttt{val\_bpb} of 0.9806, and starting from the same initial teams and hypotheses that \name forms reaches 0.9794, compared with 0.9777 for \name.
}
Per ablation trajectories are in Appendix~\ref{app:ablation_trajectories}.

\begin{table}[h]
\centering
\caption{Ablation results of \name for biomedicine and GPT training optimization tasks.
}
\label{tab:ablations}

\resizebox{\linewidth}{!}{%
\begin{tabular}{llcccc>{\columncolor{modelcolor}}c}
\toprule
\textbf{Task} & \textbf{Metric}
& \textbf{No analyst}
& \textbf{No cross-agent}
& \textbf{No self-org.}
& \textbf{Independent agents}
& \cellcolor{modelcolor}\textbf{\name} \\
\midrule

{\textbf{TDC-hERG}}
& AUROC ($\uparrow$)
& 0.738 & 0.819 & 0.807 & 0.853 & \textbf{0.867} \\

{\textbf{Cell-Cell Communication}}
& Odds Ratio ($\uparrow$)
& 0.858 & 0.908 & 0.628 & 0.435 & \textbf{0.924} \\

{\textbf{Human Plasma-Protein Binding}}
& Pearson correlation ($\uparrow$)
& 0.813 & 0.714 & 0.811 & 0.784 & \textbf{0.873} \\

\textbf{GPT Training Optimization}
& Best \texttt{val\_bpb} $(\downarrow)$
& 0.9817 & 0.9814 & 0.9833 & 0.9833 & \textbf{0.9777}  \\
\bottomrule
\end{tabular}%
}
\end{table}

\section{Limitations and Future Work}

\name is not designed to be more LLM call efficient than single-agent baselines. As shown in Table~\ref{tab:llm_usage_by_task}, \name uses more LLM tokens than Autoresearch, though within the same order of magnitude, reflecting its use of multiple agents for parallel reasoning, discussion, and team reorganization. Instead, \name is designed to improve experimental search under a fixed experimental compute budget by enabling teams of agents to explore and collaborate over the design space.
\revision{Our comparisons match experimental compute but not LLM token usage, and the higher token usage of \name may limit accessibility. Token usage could be reduced with LLM routing~\cite{ong2025routellm}, deactivating low contributing agents~\cite{liu2024dylan}, or pruning messages that do not affect decisions~\cite{zhang2025cut}.
Although \name uses Claude Code in the main experiments, runs with an OpenAI Codex backend reach consistent results (Appendix~\ref{app:repeated_runs}).}
As part of the matched experimental compute budget used for fair comparison on BioML-Bench, we restricted \name to one H100 GPU per task so GPU-bound experiments in that evaluation were executed sequentially. This setting evaluates experiment selection under matched compute but does not fully exercise \name's capacity for parallel experimentation. When multiple GPUs are available, \name can dispatch experiments concurrently. \revision{With 2 and 4 GPUs, wall-clock time per experiment on GPT nanochat drops by $1.52\times$ and $2.18\times$, and the speedup is sublinear because the LLM reasoning in each experiment does not shrink (Appendix~\ref{app:teamsize_trajectories}).} Evaluating how \name scales with larger GPU budgets \revision{and asynchronous dispatch} remains future work.
Additionally, the number of agents is set before \revision{the run, and nine agents is not optimal for every task}. In future work, we will \revision{scale team size dynamically with task difficulty}. We provide preliminary exploration of different \name team sizes \revision{and practical guidance} in Appendix~\ref{app:teamsize_trajectories}.

\vspace{-2mm}
\section{Conclusion}
We introduced \name, a self-organizing agent team for long-horizon scientific experimentation. \name addresses a limitation of existing AI agents, which can run individual experiments but often struggle to sustain experimental search as evidence accumulates and productive directions change. Across BioML-Bench, GPT training optimization, and ProteinGym, \name improves over state-of-the-art AI agents under matched experimental budgets.

\name makes long-running experimentation a collective search process. Agents evaluate proposals before execution, record successful and failed directions, share evidence through a common state, and reorganize teams when progress stalls. This design helps agents use experimental trials more effectively, avoid repeated dead ends, and continue identifying productive modifications after individual agents plateau.

{\small \section*{Acknowledgments}

A.F. is supported by the Kempner Graduate Fellowship at Harvard University.
We gratefully acknowledge the support by NSF CAREER Award 2339524, ARPA-H Biomedical Data Fabric (BDF) Toolbox Program, Amazon Faculty Research, Google Research Scholar Program, AstraZeneca Research, GlaxoSmithKline Award, Roche Alliance with Distinguished Scientists (ROADS) Program, Sanofi iDEA-iTECH Award, Boehringer Ingelheim Award, Merck Award, Optum AI Research Collaboration Award, Pfizer Research, Gates Foundation (INV-079038), Chan Zuckerberg Initiative, Collaborative Center for XDP at Massachusetts General Hospital, John and Virginia Kaneb Fellowship at Harvard Medical School, Biswas Computational Biology Initiative in partnership with the Milken Institute, Harvard Medical School Dean’s Innovation Fund for the Use of Artificial Intelligence, and the Kempner Institute for the Study of Natural and Artificial Intelligence at Harvard University.  This work was delivered as part of the AURORA project supported by the Cancer Grand Challenges partnership funded by Cancer Research UK ([CGCAI1-Mar26/100001]).
Any opinions, findings, conclusions or recommendations expressed in this material are those of the authors and do not necessarily reflect the views of the funders.
}

\bibliographystyle{unsrtnat}
\bibliography{refs}

@article{yang2026understanding,
  title={Understanding Agent Scaling in {LLM}-Based Multi-Agent Systems via Diversity},
  author={Yang, Yingxuan and Qu, Chengrui and Wen, Muning and Shi, Laixi and Wen, Ying and Zhang, Weinan and Wierman, Adam and Gu, Shangding},
  journal={arXiv preprint arXiv:2602.03794},
  year={2026}
}

@article{gao2024empowering,
  title={Empowering biomedical discovery with AI agents},
  author={Gao, Shanghua and Fang, Ada and Huang, Yepeng and Giunchiglia, Valentina and Noori, Ayush and Schwarz, Jonathan Richard and Ektefaie, Yasha and Kondic, Jovana and Zitnik, Marinka},
  journal={Cell},
  volume={187},
  number={22},
  pages={6125--6151},
  year={2024},
  publisher={Elsevier}
}

@inproceedings{
huang2025on,
title={On the Resilience of {LLM}-Based Multi-Agent Collaboration with Faulty Agents},
author={Jen-tse Huang and Jiaxu Zhou and Tailin Jin and Xuhui Zhou and Zixi Chen and Wenxuan Wang and Youliang Yuan and Michael Lyu and Maarten Sap},
booktitle={Forty-second International Conference on Machine Learning},
year={2025},
url={https://openreview.net/forum?id=bkiM54QftZ}
}

@article{pappu2026multi,
  title={Multi-Agent Teams Hold Experts Back},
  author={Pappu, Aneesh and El, Batu and Cao, Hancheng and di Nolfo, Carmelo and Sun, Yanchao and Cao, Meng and Zou, James},
  journal={arXiv preprint arXiv:2602.01011},
  year={2026}
}

@article{zhuge2024language,
  title={Language agents as optimizable graphs},
  author={Zhuge, Mingchen and Wang, Wenyi and Kirsch, Louis and Faccio, Francesco and Khizbullin, Dmitrii and Schmidhuber, J{\"u}rgen},
  journal={arXiv preprint arXiv:2402.16823},
  year={2024}
}

@article{wuchty2007increasing,
  title={The increasing dominance of teams in production of knowledge},
  author={Wuchty, Stefan and Jones, Benjamin F and Uzzi, Brian},
  journal={Science},
  volume={316},
  number={5827},
  pages={1036--1039},
  year={2007},
  publisher={American Association for the Advancement of Science}
}

@article{cummings2005collaborative,
  title={Collaborative research across disciplinary and organizational boundaries},
  author={Cummings, Jonathon N and Kiesler, Sara},
  journal={Social studies of science},
  volume={35},
  number={5},
  pages={703--722},
  year={2005},
  publisher={Sage Publications Sage CA: Thousand Oaks, CA}
}

@article{hall2018science,
  title={The science of team science: A review of the empirical evidence and research gaps on collaboration in science.},
  author={Hall, Kara L and Vogel, Amanda L and Huang, Grace C and Serrano, Katrina J and Rice, Elise L and Tsakraklides, Sophia P and Fiore, Stephen M},
  journal={American psychologist},
  volume={73},
  number={4},
  pages={532},
  year={2018},
  publisher={American Psychological Association}
}

@article{xu2022flat,
  title={Flat teams drive scientific innovation},
  author={Xu, Fengli and Wu, Lingfei and Evans, James},
  journal={Proceedings of the National Academy of Sciences},
  volume={119},
  number={23},
  pages={e2200927119},
  year={2022},
  publisher={National Academy of Sciences}
}

@article{swanson2025virtual,
  title={The Virtual Lab of AI agents designs new SARS-CoV-2 nanobodies},
  author={Swanson, Kyle and Wu, Wesley and Bulaong, Nash L and Pak, John E and Zou, James},
  journal={Nature},
  volume={646},
  number={8085},
  pages={716--723},
  year={2025},
  publisher={Nature Publishing Group UK London}
}

@article{kim2025towards,
  title={Towards a science of scaling agent systems},
  author={Kim, Yubin and Gu, Ken and Park, Chanwoo and Park, Chunjong and Schmidgall, Samuel and Heydari, A Ali and Yan, Yao and Zhang, Zhihan and Zhuang, Yuchen and Malhotra, Mark and others},
  journal={arXiv preprint arXiv:2512.08296},
  year={2025}
}

@article{lu2026towards,
  title={Towards end-to-end automation of AI research},
  author={Lu, Chris and Lu, Cong and Lange, Robert Tjarko and Yamada, Yutaro and Hu, Shengran and Foerster, Jakob and Ha, David and Clune, Jeff},
  journal={Nature},
  volume={651},
  number={8107},
  pages={914--919},
  year={2026},
  publisher={Nature Publishing Group UK London}
}

@article{huang2025biomni,
  title={{Biomni}: A general-purpose biomedical {AI} agent},
  author={Huang, Kexin and Zhang, Serena and Wang, Hanchen and Qu, Yuanhao and Lu, Yingzhou and Roohani, Yusuf and Li, Ryan and Qiu, Lin and Li, Gavin and Zhang, Junze and others},
  journal={biorxiv},
  year={2025}
}

@article{mitchener2025kosmos,
  title={{Kosmos}: An {AI} scientist for autonomous discovery},
  author={Mitchener, Ludovico and Yiu, Angela and Chang, Benjamin and Bourdenx, Mathieu and Nadolski, Tyler and Sulovari, Arvis and Landsness, Eric C and Barabasi, Daniel L and Narayanan, Siddharth and Evans, Nicky and others},
  journal={arXiv preprint arXiv:2511.02824},
  year={2025}
}

@article{qu2026coral,
  title={CORAL: Towards Autonomous Multi-Agent Evolution for Open-Ended Discovery},
  author={Qu, Ao and Zheng, Han and Zhou, Zijian and Yan, Yihao and Tang, Yihong and Ong, Shao Yong and Hong, Fenglu and Zhou, Kaichen and Jiang, Chonghe and Kong, Minwei and others},
  journal={arXiv preprint arXiv:2604.01658},
  year={2026}
}

@article {jin2025stella,
	author = {Jin, Ruofan and Xu, Mingyang and Meng, Fei and Wan, Guancheng and Cai, Qingran and Jiang, Yize and Han, Jin and Chen, Yuanyuan and Lu, Wanqing and Wang, Mengyang and Lan, Zhiqian and Jiang, Yuxuan and Liu, Junhong and Wang, Dongyao and Cong, Le and Zhang, Zaixi},
	title = {STELLA: Towards a Biomedical World Model with Self-Evolving Multimodal Agents},
	elocation-id = {2025.07.01.662467},
	year = {2026},
	doi = {10.1101/2025.07.01.662467},
	publisher = {Cold Spring Harbor Laboratory},
	URL = {https://www.biorxiv.org/content/early/2026/01/25/2025.07.01.662467},
	eprint = {https://www.biorxiv.org/content/early/2026/01/25/2025.07.01.662467.full.pdf},
	journal = {bioRxiv}
}

@article {sui2026medea,
	author = {Sui, Pengwei and Li, Michelle M. and Gao, Shanghua and Shen, Wanxiang and Giunchiglia, Valentina and Shen, Andrew and Huang, Yepeng and Kong, Zhenglun and Zitnik, Marinka},
	title = {Medea: An omics AI agent for therapeutic discovery},
	elocation-id = {2026.01.16.696667},
	year = {2026},
	doi = {10.64898/2026.01.16.696667},
	publisher = {Cold Spring Harbor Laboratory},
	URL = {https://www.biorxiv.org/content/early/2026/01/20/2026.01.16.696667},
	eprint = {https://www.biorxiv.org/content/early/2026/01/20/2026.01.16.696667.full.pdf},
	journal = {bioRxiv}
}

@article{gottweis2025towards,
  title={Towards an AI co-scientist},
  author={Gottweis, Juraj and Weng, Wei-Hung and Daryin, Alexander and Tu, Tao and Palepu, Anil and Sirkovic, Petar and Myaskovsky, Artiom and Weissenberger, Felix and Rong, Keran and Tanno, Ryutaro and others},
  journal={arXiv preprint arXiv:2502.18864},
  year={2025}
}

@article{penades2025ai,
  title={AI mirrors experimental science to uncover a mechanism of gene transfer crucial to bacterial evolution},
  author={Penad{\'e}s, Jos{\'e} R and Gottweis, Juraj and He, Lingchen and Patkowski, Jonasz B and Daryin, Alexander and Weng, Wei-Hung and Tu, Tao and Palepu, Anil and Myaskovsky, Artiom and Pawlosky, Annalisa and others},
  journal={Cell},
  volume={188},
  number={23},
  pages={6654--6665},
  year={2025},
  publisher={Elsevier}
}

@misc{karpathy2026autoresearch,
  author       = {Karpathy, Andrej},
  title        = {Autoresearch: {AI} agents running research on single-{GPU} nanochat training automatically},
  year         = {2026},
  howpublished = {\url{https://github.com/karpathy/autoresearch}},
  note         = {GitHub repository}
}

@article{gao2025txagent,
  title={TxAgent: an AI agent for therapeutic reasoning across a universe of tools},
  author={Gao, Shanghua and Zhu, Richard and Kong, Zhenglun and Noori, Ayush and Su, Xiaorui and Ginder, Curtis and Tsiligkaridis, Theodoros and Zitnik, Marinka},
  journal={arXiv preprint arXiv:2503.10970},
  year={2025}
}

@article {miller2025bioml,
	author = {Miller, Henry E. and Greenig, Matthew and Tenmann, Benjamin and Wang, Bo},
	title = {{BioML}-Bench: Evaluation of {AI} Agents for End-to-End Biomedical {ML}},
	elocation-id = {2025.09.01.673319},
	year = {2025},
	doi = {10.1101/2025.09.01.673319},
	publisher = {Cold Spring Harbor Laboratory},
	URL = {https://www.biorxiv.org/content/early/2025/09/28/2025.09.01.673319},
	eprint = {https://www.biorxiv.org/content/early/2025/09/28/2025.09.01.673319.full.pdf},
	journal = {bioRxiv}
}

@article{feng2026internagent,
  title={Internagent-1.5: A unified agentic framework for long-horizon autonomous scientific discovery},
  author={Feng, Shiyang and Ma, Runmin and Yan, Xiangchao and Fan, Yue and Hu, Yusong and Huang, Songtao and Zhang, Shuaiyu and Cao, Zongsheng and Peng, Tianshuo and Yuan, Jiakang and others},
  journal={arXiv preprint arXiv:2602.08990},
  year={2026}
}

@article{novikov2025alphaevolve,
  title={{AlphaEvolve}: A coding agent for scientific and algorithmic discovery},
  author={Novikov, Alexander and V{\~u}, Ng{\^a}n and Eisenberger, Marvin and Dupont, Emilien and Huang, Po-Sen and Wagner, Adam Zsolt and Shirobokov, Sergey and Kozlovskii, Borislav and Ruiz, Francisco JR and Mehrabian, Abbas and others},
  journal={arXiv preprint arXiv:2506.13131},
  year={2025}
}

@article{riedl2025emergent,
  title={Emergent coordination in multi-agent language models},
  author={Riedl, Christoph},
  journal={arXiv preprint arXiv:2510.05174},
  year={2025}
}

@inproceedings{kermut,
 author = {Groth, Peter M\o rch and Kerrn, Mads Herbert and Olsen, Lars and Salomon, Jesper and Boomsma, Wouter},
 booktitle = {Advances in Neural Information Processing Systems},
 doi = {10.52202/079017-0929},
 editor = {A. Globerson and L. Mackey and D. Belgrave and A. Fan and U. Paquet and J. Tomczak and C. Zhang},
 pages = {29514--29565},
 publisher = {Curran Associates, Inc.},
 title = {Kermut: Composite kernel regression for protein variant effects},
 url = {https://proceedings.neurips.cc/paper_files/paper/2024/file/34547650b2ca69d91f3b3c3ae8b21962-Paper-Conference.pdf},
 volume = {37},
 year = {2024}
}

@inproceedings{proteingym,
 author = {Notin, Pascal and Kollasch, Aaron and Ritter, Daniel and van Niekerk, Lood and Paul, Steffanie and Spinner, Han and Rollins, Nathan and Shaw, Ada and Orenbuch, Rose and Weitzman, Ruben and Frazer, Jonathan and Dias, Mafalda and Franceschi, Dinko and Gal, Yarin and Marks, Debora},
 booktitle = {Advances in Neural Information Processing Systems},
 editor = {A. Oh and T. Naumann and A. Globerson and K. Saenko and M. Hardt and S. Levine},
 pages = {64331--64379},
 publisher = {Curran Associates, Inc.},
 title = {ProteinGym: Large-Scale Benchmarks for Protein Fitness Prediction and Design},
 url = {https://proceedings.neurips.cc/paper_files/paper/2023/file/cac723e5ff29f65e3fcbb0739ae91bee-Paper-Datasets_and_Benchmarks.pdf},
 volume = {36},
 year = {2023}
}

@article{jiang2025aide,
  title={{AIDE}: {AI}-driven exploration in the space of code},
  author={Jiang, Zhengyao and Schmidt, Dominik and Srikanth, Dhruv and Xu, Dixing and Kaplan, Ian and Jacenko, Deniss and Wu, Yuxiang},
  journal={arXiv preprint arXiv:2502.13138},
  year={2025}
}

@article{huang2023mlagentbench,
  title={Mlagentbench: Evaluating language agents on machine learning experimentation},
  author={Huang, Qian and Vora, Jian and Liang, Percy and Leskovec, Jure},
  journal={arXiv preprint arXiv:2310.03302},
  year={2023}
}

@inproceedings{mitchell2019model,
  title={Model cards for model reporting},
  author={Mitchell, Margaret and Wu, Simone and Zaldivar, Andrew and Barnes, Parker and Vasserman, Lucy and Hutchinson, Ben and Spitzer, Elena and Raji, Inioluwa Deborah and Gebru, Timnit},
  booktitle={Proceedings of the conference on fairness, accountability, and transparency},
  pages={220--229},
  year={2019}
}

@article{zhu2026toward,
  title={Toward ultra-long-horizon agentic science: Cognitive accumulation for machine learning engineering},
  author={Zhu, Xinyu and Cai, Yuzhu and Liu, Zexi and Zheng, Bingyang and Wang, Cheng and Ye, Rui and Chen, Jiaao and Wang, Hanrui and Wang, Wei-Chen and Zhang, Yuzhi and others},
  journal={arXiv preprint arXiv:2601.10402},
  year={2026}
}

@article{berdoz2026can,
  title={Can AI Agents Agree?},
  author={Berdoz, Fr{\'e}d{\'e}ric and Rugli, Leonardo and Wattenhofer, Roger},
  journal={arXiv preprint arXiv:2603.01213},
  year={2026}
}

@article{schmidgall2025agent,
  title={Agent laboratory: Using {LLM} agents as research assistants},
  author={Schmidgall, Samuel and Su, Yusheng and Wang, Ze and Sun, Ximeng and Wu, Jialian and Yu, Xiaodong and Liu, Jiang and Moor, Michael and Liu, Zicheng and Barsoum, Emad},
  journal={Findings of the Association for Computational Linguistics: EMNLP 2025},
  pages={5977--6043},
  year={2025},
  publisher={Association for Computational Linguistics}
}

@article{ghareeb2025robin,
  title={Robin: A multi-agent system for automating scientific discovery},
  author={Ghareeb, Ali Essam and Chang, Benjamin and Mitchener, Ludovico and Yiu, Angela and Szostkiewicz, Caralyn J and Laurent, Jon M and Razzak, Muhammed T and White, Andrew D and Hinks, Michaela M and Rodriques, Samuel G},
  journal={arXiv preprint arXiv:2505.13400},
  year={2025}
}

@article{pu2025piflow,
  title={Piflow: Principle-aware scientific discovery with multi-agent collaboration},
  author={Pu, Yingming and Lin, Tao and Chen, Hongyu},
  journal={arXiv preprint arXiv:2505.15047},
  year={2025}
}

@article{bu2026empowering,
  title={Empowering AI data scientists using a multi-agent LLM framework with self-evolving capabilities for autonomous, tool-aware biomedical data analyses},
  author={Bu, Dechao and Sun, Jingbo and Li, Kun and He, Zihao and Huang, Wei and Hu, Jinlin and Zhang, Shanshan and Lei, Shuangshuang and Huo, Peipei and Wang, Zhihao and others},
  journal={Nature Biomedical Engineering},
  pages={1--16},
  year={2026},
  publisher={Nature Publishing Group UK London}
}

@article{ding2025scitoolagent,
  title={Scitoolagent: a knowledge-graph-driven scientific agent for multitool integration},
  author={Ding, Keyan and Yu, Jing and Huang, Junjie and Yang, Yuchen and Zhang, Qiang and Chen, Huajun},
  journal={Nature Computational Science},
  volume={5},
  number={10},
  pages={962--972},
  year={2025},
  publisher={Nature Publishing Group US New York}
}

@article{liu2025genomas,
  title={{GenoMAS}: A multi-agent framework for scientific discovery via code-driven gene expression analysis},
  author={Liu, Haoyang and Li, Yijiang and Wang, Haohan},
  journal={arXiv preprint arXiv:2507.21035},
  year={2025}
}

@article{han2025chembomas,
  title={{ChemBOMAS}: Accelerated {BO} in chemistry with {LLM}-enhanced multi-agent system},
  author={Han, Dong and Ai, Zhehong and Cai, Pengxiang and Lu, Shanya and Chen, Jianpeng and Ye, Zihao and Sun, Shuzhou and Gao, Ben and Ge, Lingli and Wang, Weida and others},
  journal={arXiv preprint arXiv:2509.08736},
  year={2025}
}

@article{xia2025sr,
  title={{SR}-Scientist: Scientific equation discovery with agentic {AI}},
  author={Xia, Shijie and Sun, Yuhan and Liu, Pengfei},
  journal={arXiv preprint arXiv:2510.11661},
  year={2025}
}

@article{huang2025cascade,
  title={{CASCADE}: Cumulative agentic skill creation through autonomous development and evolution},
  author={Huang, Xu and Chen, Junwu and Fei, Yuxing and Li, Zhuohan and Schwaller, Philippe and Ceder, Gerbrand},
  journal={arXiv preprint arXiv:2512.23880},
  year={2025}
}

@misc{wu2026selfaiselfdirectedframeworklonghorizon,
      title={{SelfAI}: A self-directed framework for long-horizon scientific discovery},
      author={Xiao Wu and Ting-Zhu Huang and Liang-Jian Deng and Xiaobing Yu and Yu Zhong and Shangqi Deng and Ufaq Khan and Jianghao Wu and Xiaofeng Liu and Imran Razzak and Xiaojun Chang and Yutong Xie},
      year={2025},
      eprint={2512.00403},
      archivePrefix={arXiv},
      primaryClass={cs.LG},
      url={https://arxiv.org/abs/2512.00403}, 
}

@misc{lyu2026evoscientistmultiagentevolvingai,
      title={{EvoScientist}: Towards Multi-Agent Evolving {AI} Scientists for End-to-End Scientific Discovery},
      author={Yougang Lyu and Xi Zhang and Xinhao Yi and Yuyue Zhao and Shuyu Guo and Wenxiang Hu and Jan Piotrowski and Jakub Kaliski and Jacopo Urbani and Zaiqiao Meng and Lun Zhou and Xiaohui Yan},
      year={2026},
      eprint={2603.08127},
      archivePrefix={arXiv},
      primaryClass={cs.CL},
      url={https://arxiv.org/abs/2603.08127}, 
}

@misc{bragg2026astabenchrigorousbenchmarkingai,
      title={{AstaBench}: Rigorous Benchmarking of {AI} Agents with a Scientific Research Suite},
      author={Jonathan Bragg and Mike D'Arcy and Nishant Balepur and Dan Bareket and Bhavana Dalvi and Sergey Feldman and Dany Haddad and Jena D. Hwang and Peter Jansen and Varsha Kishore and Bodhisattwa Prasad Majumder and Aakanksha Naik and Sigal Rahamimov and Kyle Richardson and Amanpreet Singh and Harshit Surana and Aryeh Tiktinsky and Rosni Vasu and Guy Wiener and Chloe Anastasiades and Stefan Candra and Jason Dunkelberger and Dan Emery and Rob Evans and Malachi Hamada and Regan Huff and Rodney Kinney and Matt Latzke and Jaron Lochner and Ruben Lozano-Aguilera and Cecile Nguyen and Smita Rao and Amber Tanaka and Brooke Vlahos and Peter Clark and Doug Downey and Yoav Goldberg and Ashish Sabharwal and Daniel S. Weld},
      year={2025},
      eprint={2510.21652},
      archivePrefix={arXiv},
      primaryClass={cs.AI},
      url={https://arxiv.org/abs/2510.21652}, 
}

@misc{yan2025lmrbenchevaluatingllmagents,
      title={{LMR-BENCH}: Evaluating {LLM} Agent's Ability on Reproducing Language Modeling Research},
      author={Shuo Yan and Ruochen Li and Ziming Luo and Zimu Wang and Daoyang Li and Liqiang Jing and Kaiyu He and Peilin Wu and George Michalopoulos and Yue Zhang and Ziyang Zhang and Mian Zhang and Zhiyu Chen and Xinya Du},
      year={2025},
      eprint={2506.17335},
      archivePrefix={arXiv},
      primaryClass={cs.SE},
      url={https://arxiv.org/abs/2506.17335}, 
}

@misc{zhu2025multiagentbenchevaluatingcollaborationcompetition,
      title={{MultiAgentBench}: Evaluating the Collaboration and Competition of {LLM} agents},
      author={Kunlun Zhu and Hongyi Du and Zhaochen Hong and Xiaocheng Yang and Shuyi Guo and Zhe Wang and Zhenhailong Wang and Cheng Qian and Xiangru Tang and Heng Ji and Jiaxuan You},
      year={2025},
      eprint={2503.01935},
      archivePrefix={arXiv},
      primaryClass={cs.MA},
      url={https://arxiv.org/abs/2503.01935}, 
}

@misc{tran2025multiagentcollaborationmechanismssurvey,
      title={Multi-Agent Collaboration Mechanisms: A Survey of {LLMs}},
      author={Khanh-Tung Tran and Dung Dao and Minh-Duong Nguyen and Quoc-Viet Pham and Barry O'Sullivan and Hoang D. Nguyen},
      year={2025},
      eprint={2501.06322},
      archivePrefix={arXiv},
      primaryClass={cs.AI},
      url={https://arxiv.org/abs/2501.06322}, 
}

@misc{liu2025visionautoresearchllm,
      title={A Vision for Auto Research with {LLM} Agents},
      author={Chengwei Liu and Chong Wang and Jiayue Cao and Jingquan Ge and Kun Wang and Lyuye Zhang and Ming-Ming Cheng and Penghai Zhao and Tianlin Li and Xiaojun Jia and Xiang Li and Xingshuai Li and Yang Liu and Yebo Feng and Yihao Huang and Yijia Xu and Yuqiang Sun and Zhenhong Zhou and Zhengzi Xu},
      year={2025},
      eprint={2504.18765},
      archivePrefix={arXiv},
      primaryClass={cs.AI},
      url={https://arxiv.org/abs/2504.18765}, 
}

@misc{gridach2025agenticaiscientificdiscovery,
      title={Agentic {AI} for Scientific Discovery: A Survey of Progress, Challenges, and Future Directions},
      author={Mourad Gridach and Jay Nanavati and Khaldoun Zine El Abidine and Lenon Mendes and Christina Mack},
      year={2025},
      eprint={2503.08979},
      archivePrefix={arXiv},
      primaryClass={cs.CL},
      url={https://arxiv.org/abs/2503.08979}, 
}

@inproceedings{du2024improving,
  title={Improving factuality and reasoning in language models through multiagent debate},
  author={Du, Yilun and Li, Shuang and Torralba, Antonio and Tenenbaum, Joshua B and Mordatch, Igor},
  booktitle={Forty-first international conference on machine learning},
  year={2024}
}

@misc{chen2024reconcile,
      title={{ReConcile}: Round-Table Conference Improves Reasoning via Consensus among Diverse {LLMs}},
      author={Justin Chih-Yao Chen and Swarnadeep Saha and Mohit Bansal},
      year={2024},
      eprint={2309.13007},
      archivePrefix={arXiv},
      primaryClass={cs.CL},
      url={https://arxiv.org/abs/2309.13007}, 
}

@misc{anthropic_claude_code,
  author       = {{Anthropic}},
  title        = {{Claude Code}: Overview},
  year         = {2026},
  howpublished = {\url{https://code.claude.com/docs/en/overview}},
  note         = {Product documentation. Accessed: 2026-05-06}
}

@misc{anthropic_claude_sonnet_46,
  author       = {{Anthropic}},
  title        = {{Claude Sonnet 4.6}},
  year         = {2026},
  howpublished = {\url{https://www.anthropic.com/claude/sonnet}},
  note         = {Model documentation. Model ID: \texttt{claude-sonnet-4-6}. Accessed: 2026-05-06}
}

@misc{uw-madison-gi-tract-image-segmentation,
    author = {Lee, S. L. and Yadav, P. and Li, Y. and Meudt, J. J. and Strang, J. and Hebel, D. and Alfson, A. and Olson, S. J. and Kruser, T. R. and Smilowitz, J. B. and Borchert, K. and Loritz, B. and Gharzai, L. and Karimpour, S. and Bayouth, J. and Bassetti, M. F.},
    title = {UW-Madison GI Tract Image Segmentation },
    year = {2022},
    howpublished = {\url{https://kaggle.com/competitions/uw-madison-gi-tract-image-segmentation}},
    note = {Kaggle}
}

@misc{osic-pulmonary-fibrosis-progression,
    author = {Ahmed Shahin and Carmela Wegworth and David and Elizabeth Estes and Julia Elliott and Justin Zita and SimonWalsh and Slepetys and Will Cukierski},
    title = {OSIC Pulmonary Fibrosis Progression},
    year = {2020},
    howpublished = {\url{https://kaggle.com/competitions/osic-pulmonary-fibrosis-progression}},
    note = {Kaggle}
}

@misc{histopathologic-cancer-detection,
    author = {Will Cukierski},
    title = {Histopathologic Cancer Detection},
    year = {2018},
    howpublished = {\url{https://kaggle.com/competitions/histopathologic-cancer-detection}},
    note = {Kaggle}
}

@misc{rsna-miccai-brain-tumor-radiogenomic-classification,
    author = {Adam Flanders and Chris Carr and Evan Calabrese and FelipeKitamura, MD, PhD and inversion and JeffRudie and John Mongan and Julia Elliott and Luciano Prevedello and Michelle Riopel and sprint and Spyridon Bakas and Ujjwal},
    title = {RSNA-MICCAI Brain Tumor Radiogenomic Classification},
    year = {2021},
    howpublished = {\url{https://kaggle.com/competitions/rsna-miccai-brain-tumor-radiogenomic-classification}},
    note = {Kaggle}
}

@article{huang2021therapeutics,
  title={Therapeutics data commons: Machine learning datasets and tasks for drug discovery and development},
  author={Huang, Kexin and Fu, Tianfan and Gao, Wenhao and Zhao, Yue and Roohani, Yusuf and Leskovec, Jure and Coley, Connor W and Xiao, Cao and Sun, Jimeng and Zitnik, Marinka},
  journal={arXiv preprint arXiv:2102.09548},
  year={2021}
}

@misc{polaris_hub,
  author       = {{Polaris}},
  title        = {{Polaris}: The benchmarking platform for drug discovery},
  year         = {2026},
  howpublished = {\url{https://polarishub.io/}},
  note         = {Accessed: May 2026}
}

@article{luecken2025defining,
  title={Defining and benchmarking open problems in single-cell analysis},
  author={Luecken, Malte D and Gigante, Scott and Burkhardt, Daniel B and Cannoodt, Robrecht and Strobl, Daniel C and Markov, Nikolay S and Zappia, Luke and Palla, Giovanni and Lewis, Wesley and Dimitrov, Daniel and others},
  journal={Nature Biotechnology},
  volume={43},
  number={7},
  pages={1035--1040},
  year={2025},
  publisher={Nature Publishing Group US New York}
}

@article{lin2023evolutionary,
  title={Evolutionary-scale prediction of atomic-level protein structure with a language model},
  author={Lin, Zeming and Akin, Halil and Rao, Roshan and Hie, Brian and Zhu, Zhongkai and Lu, Wenting and Smetanin, Nikita and Verkuil, Robert and Kabeli, Ori and Shmueli, Yaniv and others},
  journal={Science},
  volume={379},
  number={6637},
  pages={1123--1130},
  year={2023},
  publisher={American Association for the Advancement of Science}
}

@article{dauparas2022robust,
  title={Robust deep learning--based protein sequence design using ProteinMPNN},
  author={Dauparas, Justas and Anishchenko, Ivan and Bennett, Nathaniel and Bai, Hua and Ragotte, Robert J and Milles, Lukas F and Wicky, Basile IM and Courbet, Alexis and de Haas, Rob J and Bethel, Neville and others},
  journal={Science},
  volume={378},
  number={6615},
  pages={49--56},
  year={2022},
  publisher={American Association for the Advancement of Science}
}

@article{chithrananda2020chemberta,
  title={ChemBERTa: large-scale self-supervised pretraining for molecular property prediction},
  author={Chithrananda, Seyone and Grand, Gabriel and Ramsundar, Bharath},
  journal={arXiv preprint arXiv:2010.09885},
  year={2020}
}

@article{heid2023chemprop,
  title={Chemprop: a machine learning package for chemical property prediction},
  author={Heid, Esther and Greenman, Kevin P and Chung, Yunsie and Li, Shih-Cheng and Graff, David E and Vermeire, Florence H and Wu, Haoyang and Green, William H and McGill, Charles J},
  journal={Journal of chemical information and modeling},
  volume={64},
  number={1},
  pages={9--17},
  year={2024},
  publisher={ACS Publications}
}

@misc{rdkit,
  title        = {{RDKit}: Open-source cheminformatics},
  year         = {2026},
  howpublished = {\url{https://www.rdkit.org}},
  note         = {Accessed: May 2026}
}

@inproceedings{xgboost,
 author = {Chen, Tianqi and Guestrin, Carlos},
 title = {{XGBoost}: A Scalable Tree Boosting System},
 booktitle = {Proceedings of the 22nd ACM SIGKDD International Conference on Knowledge Discovery and Data Mining},
 series = {KDD '16},
 year = {2016},
 isbn = {978-1-4503-4232-2},
 location = {San Francisco, California, USA},
 pages = {785--794},
 numpages = {10},
 url = {http://doi.acm.org/10.1145/2939672.2939785},
 doi = {10.1145/2939672.2939785},
 acmid = {2939785},
 publisher = {ACM},
 address = {New York, NY, USA},
}

@article{ke2017lightgbm,
  title={Lightgbm: A highly efficient gradient boosting decision tree},
  author={Ke, Guolin and Meng, Qi and Finley, Thomas and Wang, Taifeng and Chen, Wei and Ma, Weidong and Ye, Qiwei and Liu, Tie-Yan},
  journal={Advances in neural information processing systems},
  volume={30},
  year={2017}
}

@article{dorogush2018catboost,
  title={CatBoost: gradient boosting with categorical features support},
  author={Dorogush, Anna Veronika and Ershov, Vasily and Gulin, Andrey},
  journal={arXiv preprint arXiv:1810.11363},
  year={2018}
}

@article{paszke2019pytorch,
  title={Pytorch: An imperative style, high-performance deep learning library},
  author={Paszke, Adam and Gross, Sam and Massa, Francisco and Lerer, Adam and Bradbury, James and Chanan, Gregory and Killeen, Trevor and Lin, Zeming and Gimelshein, Natalia and Antiga, Luca and others},
  journal={Advances in neural information processing systems},
  volume={32},
  year={2019}
}

@article{yang2023masked,
  title={Masked inverse folding with sequence transfer for protein representation learning},
  author={Yang, Kevin K and Zanichelli, Niccol{\`o} and Yeh, Hugh},
  journal={Protein Engineering, Design and Selection},
  volume={36},
  pages={gzad015},
  year={2023},
  publisher={Oxford University Press}
}

@article{tan2025venusrem,
    author = {Tan, Yang and Wang, Ruilin and Wu, Banghao and Hong, Liang and Zhou, Bingxin},
    title = {From high-throughput evaluation to wet-lab studies: advancing mutation effect prediction with a retrieval-enhanced model},
    journal = {Bioinformatics},
    volume = {41},
    number = {Supplement 1},
    pages = {i401-i409},
    year = {2025},
    month = {07},
    doi = {10.1093/bioinformatics/btaf189},
    url = {https://doi.org/10.1093/bioinformatics/btaf189},
    eprint = {https://academic.oup.com/bioinformatics/article-pdf/41/Supplement_1/i401/63745466/btaf189.pdf},
}

@article{li2024prosst,
  title={ProSST: Protein language modeling with quantized structure and disentangled attention},
  author={Li, Mingchen and Tan, Yang and Ma, Xinzhu and Zhong, Bozitao and Yu, Huiqun and Zhou, Ziyi and Ouyang, Wanli and Zhou, Bingxin and Tan, Pan and Hong, Liang},
  journal={Advances in Neural Information Processing Systems},
  volume={37},
  pages={35700--35726},
  year={2024}
}

@article{tsishyn2025residue,
  title={Residue conservation and solvent accessibility are (almost) all you need for predicting mutational effects in proteins},
  author={Tsishyn, Matsvei and Hermans, Pauline and Rooman, Marianne and Pucci, Fabrizio},
  journal={Bioinformatics},
  volume={41},
  number={6},
  pages={btaf322},
  year={2025},
  publisher={Oxford University Press}
}

@article{tekpinar2025prescott,
  title={PRESCOTT: a population aware, epistatic, and structural model accurately predicts missense effects},
  author={Tekpinar, Mustafa and David, Laurent and Henry, Thomas and Carbone, Alessandra},
  journal={Genome Biology},
  volume={26},
  number={1},
  pages={113},
  year={2025},
  publisher={Springer}
}

@article{chen2025xtrimopglm,
  title={xTrimoPGLM: unified 100-billion-parameter pretrained transformer for deciphering the language of proteins},
  author={Chen, Bo and Cheng, Xingyi and Li, Pan and Geng, Yangli-ao and Gong, Jing and Li, Shen and Bei, Zhilei and Tan, Xu and Wang, Boyan and Zeng, Xin and others},
  journal={Nature Methods},
  volume={22},
  number={5},
  pages={1028--1039},
  year={2025},
  publisher={Nature Publishing Group US New York}
}

@inproceedings{
su2024saprot,
title={SaProt: Protein Language Modeling with Structure-aware Vocabulary},
author={Jin Su and Chenchen Han and Yuyang Zhou and Junjie Shan and Xibin Zhou and Fajie Yuan},
booktitle={The Twelfth International Conference on Learning Representations},
year={2024},
url={https://openreview.net/forum?id=6MRm3G4NiU}
}

@inproceedings{hsu2022learning,
  title={Learning inverse folding from millions of predicted structures},
  author={Hsu, Chloe and Verkuil, Robert and Liu, Jason and Lin, Zeming and Hie, Brian and Sercu, Tom and Lerer, Adam and Rives, Alexander},
  booktitle={International conference on machine learning},
  pages={8946--8970},
  year={2022},
  organization={PMLR}
}

@article{alley2019unified,
  title={Unified rational protein engineering with sequence-based deep representation learning},
  author={Alley, Ethan C and Khimulya, Grigory and Biswas, Surojit and AlQuraishi, Mohammed and Church, George M},
  journal={Nature methods},
  volume={16},
  number={12},
  pages={1315--1322},
  year={2019},
  publisher={Nature Publishing Group US New York}
}

@article{zhang2024multi,
  title={Multi-scale representation learning for protein fitness prediction},
  author={Zhang, Zuobai and Notin, Pascal and Huang, Yining and Lozano, Aur{\'e}lie and Chenthamarakshan, Vijil and Marks, Debora and Das, Payel and Tang, Jian},
  journal={Advances in Neural Information Processing Systems},
  volume={37},
  pages={101456--101473},
  year={2024}
}

@inproceedings{rao2021msa,
  title={MSA transformer},
  author={Rao, Roshan M and Liu, Jason and Verkuil, Robert and Meier, Joshua and Canny, John and Abbeel, Pieter and Sercu, Tom and Rives, Alexander},
  booktitle={International conference on machine learning},
  pages={8844--8856},
  year={2021},
  organization={PMLR}
}

@article{vespag,
    author = {Marquet, Céline and Schlensok, Julius and Abakarova, Marina and Rost, Burkhard and Laine, Elodie},
    title = {Expert-guided protein language models enable accurate and blazingly fast fitness prediction},
    journal = {Bioinformatics},
    volume = {40},
    number = {11},
    pages = {btae621},
    year = {2024},
    month = {11},
    issn = {1367-4811},
    doi = {10.1093/bioinformatics/btae621},
    url = {https://doi.org/10.1093/bioinformatics/btae621},
    eprint = {https://academic.oup.com/bioinformatics/article-pdf/40/11/btae621/60811415/btae621.pdf},
}

@article{prillo2024ultrafast,
  title={Ultrafast classical phylogenetic method beats large protein language models on variant effect prediction},
  author={Prillo, Sebastian and Wu, Wilson and Song, Yun S},
  journal={Advances in neural information processing systems},
  volume={37},
  pages={130265--130290},
  year={2024}
}

@inproceedings{tan2019efficientnet,
  title={Efficientnet: Rethinking model scaling for convolutional neural networks},
  author={Tan, Mingxing and Le, Quoc},
  booktitle={International conference on machine learning},
  pages={6105--6114},
  year={2019},
  organization={PMLR}
}

@article{erman1980hearsay,
  title={The {Hearsay-II} speech-understanding system: Integrating knowledge to resolve uncertainty},
  author={Erman, Lee D. and Hayes-Roth, Frederick and Lesser, Victor R. and Reddy, D. Raj},
  journal={ACM Computing Surveys},
  volume={12},
  number={2},
  pages={213--253},
  year={1980}
}

@article{nii1986blackboard,
  title={Blackboard systems: The blackboard model of problem solving and the evolution of blackboard architectures},
  author={Nii, H. Penny},
  journal={AI Magazine},
  volume={7},
  number={2},
  pages={38--53},
  year={1986}
}

@article{han2025blackboard,
  title={Exploring Advanced {LLM} Multi-Agent Systems Based on Blackboard Architecture},
  author={Han, Bochen and Zhang, Songmao},
  journal={arXiv preprint arXiv:2507.01701},
  year={2025}
}

@article{salemi2025blackboard,
  title={{LLM}-Based Multi-Agent Blackboard System for Information Discovery in Data Science},
  author={Salemi, Alireza and Parmar, Mihir and Goyal, Palash and Song, Yiwen and Yoon, Jinsung and Zamani, Hamed and Pfister, Tomas and Palangi, Hamid},
  journal={arXiv preprint arXiv:2510.01285},
  year={2025}
}

@article{pugachev2025codecrdt,
  title={{CodeCRDT}: Observation-Driven Coordination for Multi-Agent {LLM} Code Generation},
  author={Pugachev, Sergey},
  journal={arXiv preprint arXiv:2510.18893},
  year={2025}
}

@article{cemri2025multi,
  title={Why Do Multi-Agent {LLM} Systems Fail?},
  author={Cemri, Mert and Pan, Melissa Z. and Yang, Shuyi and Agrawal, Lakshya A. and Chopra, Bhavya and Tiwari, Rishabh and Keutzer, Kurt and Parameswaran, Aditya and Klein, Dan and Ramchandran, Kannan and Zaharia, Matei and Gonzalez, Joseph E. and Stoica, Ion},
  journal={arXiv preprint arXiv:2503.13657},
  year={2025}
}

@inproceedings{ong2025routellm,
  title={{RouteLLM}: Learning to Route {LLMs} with Preference Data},
  author={Ong, Isaac and Almahairi, Amjad and Wu, Vincent and Chiang, Wei-Lin and Wu, Tianhao and Gonzalez, Joseph E. and Kadous, M. Waleed and Stoica, Ion},
  booktitle={International Conference on Learning Representations},
  year={2025}
}

@inproceedings{liu2024dylan,
  title={A Dynamic {LLM}-Powered Agent Network for Task-Oriented Agent Collaboration},
  author={Liu, Zijun and Zhang, Yanzhe and Li, Peng and Liu, Yang and Yang, Diyi},
  booktitle={Conference on Language Modeling},
  year={2024}
}

@inproceedings{zhang2025cut,
  title={Cut the Crap: An Economical Communication Pipeline for {LLM}-based Multi-Agent Systems},
  author={Zhang, Guibin and Yue, Yanwei and Li, Zhixun and Yun, Sukwon and Wan, Guancheng and Wang, Kun and Cheng, Dawei and Yu, Jeffrey Xu and Chen, Tianlong},
  booktitle={International Conference on Learning Representations},
  year={2025}
}

@misc{openai_codex,
  author       = {{OpenAI}},
  title        = {Codex},
  year         = {2026},
  howpublished = {\url{https://github.com/openai/codex}},
  note         = {Coding agent. Accessed: 2026-08-01}
}

\appendix

\renewcommand{\thefigure}{S\arabic{figure}}
\renewcommand{\thetable}{S\arabic{table}}
\setcounter{figure}{0}
\setcounter{table}{0}

\section*{Reproducibility Statement}\label{app:reproduce}

We provide the source code of \name in \url{https://github.com/mims-harvard/AutoScientists}, which includes the launch scripts \revision{for all experiments}. We use the official GitHub repositories for BioML-Bench (\url{https://github.com/science-machine/biomlbench}), ProteinGym (\url{https://github.com/OATML-Markslab/ProteinGym}), Autoresearch (\url{https://github.com/karpathy/autoresearch}), Biomni (\url{https://github.com/snap-stanford/biomni})\revision{,} which we use with the base LLM Claude Sonnet 4.6, and Kermut (\url{https://github.com/petergroth/kermut}). \revision{We rerun MLAgentBench, AIDE, and STELLA on BioML-Bench with the base LLM GPT-5.4 and run the Codex variant of \name with the OpenAI Codex coding agent~\cite{openai_codex}, and the repository includes the launch scripts for these runs.} Further details are available in Table~\ref{tab:existing_assets}. We use only publicly available data in benchmarking from BioML-Bench, Kaggle (\url{https://www.kaggle.com}), TDC (\url{https://tdcommons.ai}), Polaris (\url{https://polarishub.io}), Open Problems (\url{https://openproblems.bio}), and ProteinGym (\url{https://proteingym.org}). For Autoresearch and \name we launch experiments with new Claude Code sessions with the prompt ``Read the program file and start the experimental loop, do not stop until the specified wall clock time is up'', and the AI agents work with no human intervention for the duration of the specified wall clock time.

\begin{table}[p]
\centering
\scriptsize
\setlength{\tabcolsep}{3pt}
\renewcommand{\arraystretch}{1.15}
\caption{Existing assets used in this work. We report the asset type, how it was used, citation, version or commit when available, license, and URL or terms of use.}
\label{tab:existing_assets}

\begin{tabularx}{\textwidth}{L{2.2cm} L{1.7cm} Y L{1.25cm} L{1.45cm} L{1.55cm} L{0.9cm}}
\toprule
\textbf{Asset} & \textbf{Type} & \textbf{Use in this paper} & \textbf{Citation} & \textbf{Version / commit} & \textbf{License} & \textbf{URL} \\
\midrule
BioML-Bench & Benchmark / datasets / code & Evaluation on 24 biomedical ML tasks. & \cite{miller2025bioml} & \texttt{673f3d8} & MIT & \href{https://github.com/science-machine/biomlbench}{link} \\
ProteinGym & Benchmark / datasets / code & Protein engineering tasks and 217-assay benchmark. & \cite{proteingym} & PG\_v1.3 & MIT & \href{https://proteingym.org/}{link} \\
Autoresearch & Codebase / baseline & Single-agent baseline and GPT nanochat comparison. & \cite{karpathy2026autoresearch} & \texttt{228791f} & MIT & \href{https://github.com/karpathy/autoresearch}{link} \\
Biomni & Codebase / baseline & Biomedical agent baseline on BioML-Bench. & \cite{huang2025biomni} & 0.0.8 & Apache-2.0 & \href{https://github.com/snap-stanford/biomni}{link} \\
\revision{MLAgentBench} & \revision{Codebase / baseline} & \revision{Baseline agent rerun on BioML-Bench.} & \revision{\cite{huang2023mlagentbench}} & \revision{\texttt{5d71205}} & \revision{MIT} & \revision{\href{https://github.com/snap-stanford/MLAgentBench}{link}} \\
\revision{AIDE} & \revision{Codebase / baseline} & \revision{Baseline agent rerun on BioML-Bench.} & \revision{\cite{jiang2025aide}} & \revision{\texttt{5d66a21}} & \revision{MIT} & \revision{\href{https://github.com/WecoAI/aideml}{link}} \\
\revision{STELLA} & \revision{Codebase / baseline} & \revision{Baseline agent rerun on BioML-Bench.} & \revision{\cite{jin2025stella}} & \revision{\texttt{1a8386e}} & \revision{Apache-2.0} & \revision{\href{https://github.com/zaixizhang/STELLA}{link}} \\
Kermut & Codebase / baseline model & Seed method for \name-Kermut. & \cite{kermut} & \texttt{7e9e2e6} & MIT & \href{https://github.com/petergroth/kermut}{link} \\
Claude Code / Claude Sonnet 4.6 & LLM / coding agent & Base coding agent backend. & \cite{anthropic_claude_code,anthropic_claude_sonnet_46} & Claude Sonnet 4.6 & API terms & \href{https://code.claude.com/docs/en/overview}{link} \\
\revision{OpenAI Codex} & \revision{LLM / coding agent} & \revision{Alternative coding agent backend (Appendix~\ref{app:repeated_runs}).} & \revision{\cite{openai_codex}} & \revision{GPT} & \revision{API terms} & \revision{\href{https://github.com/openai/codex}{link}} \\

Kaggle UW-Madison GI tract & Dataset & Biomedical imaging benchmark task. & \cite{uw-madison-gi-tract-image-segmentation} & Jul. 2022 & Non-commercial academic research & \href{https://www.kaggle.com/competitions/uw-madison-gi-tract-image-segmentation}{link} \\
Kaggle OSIC pulmonary fibrosis & Dataset & Biomedical imaging benchmark task. & \cite{osic-pulmonary-fibrosis-progression} & Oct. 2020 & Non-commercial academic research & \href{https://www.kaggle.com/c/osic-pulmonary-fibrosis-progression}{link} \\
Kaggle histopathologic cancer detection & Dataset & Biomedical imaging benchmark task. & \cite{histopathologic-cancer-detection} & Mar. 2019 & Non-commercial academic research & \href{https://www.kaggle.com/c/histopathologic-cancer-detection}{link} \\
Kaggle RSNA-MICCAI brain tumor & Dataset & Biomedical imaging benchmark task. & \cite{rsna-miccai-brain-tumor-radiogenomic-classification} & Oct. 2021 & Non-commercial academic research & \href{https://www.kaggle.com/c/rsna-miccai-brain-tumor-radiogenomic-classification}{link} \\

TDCommons & Dataset & Drug discovery benchmark tasks. & \cite{huang2021therapeutics} & \texttt{c310c35} & MIT & \href{https://tdcommons.ai/}{link} \\
Polaris Hub tasks & Dataset & Drug discovery benchmark tasks. & \cite{polaris_hub} & 0.13.0 & Apache-2.0 & \href{https://polarishub.io/}{link} \\
Open Problems & Dataset / benchmark & Single-cell omics benchmark tasks. & \cite{luecken2025defining} & \texttt{5d53ffb} & MIT & \href{https://openproblems.bio/}{link} \\

ESM-2 models & Pretrained model & Embeddings and fine-tuning components. & \cite{lin2023evolutionary} & 650M, 3B & MIT & \href{https://github.com/facebookresearch/esm}{link} \\
ProteinMPNN & Pretrained model / features & Conditional probability and structure kernel features & \cite{dauparas2022robust} & \texttt{8907e66} & MIT & \href{https://github.com/dauparas/ProteinMPNN}{link} \\
ChemBERTa & Pretrained model & Molecular embeddings. & \cite{chithrananda2020chemberta} & Zinc-Base-V1 & MIT & \href{https://huggingface.co/seyonec/ChemBERTa-zinc-base-v1}{link} \\
Chemprop & Codebase / model & Models for BioML-Bench. & \cite{heid2023chemprop} & v2.2.3 & MIT & \href{https://github.com/chemprop/chemprop}{link} \\
RDKit & Software library & Molecular fingerprints and descriptors. & \cite{rdkit} & Q1 2026 & BSD-3-Clause & \href{https://github.com/rdkit/rdkit}{link} \\
XGBoost & Software library & Gradient-boosted tree models. & \cite{xgboost} & 3.2.0 & Apache-2.0 & \href{https://github.com/dmlc/xgboost}{link} \\
LightGBM & Software library & Gradient-boosted tree models. & \cite{ke2017lightgbm} & 4.6.0 & MIT & \href{https://github.com/lightgbm-org/LightGBM}{link} \\
CatBoost & Software library & Gradient-boosted tree models. & \cite{dorogush2018catboost} & 1.2.10 & Apache-2.0 & \href{https://github.com/catboost/catboost}{link} \\

EfficientNet & Pretrained model & Embeddings and fine-tuning components. & \cite{tan2019efficientnet} & B0, B3, B4 & Apache-2.0 & \href{https://huggingface.co/docs/transformers/model_doc/efficientnet}{link} \\

PyTorch / PyTorch Geometric & Software libraries & Neural network training and implementation. & \cite{paszke2019pytorch} & 2.11.0 & BSD-3-Clause & \href{https://pytorch.org/}{link} \\
\bottomrule
\end{tabularx}
\end{table}

\section*{Impact Statement}\label{app:impact}
\name may accelerate machine learning and AI for Science by helping researchers explore modeling choices, use experimental compute more efficiently, and document both successful and failed hypotheses through shared logs, dead-end registries, model cards, and research reports. These benefits are most relevant for computational biomedical and scientific ML settings where experimental iteration is costly. The main risks are over-trusting automatically discovered models, overfitting to benchmark feedback, and amplifying erroneous hypotheses if validation is weak. This is important in biomedical applications, where outputs should not be treated as clinically or biologically actionable without expert review and independent validation. \revision{Our experiments are restricted to computational benchmarks on publicly available datasets, and deploying \name beyond these settings would require safeguards that we did not evaluate, including sandboxed code execution with audit logs, human approval for sensitive actions, hidden test sets, and independent replication.}

\section{Implementation Details and Algorithmic Protocols}
\label{app:implementation}

This section gives the implementation behind the method in Section~\ref{sec:multi-agent-defn}. An \name run proceeds in three phases. In \emph{Launch} (App.~\ref{app:launch}), the system provisions a roster of experiment agents and analysts with no team assignments and posts a single discussion trigger to the forum $\mathcal{F}$. In \emph{Self-organized team formation} (App.~\ref{app:self_org}), all agents enter a structured discussion in which they propose research directions, vote on proposals, and write the team roster $R$, and the same protocol reopens whenever an analyst detects stagnation. In the \emph{Normal cycle}, once $R$ exists, agents are repeatedly invoked through a fixed rotation. On each invocation, an analyst generates new proposals for the team queue via coverage audit, empirical priors ranking, and diversity rules (App.~\ref{app:analyst_details}), and an experiment agent claims a queued experiment, runs it, and applies a noise-aware promotion gate to decide whether to update the champion (App.~\ref{app:noise_gate}). All coordination flows through the shared state $\mathcal{S}$, which agents access through a list--decide--read protocol (App.~\ref{app:files}) and which contains per team queues $Q_k$ governed by optimistic locking (App.~\ref{app:queue}). Algorithms~\ref{alg:heartbeat}--\ref{alg:analyst} give the per invocation pseudocode and are discussed in App.~\ref{app:heartbeat}.

\subsection{Setup and Launch}
\label{app:launch}

An \name deployment is initialized by a deterministic bootstrap procedure that performs three steps. First, the shared state $\mathcal{S}$ is initialized with the task specification and an empty experiment log. Second, the agent roster is created, comprising experiment agents (each bound to one GPU if needed) and analysts, and each agent is issued a unique credential and a role-specific heartbeat description that determines its behavior at every invocation. The agents are launched with no team assignments, so the roster $R$ that maps agents to teams is initially empty. Third, a single \texttt{[DISCUSSION-TRIGGER]} is posted to the forum $\mathcal{F}$, initiating the bootstrap discussion in which agents propose research directions and self-organize into teams (Section~\ref{sec:multi-agent-defn}, App.~\ref{app:self_org}).

After bootstrap, agents are repeatedly invoked in a fixed rotation by a simple loop. Each invocation is a single LLM session that wakes up, executes one heartbeat, and exits. Long-horizon coordination is achieved by repeating these invocations. The loop passes only the agent's identity, and the agent reads its persistent state (credentials, role assignment, and GPU binding for experiment agents) and the shared state $\mathcal{S}$ on every invocation to discover team membership, the current champion $p^*$, the queue state, and recent forum activity. Repeat discussion rounds are triggered by agents themselves when stagnation is detected (Appendix~\ref{app:heartbeat}).

\begin{algorithm}[H]
\caption{Heartbeat dispatch (per agent invocation). The discussion branch on line~3 is detailed in Algorithm~\ref{alg:selforg}, and the role-specific normal cycle on line~9 is detailed in Algorithm~\ref{alg:analyst} (analysts) or Algorithm~\ref{alg:experiment} (experiment agents).}
\label{alg:heartbeat}
\begin{algorithmic}[1]
\Require agent identity $i$
\State Read roster $R$ and recent forum $\mathcal{F}$ from $\mathcal{S}$
\If{$\mathcal{F}$ has an unresolved \texttt{[DISCUSSION-TRIGGER]} \textbf{or} $R$ is empty}
  \State \textbf{run} Algorithm~\ref{alg:selforg}; \Return \Comment{discussion}
\ElsIf{$i$ is not assigned to any team in $R$}
  \State \Return \Comment{no team exit}
\ElsIf{$i$ is an experiment agent \textbf{and} an unposted result exists}
  \State post pending \texttt{[RESULT]}; \Return \Comment{resume}
\EndIf
\State Read $\mathcal{T}_k$, $p^*$, $Q_k$, knowledge files \Comment{App.~\ref{app:files}}
\State \textbf{run} Algorithm~\ref{alg:analyst} if $i$ is an analyst, \textbf{else} Algorithm~\ref{alg:experiment} \Comment{normal cycle}
\State persist updated state and exit
\end{algorithmic}
\end{algorithm}

\begin{algorithm}[H]
\caption{Self-organized team formation (discussion branch). Detailed in App.~\ref{app:self_org}.}
\label{alg:selforg}
\begin{algorithmic}[1]
\State Read task specification, current champion $p^*$ if any, and the active \texttt{[DISCUSSION-TRIGGER]} thread from $\mathcal{F}$
\State post candidate research directions to $\mathcal{F}$
\State post ranked hypotheses about the proposed directions, with reasoning
\State critique earlier posts and identify gaps in the proposed decomposition
\State post a self-termination vote on the trigger thread, either \texttt{[DISCUSS-MORE]} or \texttt{[DISCUSS-DONE]}
\If{a majority of agents have cast \texttt{[DISCUSS-DONE]} on the trigger thread \textbf{and} $i$ is the alphabetically last analyst that participated}
  \State consolidate proposals into a roster $R = \{(\mathcal{T}_k, \mathrm{direction}_k, \mathrm{members}_k)\}_{k=1}^K$
  \State write $R$ to $\mathcal{S}$
\EndIf
\end{algorithmic}
\end{algorithm}

\begin{algorithm}[H]
\caption{Experiment agent cycle. Queue protocol in App.~\ref{app:queue} and noise-aware gate in App.~\ref{app:noise_gate}.}
\label{alg:experiment}
\begin{algorithmic}[1]
\State claim a queued experiment $q$ from $Q_k$; \Return\ if none
\State apply diff to $p^*$, train candidate $p'$, compute $\Delta = \ell(p') - \ell(p^*)$
\If{$\Delta > M\sigma$}
  \State promote $p'$ to $p^*$
\ElsIf{$0 < \Delta \leq M\sigma$}
  \State rerun on a second seed, and promote if both runs improve
\Else
  \State discard
\EndIf
\State record outcome to log $\mathcal{L}$, release claim on $Q_k$, post \texttt{[RESULT]} to $\mathcal{F}$
\end{algorithmic}
\end{algorithm}

\subsection{Heartbeat Protocol}
\label{app:heartbeat}

Algorithm~\ref{alg:heartbeat} specifies the per invocation dispatch, Algorithm~\ref{alg:selforg} specifies what an agent does in the discussion branch, and Algorithms~\ref{alg:analyst} and \ref{alg:experiment} specify the role-specific normal cycle. After reading the current roster and recent forum activity, the agent enters a discussion branch when an analyst has posted a \texttt{[DISCUSSION-TRIGGER]} in response to stagnation (Algorithm~\ref{alg:analyst}) or when the roster has not yet been formed during cold start bootstrap; a no team exit when the agent is not assigned to any team; a resume branch when an experiment agent has an unposted result from a prior session; or the role-specific normal cycle. In the normal cycle, the two roles divide labor, where an experiment agent claims a queued experiment, applies the code change, trains, gates the result against the noise floor, and records the outcome (Algorithm~\ref{alg:experiment}), while an analyst, rather than running experiments, maintains the team's knowledge by auditing untested parameters, ranking research directions by empirical effect size, and queueing new proposals for experiment agents to execute (Algorithm~\ref{alg:analyst}). In the discussion branch (Algorithm~\ref{alg:selforg}), both roles contribute research directions, hypotheses, and a self-termination vote, and the alphabetically last analyst that participated in the discussion is additionally responsible for consolidating the proposals into the new roster $R$ (App.~\ref{app:self_org}). After the branch returns, the agent persists its updated state and exits.

\begin{algorithm}[t]
\caption{Analyst cycle (when $i$ is an analyst). Each step detailed in App.~\ref{app:analyst_details}.}
\label{alg:analyst}
\begin{algorithmic}[1]
\If{recent experiments produced no improvement, or proposals have concentrated on a narrow class of changes}
  \State post \texttt{[DISCUSSION-TRIGGER]} to $\mathcal{F}$ \Comment{App.~\ref{app:self_org}}
\EndIf
\State audit untested parameters of $p^*$
\State compute research direction priors $\mu_{a,d}$ from experiment log $\mathcal{L}$
\State propose 2 experiments under ambition quota and diversity rules, and if $p^*$ changed since the analyst's last cycle, $\geq 1$ proposal must target the property responsible
\State append proposals to $Q_k$ \Comment{App.~\ref{app:queue}}
\end{algorithmic}
\end{algorithm}

\subsection{Self-Organized Team Formation}
\label{app:self_org}

The roster $R = \{(\mathcal{T}_k, \mathrm{direction}_k, \mathrm{members}_k)\}_{k=1}^K$ is not specified by an external coordinator, and agents write it themselves through a structured discussion that opens whenever the discussion branch of Algorithm~\ref{alg:heartbeat} is entered. \revision{The same protocol applies in} \emph{cold start bootstrap} at the beginning of a run, when $R$ is empty and every agent is routed to the discussion branch, and \revision{in} \emph{mid-run reformation}, when an analyst's stagnation check (Algorithm~\ref{alg:analyst}) posts a fresh \texttt{[DISCUSSION-TRIGGER]} to $\mathcal{F}$ and subsequent agents observe it.

\xhdr{Discussion contributions}
Each agent entering the discussion branch reads the task specification, the current champion $p^*$ if any, and the active \texttt{[DISCUSSION-TRIGGER]} thread. It then posts (i)~candidate research directions, (ii)~hypotheses ranking those research directions by expected effect with the reasoning that justifies the ranking, and (iii)~a self-termination vote on the trigger thread, either \texttt{[DISCUSS-MORE]} (more discussion is needed before a roster can be written) or \texttt{[DISCUSS-DONE]} (the discussion has converged). Later agents critique earlier posts, identify gaps, propose alternative \revision{groupings of research directions}, and add their own vote.

\xhdr{Termination and roster writing}
The discussion terminates when a majority of agents have cast \texttt{[DISCUSS-DONE]} on the trigger thread. The alphabetically last analyst that participated in the discussion is responsible for consolidating the proposals into the new roster $R$ and writing it to $\mathcal{S}$. The alphabetical convention is a tiebreaker that ensures exactly one agent assumes the consolidator role without further coordination, and the system has no privileged consolidator agent.

\xhdr{Reformation outcomes}
A new roster may create, merge, split, retire, or rebalance teams relative to the previous roster, depending on the discussion's conclusions. Teams whose research directions have produced no recent improvements can be retired, newly discovered productive research directions can spawn new teams, and agents can be reassigned across teams. The persistence of $R$ in $\mathcal{S}$ ensures that subsequent invocations begin from the written decomposition.

\subsection{File Discovery}
\label{app:files}

\revision{In each} heartbeat cycle, agents access the shared state $\mathcal{S}$ through a list--decide--read protocol, where the agent first retrieves only the lightweight metadata for items in $\mathcal{S}$ (path, version, timestamp, author), then selects the items relevant to its current task, and finally fetches the contents of those items. This indirection allows new artifact types to appear in $\mathcal{S}$ over the course of a run (for instance, a newly created knowledge file or dead-end registry) without changes to the access protocol.

\subsection{Queue and Claim Protocol}
\label{app:queue}

The team queue $Q_k$ is a structured record of pending experiments (each with an identifier, priority, diff description, proposing agent, and link to its proposal post) and the experiments currently claimed by agents. Concurrent reads and writes are serialized by optimistic locking on a version token, so a \revision{stale write} is rejected and the agent retries against the latest version\revision{. Each queue update is therefore atomic} and no experiment is claimed twice. Each experiment agent releases its own claim immediately after recording the experiment's outcome.

\subsection{Noise-Aware Champion Validation}
\label{app:noise_gate}

The evaluation metric $\ell$ is stochastic, since running the same program with two different random seeds yields different values. A naive rule that promotes whenever $\ell(p') > \ell(p^*)$ would therefore promote candidates whose ``improvement'' is just noise. \revision{All agents build on the shared champion $p^*$, so an erroneous promotion compounds because} every downstream comparison is then made against a corrupted baseline. We term this \emph{champion pollution} and prevent it by gating every proposed promotion against an empirically measured noise floor.

For a candidate $p'$ produced by an experiment agent (Algorithm~\ref{alg:experiment}), let $\Delta = \ell(p') - \ell(p^*)$ denote its signed improvement over the current champion. Since $\ell$ is oriented so that higher is better (Section~\ref{sec:multi-agent-defn}), $\Delta > 0$ if $p'$ is better. Let $\sigma$ denote the per run standard deviation of $\ell$ at fixed code (the \emph{noise floor}, with calibration described below), and let $M = 2$, so that $[0, M\sigma]$ is the band of improvements small enough to be confounded with noise. The gate decides
\begin{equation}
\text{promote}(p') =
\begin{cases}
\texttt{true} & \text{if } \Delta > M\sigma \\
\mathrm{confirm}(p', \mathrm{seed}_2) & \text{if } 0 < \Delta \leq M\sigma \\
\texttt{false} & \text{if } \Delta \leq 0,
\end{cases}
\end{equation}
where $\mathrm{confirm}(p', \mathrm{seed}_2)$ reruns $p'$ with a new random seed and returns \texttt{true} only if both runs improve over $p^*$. Improvements above the noise band are accepted directly, improvements inside the noise band are accepted only if a second seed rerun confirms, and non-improvements are rejected. Candidates that fall in the noise band but fail confirmation are recorded as \emph{near-misses}.

The noise floor $\sigma$ is calibrated lazily, without spending experiments specifically for calibration. Until $\sigma$ has been estimated, the gate uses a conservative default noise band so that no candidate can be promoted on weak evidence. Each time the gate's middle branch fires, the resulting pair of duplicate seed measurements $(\ell_{1,i}, \ell_{2,i})$ for the same candidate code is recorded. Once at least three such pairs have accumulated ($n \geq 3$), $\sigma$ is set to the within pair pooled standard deviation
\begin{equation}
\sigma = \sqrt{\frac{1}{2n}\sum_{i=1}^{n} (\ell_{1,i} - \ell_{2,i})^2},
\end{equation}
which estimates the per seed noise of $\ell$ at fixed code without conflating it with code-to-code variation. To prevent later pairs from retroactively reclassifying earlier promotion decisions, $\sigma$ is locked once five pairs have been recorded.

\subsection{Analyst Proposal Protocol}
\label{app:analyst_details}

On each invocation, an analyst generates two new experiment proposals for the team queue through the following cycle.

The analyst first builds a working picture of what has been tried and what has not. It scans the champion code $p^*$ for all numeric parameters (top-level constants, class fields, and inline numeric literals) and matches each against the experiment log $\mathcal{L}$ to identify parameters that have never been varied (\emph{baseline coverage audit}). It then computes the empirical mean effect size per (axis, direction) pair from prior experiments,
\begin{equation}
\mu_{a,d} = \frac{1}{|E_{a,d}|} \sum_{e \in E_{a,d}} |\Delta_e|,
\end{equation}
designates research directions with $|E_{a,d}| < 3$ as \emph{cold} and gives them an exploration bonus, deprioritizes directions whose effect size falls below the noise floor ($\mu_{a,d} < \sigma$), and \revision{sorts the team queue $Q_k$} by the resulting ranking (\emph{empirical \revision{research direction} priors}).

The analyst then drafts two new proposals subject to two filters. The \emph{ambition quota} requires at least one of the two to satisfy a bold move criterion, \revision{such as} a parameter count change of at least 10\%, a fix to a confirmed bug in the champion code, an experiment on \revision{a research direction} flagged as untested in two or more prior discussion threads, or a hypothesis tension probe whose outcome will confirm or falsify the team's current hypothesis. If neither proposal \revision{meets this criterion}, the analyst must post a public \texttt{[EXEMPT]} comment justifying why no bold candidate exists, rather than accept incrementalism. The \emph{diversity constraints} additionally require that (i)~the two proposals target different research directions, (ii)~\revision{no three or more consecutive recent proposals on the same axis push in the same direction}, and (iii)~no proposal falls inside a previously rejected range in the dead-end registry $\mathcal{D}_k$ unless the proposer explicitly states what differs from the prior failure.

When the champion has been updated since the analyst's previous cycle (a \emph{KEEP}, which is a successful promotion of a candidate to a new $p^*$), a final \emph{post KEEP} step requires at least one of the two proposals to follow up on the property responsible for the recent improvement via a different mechanism. To enforce this, the analyst must answer which property of the successful change made it work and what other untried changes share that property. The completed proposals are appended to $Q_k$.

\section{Extended Ablation Results}
\label{app:extended_ablations}

The main paper ablation table (Table~\ref{tab:ablations}) reports the three single component removals (\texttt{No analyst}, \texttt{No cross-agent}, \texttt{No self-org.}) \revision{and the \texttt{Independent agents} variant} against the full \name system. We report \revision{three} additional comparisons on the same task suite here, (i)~\name versus autoresearch \cite{karpathy2026autoresearch} as a \revision{single-agent} reference (Table~\ref{tab:ablations_singleagent} for GPT training optimization and Table~\ref{tab:biomlbench_all_tasks} for BioML-Bench), \revision{(ii)~two variants of the no self-organization ablation that change only the initial teams (Table~\ref{tab:ablations_frozen_teams}),} and \revision{(iii)}~a team size sweep with working agent counts $n\in\{2,4,14\}$ against the default crew ($n=9$) (Table~\ref{tab:ablations_teamsize}).

\begin{table}[h]
\centering
\caption{\textbf{\name versus Autoresearch} \cite{karpathy2026autoresearch} on GPT Training Optimization across two settings.}
\label{tab:ablations_singleagent}
\resizebox{\linewidth}{!}{%
\begin{tabular}{llccc>{\columncolor{modelcolor}}c}
\toprule
\textbf{Setting} & \textbf{Metric}
& \textbf{Start \texttt{val\_bpb}} & \textbf{\# Experiments}
& \textbf{Autoresearch}
& \cellcolor{modelcolor}\textbf{\name} \\
\midrule
From Autoresearch baseline & Best \texttt{val\_bpb} $(\downarrow)$ & 0.998  & 50  & 0.9790 & 0.9777 \\
From \name champion    & Best \texttt{val\_bpb} $(\downarrow)$ & 0.9777 & 100 & 0.9777 \scriptsize{(0 KEEPs)} & $\mathbf{0.9730}$ \scriptsize{(7 KEEPs)} \\
\bottomrule
\end{tabular}%
}
\end{table}

\subsection{Comparison Against Autoresearch at Matched and Extended Compute}
\label{app:singleagent_compare}
Table~\ref{tab:ablations_singleagent} reports two settings. (i)~\emph{From the Autoresearch baseline}, both systems start from the original GPT training setup released with Autoresearch~\cite{karpathy2026autoresearch} (the unmodified \texttt{nanochat} code at \texttt{val\_bpb}~$=0.998$) and run 50 experiments, and \name reaches 0.9777 whereas Autoresearch reaches 0.9790. (ii)~\emph{From the same \name champion} at \texttt{val\_bpb}~$=0.9777$ (the result of \name's setting~(i) run) for 100 additional experiments, \name accepts seven KEEPs and reaches 0.9730, while Autoresearch accepts zero KEEPs and produces no improvement over the starting champion. \name's advantage over the single-agent loop \revision{grows with the number of experiments, because the single-agent loop saturates while \name continues to find productive directions.}

\revision{
\subsection{Separating Reorganization from Initial Teams}
\label{app:frozen_teams}

The no self-organization ablation in Table~\ref{tab:ablations} differs from \name in that its teams are predefined at the start of the run and cannot be reorganized during the run. To separate these factors, we run two further variants on GPT nanochat training optimization. Both keep the same agent roster, parallel execution, shared state (champion, experiment log, and dead-end registry), and cross-agent feedback as \name, keep mid-run reorganization disabled, and use about 50 experiments, beyond the point at which \name reaches its final \texttt{val\_bpb} (by experiment $\approx$45, and $\approx$0.978 by experiment 34). In the first variant, agents discuss and form their own initial teams, and these teams are then kept fixed. In the second, the run is seeded with the exact teams and hypotheses that \name self-organized, and only mid-run reorganization is disabled\revision{. We} verified that these runs contain no discussion posts and that the roster stays fixed throughout.

As shown in Table~\ref{tab:ablations_frozen_teams}, self-organized initial teams improve over predefined initial teams (0.9806 vs.\ 0.9833), so self-organization at the start of the run contributes, but it does not close the gap to \name. With the initial teams copied exactly from \name, the fixed team variant still reaches only 0.9794, compared with 0.9777 for the full system. Because the initial teams, parallelism, shared state, and feedback are matched, this remaining gap is consistent with mid-run reorganization contributing beyond better initial teams.

To check that these differences exceed seed variance, we reran five configurations with independent random seeds and estimated a per seed standard deviation of $\sigma \approx 0.0007$ \texttt{val\_bpb}. The gaps between \name and each ablation in Tables~\ref{tab:ablations} and~\ref{tab:ablations_frozen_teams} exceed the corresponding $2\sigma$ band ($\approx 0.0015$), although each condition is a single run.

\begin{table}[h]
\centering
\caption{\revision{\textbf{Separating mid-run reorganization from the initial teams on GPT nanochat training optimization.} All variants keep parallel agents, the shared state, and cross-agent feedback, and the three no self-organization variants disable mid-run reorganization and differ only in how the initial teams are formed. The per seed standard deviation is $\sigma \approx 0.0007$ \texttt{val\_bpb}.}}
\label{tab:ablations_frozen_teams}
\small
\revision{%
\begin{tabular}{lc}
\toprule
\textbf{Condition} & \textbf{Best \texttt{val\_bpb}} $(\downarrow)$ \\
\midrule
No self-organization, predefined initial teams (Table~\ref{tab:ablations}) & 0.9833 \\
No self-organization, agent discussed initial teams, then fixed & 0.9806 \\
No self-organization, exact \name initial teams, then fixed & 0.9794 \\
\rowcolor{modelcolor}
\name & \textbf{0.9777} \\
\bottomrule
\end{tabular}%
}
\end{table}

}

\begin{table}[h]
\centering
\caption{\textbf{\name agent team size sweep.} \revision{Three tasks, two of them from} Table~\ref{tab:ablations}, comparing the default \name crew ($n=9$) against three crew size variants ($n=2$, $n=4$, $n=14$). $n$ is the number of \revision{experiment and analyst} agents. Bold indicates the best result per row. \revision{SPIKE-SARS2 here is the cold start BioML-Bench task, in which agents receive no prior model, and differs from the Kermut seeded setting of Sec.~\ref{sec:proteingym}. Its $n{=}9$ entry is the first of three independent runs, whose mean is $0.663 \pm 0.025$ (Appendix~\ref{app:repeated_runs}).}}
\small
\label{tab:ablations_teamsize}
\begin{tabular}{llcc>{\columncolor{modelcolor}}cc}
\toprule
& & \multicolumn{4}{c}{\textbf{\name agent team size}} \\
\cmidrule(lr){3-6}
\textbf{Task} & \textbf{Metric}
& $\textbf{n=2}$ & $\textbf{n=4}$ & $\textbf{n=9}$\,\textbf{(default)} & $\textbf{n=14}$ \\
\midrule
\multirow{2}{*}{\textbf{TDC-hERG}}
& AUROC ($\uparrow$)        & 0.780 & 0.803 & \textbf{0.867} & 0.843 \\
& Leaderboard \% ($\uparrow$) & 14.3  & 14.3  & \textbf{85.7}  & 42.9 \\
\midrule
\multirow{2}{*}{\textbf{\revision{\shortstack[l]{BioML-Bench SPIKE-SARS2\\(cold start)}}}}
& Spearman's $\rho$ ($\uparrow$) & \textbf{0.874} & 0.835          & 0.670 & 0.506 \\
& Leaderboard \% ($\uparrow$)    & \textbf{100.0} & \textbf{100.0} & 81.8  & 72.7  \\
\midrule
\textbf{GPT Training Optimization}
& Best \texttt{val\_bpb} $(\downarrow)$
& \textbf{0.9777} & 0.9778 & \textbf{0.9777} & 0.9821 \\
\bottomrule
\end{tabular}%
\end{table}

\subsection{Team Size Sensitivity, Parallelism \revision{Improvement}, and Oversubscription}
\label{app:teamsize_trajectories}

We compare four working crew sizes ($n{=}2, 4, 9, 14$) on three tasks (TDC-hERG, \revision{BioML-Bench SPIKE-SARS2 (cold start)}, GPT nanochat training optimization). Crew composition (experiment agents $+$ analysts) is $n{=}2$ ($1{+}1$), $n{=}4$ ($2{+}2$), $n{=}9$ ($6{+}3$, the default \name configuration), $n{=}14$ ($9{+}5$).

\xhdr{Crew size sensitivity is task dependent}
The three tasks in Table~\ref{tab:ablations_teamsize} show different sensitivities to crew size, and the crew size that achieves the top score differs across tasks. On TDC-hERG, AUROC spans $0.780$--$0.867$ and leaderboard percentile spans $14.3$--$85.7$ across the four crews. On \revision{BioML-Bench} SPIKE-SARS2, Spearman $\rho$ spans $0.506$--$0.874$ and leaderboard percentile spans $72.7$--$100.0$. On GPT training optimization, \texttt{val\_bpb} spans $0.9777$--$0.9821$. At $n{=}14$, every task degrades relative to its best score in the table. These results suggest that the optimal crew size \revision{varies across tasks rather than being} a fixed property of the protocol. \revision{Performance is not monotonic in the number of agents, since on SPIKE-SARS2 the two smallest crews outperform the default $n{=}9$ crew, and the largest crew is never the best. Nine agents is therefore not optimal for every task, although every crew of up to nine agents outperforms single-agent Autoresearch on SPIKE-SARS2 (0.593, Table~\ref{tab:biomlbench_all_tasks}).} Dynamically scaling team size to task difficulty is a direction for future work.

\revision{\xhdr{Choosing the number of agents in practice} The value of additional agents depends on how decomposable the task is, the coordination structure, the capability of the base model, and how many distinct research directions are useful. The current implementation adapts the organization of a fixed pool of agents but does not change the number of active agents, so we recommend selecting the agent count for each task and budget. First, identify how many experiments can run in parallel and use enough experiment agents to occupy, but not exceed, the available experimental compute. Then choose the number of analysts with a small pilot sweep, taking the smallest configuration beyond which performance or proposal diversity no longer improves. Duplicate proposals, rediscovered dead ends, idle proposal queues, and rising communication cost without new champion improvements indicate that the team is too large. Unexplored directions, or an experiment queue dominated by a narrow family of ideas, suggest adding analysts.}

\begin{figure}[h]
    \centering
    \includegraphics[width=0.85\linewidth]{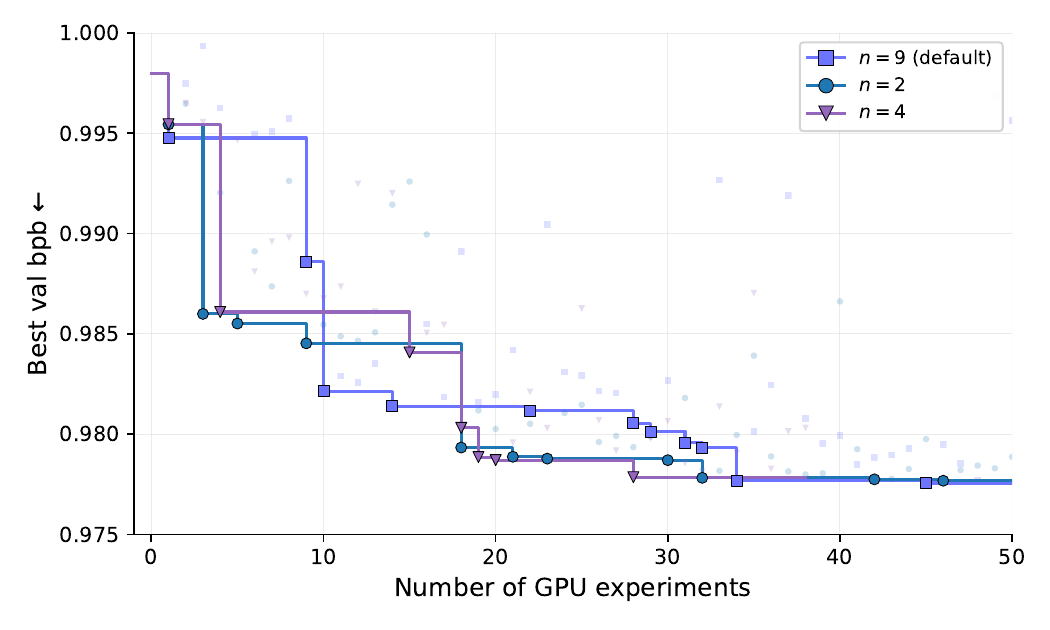}
    \caption{Team size sweep on GPT nanochat as a function of \emph{experiment number}. Solid step curves are the running best-so-far validation \texttt{val\_bpb}, opaque markers indicate accepted KEEPs, and transparent markers show per experiment attempts that did not improve the running min. Final values are $n{=}2$ 0.9777 (11 KEEPs / 51 exps), $n{=}4$ 0.9778 (7 KEEPs / 38 exps), default \name ($n{=}9$) 0.9777 (11 KEEPs / 71 exps), $n{=}14$ 0.9821 (15 KEEPs / 50 exps).}
    \label{fig:teamsize_exp}
\end{figure}

\begin{figure}[h]
    \centering
    \includegraphics[width=0.85\linewidth]{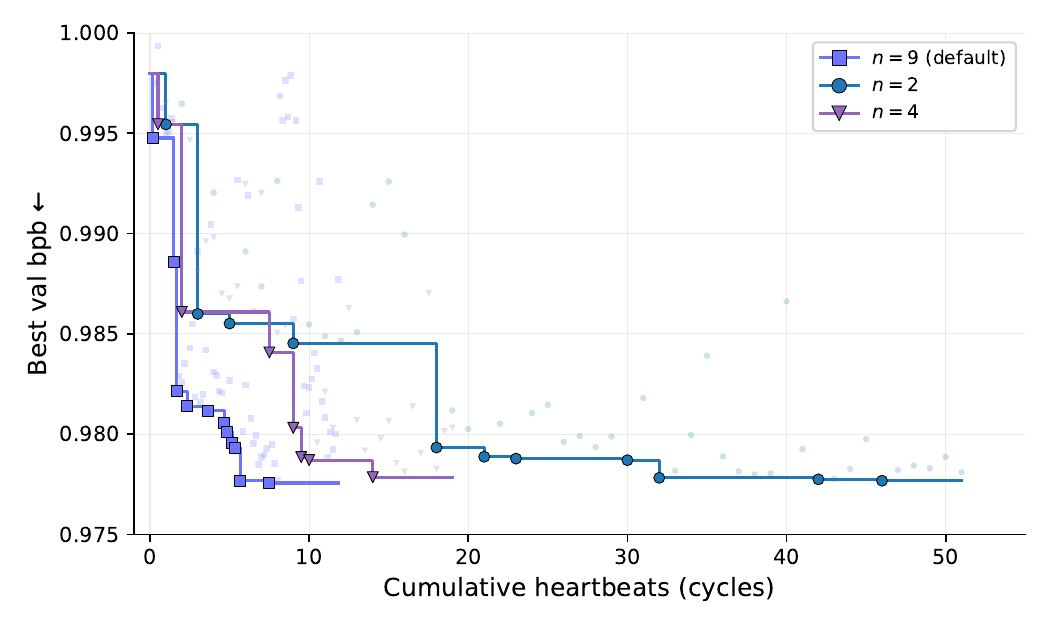}
    \caption{Same trajectories as Fig.~\ref{fig:teamsize_exp}, plotted against \emph{cumulative heartbeats} under fully parallel execution (one heartbeat = one synchronized cycle in which all $n$ working agents fire one LLM call simultaneously). $n{=}2$ finishes in $\approx$26 heartbeats, $n{=}4$ in $\approx$10, default $n{=}9$ in $\approx$8, and $n{=}14$ in $\approx$4. Larger crews finish faster under parallel execution.}
    \label{fig:teamsize_hb}
\end{figure}

\xhdr{Parallel execution of more agents is faster at similar quality}
\name agents are designed to run in parallel, and within each heartbeat all $n$ working agents fire one LLM call simultaneously. Due to resource limits in our setup, we executed the agents iteratively rather than fully in parallel, so we report two complementary views. The per experiment axis (Fig.~\ref{fig:teamsize_exp}) compares final quality under matched experimental compute, and the three smaller crews converge to similar final \texttt{val\_bpb} ($\sim 0.9777$, within the noise floor) over 51, 38, and 71 experiments for $n{=}2, 4, 9$ respectively. The per heartbeat axis (Fig.~\ref{fig:teamsize_hb}) projects the wall-clock cost of the same runs under fully parallel execution, with $\approx 26$, $\approx 10$, and $\approx 8$ heartbeats for $n{=}2, 4, 9$, a $\sim 3.25\times$ speedup of $n{=}9$ over $n{=}2$. \revision{Per experiment quality is therefore similar across crews in the productive range, while} parallel execution lets larger crews finish faster. \revision{These heartbeat counts are a projection of agent cycles under ideal parallelism, not a measurement of GPU wall-clock time.}

\revision{\xhdr{Measured wall-clock under parallel GPU execution} To measure wall-clock time under true parallel execution, we fix the agent team, base model, and GPT nanochat training setting, and vary only the number of GPUs that run experiments in parallel ($P \in \{1, 2, 4\}$). Table~\ref{tab:parallel_gpus} reports the average wall-clock time to complete one experiment after five full agent cycles. Wall-clock time per experiment decreases monotonically with more GPUs, reaching a $2.18\times$ speedup at $P{=}4$. The speedup is sublinear because each experiment involves only a few minutes of GPU training relative to the LLM reasoning and I/O of the agents, and this fixed per experiment overhead does not shrink with the number of GPUs. This overhead is a property of the implementation rather than of the coordination method, and it could be reduced with asynchronous dispatch, in which each freed GPU immediately claims the next queued experiment.

\begin{table}[h]
\centering
\caption{\revision{Measured wall-clock time per experiment on GPT nanochat when experiments run on $P$ GPUs in parallel, averaged after five full agent cycles.}}
\label{tab:parallel_gpus}
\small
\revision{%
\begin{tabular}{ccc}
\toprule
\textbf{GPUs ($P$)} & \textbf{Wall-clock per experiment} & \textbf{Speedup vs.\ $P{=}1$} \\
\midrule
1 & 9.7 min & $1.00\times$ \\
2 & 6.4 min & $1.52\times$ \\
4 & 4.4 min & $2.18\times$ \\
\bottomrule
\end{tabular}
}
\end{table}
}

\section{\name Performance Across Independent Runs}
\label{app:run_stability}

\subsection{GPT nanochat training optimization}
We executed three independent cold start runs of \name on the GPT nanochat task to assess whether \name's behavior is stable across the stochastic factors that arise during cold start coordination (workshop post acquisition order, concurrent claim races, per experiment random seeds). The three runs use identical hardware, templates, and task specification, and none uses a fixed seed or deterministic replay mechanism, so each constitutes an independent realization.

\begin{figure}[h]
    \centering
    \includegraphics[width=0.85\linewidth]{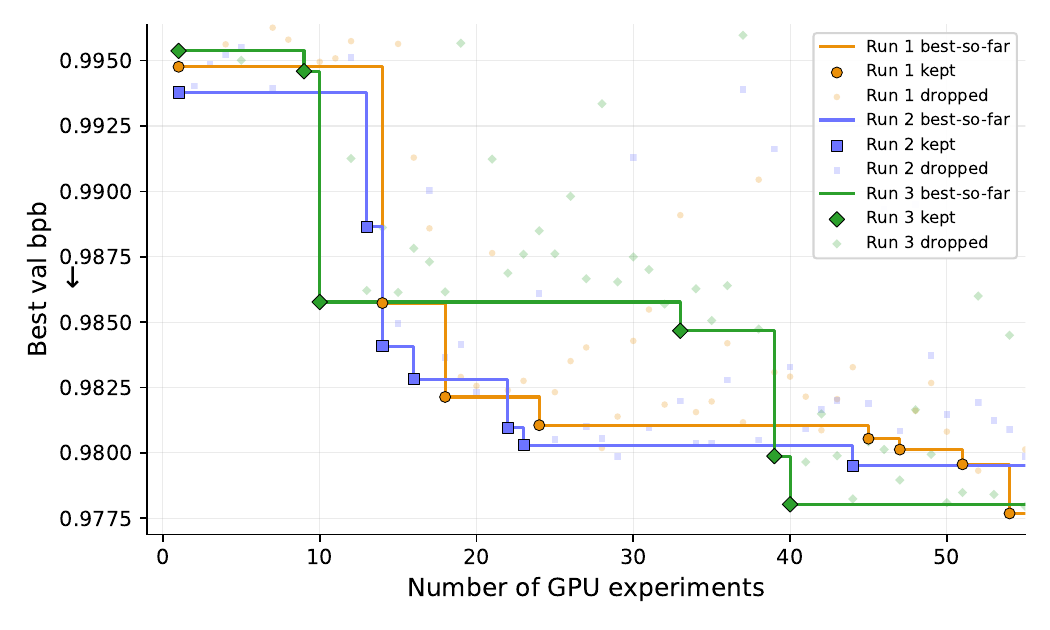}
    \caption{Per experiment trajectories of three independent cold start runs of \name on the GPT nanochat task. Solid step curves show the running best-so-far validation \texttt{val\_bpb}, opaque markers indicate accepted KEEPs, and transparent markers show per experiment attempts that did not improve the running minimum. All three runs converge to the same best-so-far region under different proposal orderings, with descent timing \revision{that depends on} when high leverage probes are discovered.}
    \label{fig:run1_run2_run3}
\end{figure}

Figure~\ref{fig:run1_run2_run3} reports the running best-so-far \texttt{val\_bpb} for all three runs. All three runs converge to the same best-so-far region, with final values of $0.9777$ (Run~1), $0.9795$ (Run~2), and $0.9780$ (Run~3) after 75, 64, and 62 experiments respectively. The mean final \texttt{val\_bpb} across runs is $0.9784$ with sample standard deviation $0.0010$ (sample variance $9.4\times 10^{-7}$, range $0.0018$). Acceptance rates are comparable across runs, at $8/75 = 10.7\%$ for Run~1, $7/64 = 10.9\%$ for Run~2, and $6/62 = 9.7\%$ for Run~3. The runs select different intermediate research directions (Newton--Schulz iteration count and short window patterns in Run~1, final LR fraction and per group learning rate allocation in Run~2, and total batch size halving and rotary base scaling in Run~3) yet reach the $0.978$--$0.980$ region, indicating that \revision{different proposal orderings reach the same productive region rather than depending on a single deterministic path.} Descent timing differs across runs because each run is gated by a small number of high leverage probes whose discovery order is stochastic\revision{.} Run~3 trails Runs~1--2 through the middle range\revision{,} recovers at experiment~37 once a coupling preserving warmdown ratio probe lands, \revision{and} then makes \revision{a large late jump} (a width up KEEP at experiment~43 worth $-0.0048$ \texttt{val\_bpb})\revision{, reaching} its final region within five experiments. The $0.0018$ gap between Runs~2 and~1 reflects a single \texttt{depth\_7} schedule probe that Run~1 reached at experiment~54 and Run~2 was halted before reaching, while Run~3 closes most of the same gap through an alternate path.

\revision{
\subsection{Independent run stability across tasks and LLM backends}
\label{app:repeated_runs}

Table~\ref{tab:repeated_runs} \revision{reports repeated runs of \name on TDC-hERG with the Claude Code and OpenAI Codex backends, on GPT nanochat, and on SPIKE-SARS2 in the cold start BioML-Bench and Kermut seeded settings. The other BioML-Bench tasks were each run once.}

\xhdr{SPIKE-SARS2 Kermut seeded} Three independent runs from the same Kermut seed reach $0.8377 \pm 0.0037$, compared with $0.8088 \pm 0.0058$ for three single-agent Autoresearch runs given the same seed, experimental compute, and number of experiments. All three \name runs converge on the same two modifications, neither of which is present in Kermut. First, they broaden the zero-shot evidence supplied to the Gaussian process, since Kermut uses a single zero-shot predictor whereas the three discovered models use fifteen, seven, and seven predictors, as kernel features, linear mean inputs, or both. Second, they replace Kermut's plain $z$-scoring of targets with a nonlinear distributional transformation (quantile-based normalization, rank-based normalization, and Yeo--Johnson, respectively). Autoresearch also discovers the broadened zero-shot features but none of the target transformations\revision{, reaching 0.8051, 0.8059, and 0.8155 in its three runs.}

\xhdr{LLM backends} To test whether \name depends on the Claude Code backend, we replace it with the OpenAI Codex coding agent~\cite{openai_codex}, keeping the rest of the system unchanged. On GPT nanochat training optimization, starting from the Autoresearch baseline ($\texttt{val\_bpb} = 0.998$), a single Codex run \revision{of 100 experiments} reaches a best \texttt{val\_bpb} of 0.9765 by experiment 19 \revision{ and makes no further improvement}, compared with 0.9777 for Claude backed \name and 0.9790 for Autoresearch within the 50-experiment budget of Table~\ref{tab:ablations_singleagent}. This run is not matched in the number of experiments to the other two, so \revision{it indicates only} that \name performs comparably with either backend. On TDC-hERG, three independent Codex runs reach a mean AUROC of $0.864 \pm 0.012$, compared with $0.853 \pm 0.020$ with Claude Sonnet 4.6 (Table~\ref{tab:repeated_runs}), and this mean is above the single runs of Biomni (0.855) and Autoresearch (0.800) (Table~\ref{tab:biomlbench_all_tasks}).

Across tasks, repeated runs identify similar high-level strategies but differ in the specific models and implementations they select, and hence in final score.

\begin{table}[h]
\centering
\caption{\revision{Repeated independent runs of \name, reported as mean $\pm$ sample standard deviation over three runs.}}
\label{tab:repeated_runs}
\small
\revision{%
\resizebox{\linewidth}{!}{%
\begin{tabular}{llcccc}
\toprule
\textbf{Task / setting} & \textbf{Metric} & \textbf{Run 1} & \textbf{Run 2} & \textbf{Run 3} & \textbf{\name} \\
\midrule
TDC-hERG (Claude Code, Sonnet 4.6) & AUROC ($\uparrow$) & 0.867 & 0.830 & 0.862 & $0.853 \pm 0.020$ \\
TDC-hERG (OpenAI Codex) & AUROC ($\uparrow$) & 0.859 & 0.855 & 0.877 & $0.864 \pm 0.012$ \\
GPT nanochat (cold start) & \texttt{val\_bpb} ($\downarrow$) & 0.9777 & 0.9795 & 0.9780 & $0.9784 \pm 0.0010$ \\
BioML-Bench SPIKE-SARS2 (cold start) & Spearman's $\rho$ ($\uparrow$) & 0.6701 & 0.6351 & 0.6845 & $0.6632 \pm 0.0254$ \\
SPIKE-SARS2 (Kermut seeded) & Spearman's $\rho$ ($\uparrow$) & 0.8407 & 0.8389 & 0.8336 & $0.8377 \pm 0.0037$ \\
\bottomrule
\end{tabular}%
}
\vspace{1mm}

}
\end{table}
}

\section{Per Experiment Trajectories for Ablation Runs}
\label{app:ablation_trajectories}

This appendix plots per experiment trajectories for the ablation runs reported in Table~\ref{tab:ablations}. All ablations start from the pristine nanochat baseline (\texttt{val\_bpb}~$=0.998$) and report the running best-so-far validation \texttt{val\_bpb}. Transparent markers show every attempted experiment, and opaque markers indicate accepted improvements (KEEPs). Autoresearch \cite{karpathy2026autoresearch} and the full \name system on the same anchor are included in each plot for reference. \revision{The ablation runs use a 50-experiment budget, and the no self-organization run is additionally extended to 71 experiments. The reference trajectories are longer (71 experiments for the full system and 83 for Autoresearch), so their final values are not a controlled terminal comparison, and their difference (0.9777 vs.\ 0.9773) is smaller than the per seed noise of \texttt{val\_bpb} ($\sigma \approx 0.0007$, Appendix~\ref{app:frozen_teams}). The controlled comparison between \name and Autoresearch at 50 experiments is in Table~\ref{tab:ablations_singleagent}. For each run we distinguish the total run length from the experiment at which the final best value is reached.}

\subsection{No Self-Organization (\texttt{abl-no-self-org})}

\begin{figure}[h]
    \centering
    \includegraphics[width=0.85\linewidth]{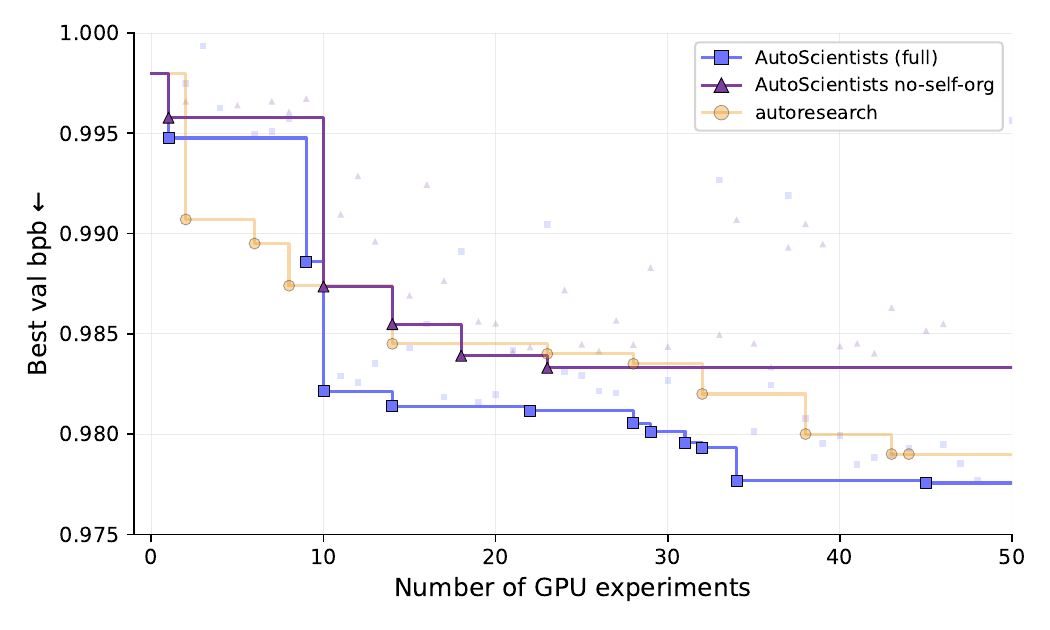}
    \caption{Per experiment trajectory for the \texttt{abl-no-self-org} ablation compared with the full \name system and single-agent autoresearch, all anchored at the pristine nanochat baseline (\texttt{val\_bpb}~$=0.998$). Solid step curves are the running best-so-far validation \texttt{val\_bpb}, opaque markers indicate accepted KEEPs, and transparent markers show per experiment attempts that did not improve the running min. Final values are autoresearch 0.9773 (15 KEEPs / 83 exps), full \name 0.9777 (11 KEEPs / 71 exps), \texttt{abl-no-self-org} 0.9833 (5 KEEPs / \revision{71} exps). \revision{The plot shows the first 50 experiments of each trajectory. The no self-organization run was extended to 71 experiments, the same length as the full system run, and made no further improvement after the last KEEP shown here.}}
    \label{fig:ablation_no_self_org}
\end{figure}

Removing self-organization (teams are fixed \revision{at the start of the run}, with no mid-run reformation) reduces the system to 5 KEEPs and a final \texttt{val\_bpb} of 0.9833 over \revision{71 experiments, with no further improvement after its last KEEP} (Fig.~\ref{fig:ablation_no_self_org}). The full \name system on the same anchor reaches 11 KEEPs and 0.9777 in 71 experiments\revision{, reaching 0.9777 by experiment $\approx$45 and $\approx$0.978 by experiment 34}. The ablated system tracks the full system for the first $\sim$10 experiments (when initial high effect size research directions are still being swept) but stalls thereafter, \revision{consistent with \texttt{ISSUES.md}, which shows that} hypothesis falsification triggers fired (analyst1 on arch v3, analyst2 on optim v3) but \revision{could not take effect without team reformation.} \revision{We extended this run from 47 to 71 experiments, the same total length as the full system run, and it made no further improvement, so the gap is not caused by a difference in run length.}

\subsection{No Cross-Agent Communication (\texttt{abl-no-cross-agent})}

\begin{figure}[h]
    \centering
    \includegraphics[width=0.85\linewidth]{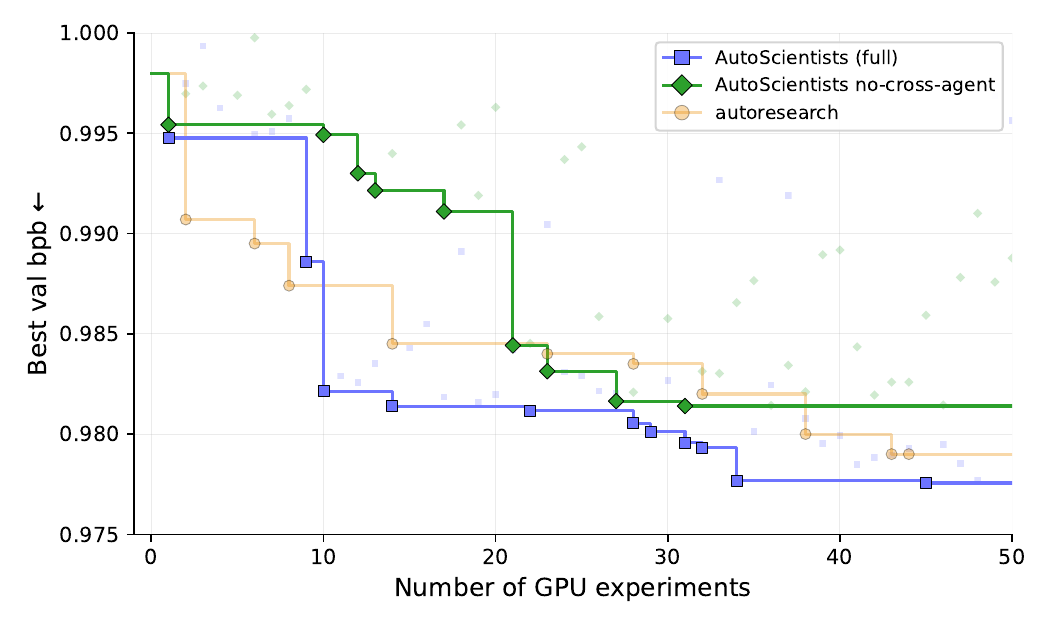}
    \caption{Per experiment trajectory for the \texttt{abl-no-cross-agent} ablation compared with the full \name system and single-agent autoresearch, all anchored at the pristine nanochat baseline (\texttt{val\_bpb}~$=0.998$). Solid step curves are the running best-so-far validation \texttt{val\_bpb}, opaque markers indicate accepted KEEPs, and transparent markers show per experiment attempts that did not improve the running min. Final values are autoresearch 0.9773 (15 KEEPs / 83 exps), full \name 0.9777 (11 KEEPs / 71 exps), \texttt{abl-no-cross-agent} 0.9814 (9 KEEPs / 50 exps). \revision{The ablation uses a 50-experiment budget, while the full system and Autoresearch curves are longer reference trajectories and are not a controlled terminal comparison.}}
    \label{fig:ablation_no_cross_agent}
\end{figure}

\revision{Removing cross-agent communication leaves coordination only through the champion record, dead-end registry, team queue, and result log. Only \texttt{[PROPOSAL]} and \texttt{[RESULT]} posts are allowed, and all other forum activity, including comments, notifications, gap analyses, rankings, synthesis posts, near-miss reports, suggestions, and structural change proposals, is disabled, as is the stagnation triggered discussion mechanism.} The resulting system reaches 9 KEEPs and a final \texttt{val\_bpb} of 0.9814 over 50 unique experiments (Fig.~\ref{fig:ablation_no_cross_agent}), against 11 KEEPs and 0.9777 in 71 experiments for the full system on the same anchor. Unlike the no self-organization ablation, the no cross-agent run continues to find improvements throughout the budget\revision{, since the shared artifacts alone carry} enough signal for \revision{decisions within individual research directions} to converge to a similar neighborhood. The cost is throughput, since the run needed roughly $1.85\times$ more experiments than a comparable full system run (\revision{measured against an internal full system run from the pristine baseline} that reached 0.980289 in 27 experiments) and never bridged the last $\sim$0.001 gap, because teams duplicated brackets that cross-team comments and team restructuring proposals would otherwise have deduplicated. The single largest individual win in this run was \texttt{sched\_batch\_half} (TOTAL\_BATCH\_SIZE $2^{19}\!\to\!2^{18}$, $\Delta=-0.0067$), which agents discovered independently after several teams converged on a step limited diagnosis from the artifact log alone.

\subsection{No Analyst Role (\texttt{abl-no-analyst})}

\begin{figure}[h]
    \centering
    \includegraphics[width=0.85\linewidth]{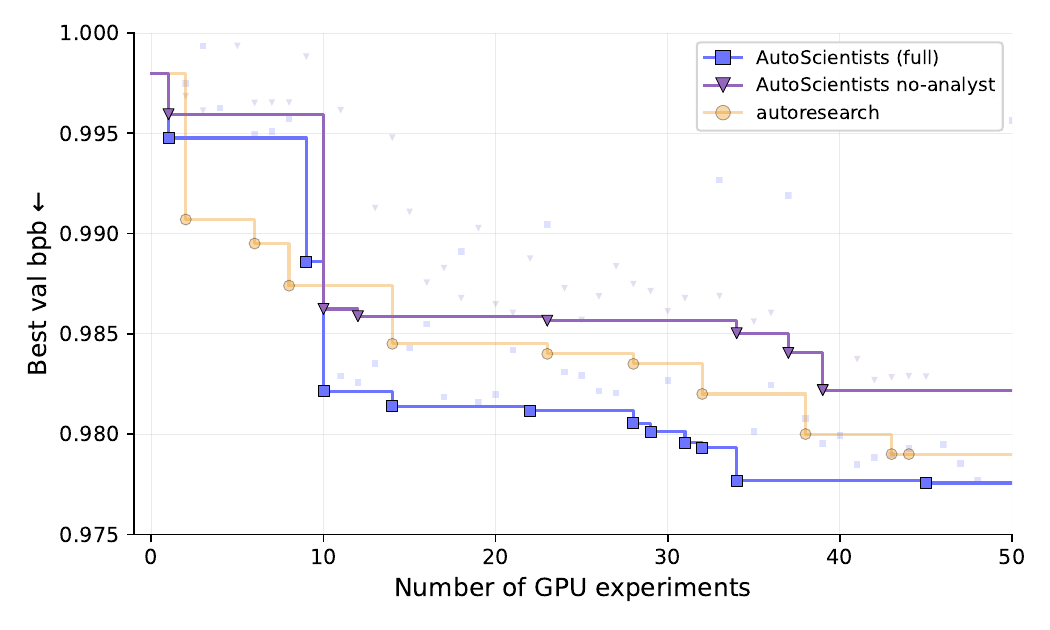}
    \caption{Per experiment trajectory for the \texttt{abl-no-analyst} ablation (purple) compared with the full \name system (red) and single-agent autoresearch (blue), all anchored at the pristine nanochat baseline (\texttt{val\_bpb}~$=0.998$). Solid step curves are the running best-so-far validation \texttt{val\_bpb}, opaque markers indicate accepted KEEPs, and transparent markers show per experiment attempts that did not improve the running min. Final values are autoresearch 0.9773 (15 KEEPs / 83 exps), full \name 0.9777 (11 KEEPs / 71 exps), \texttt{abl-no-analyst} 0.9817 (7 KEEPs / 50 exps). \revision{The ablation uses a 50-experiment budget, while the full system and Autoresearch curves are longer reference trajectories and are not a controlled terminal comparison.}}
    \label{fig:ablation_no_analyst}
\end{figure}

Removing the analyst role (no analyst agents at all, and experiment agents handle proposal generation, \revision{research direction} prioritization, and post KEEP induction in addition to executing experiments) leaves the system with five experiment agents operating against the shared workshop and artifact surface alone. The resulting system reaches 7 KEEPs and a final \texttt{val\_bpb} of 0.9817 over 50 unique experiments (Fig.~\ref{fig:ablation_no_analyst}), against 11 KEEPs and 0.9777 in 71 experiments for the full system on the same anchor. \revision{In this run,} experiment agents self-organized cross-team probes when their team's native research directions were exhausted. After the schedule team closed its TOTAL\_BATCH / WARMUP / WARMDOWN / FINAL\_LR\_FRAC bracket, gpu3 rebracketed the architecture team's MLP\_RATIO under the new compute regime (yielding KEEPs at MLP=5 and MLP=6) and gpu6 probed the optimizer team's UNEMBEDDING\_LR under the new champion (yielding a third cross-team KEEP). Three of the seven accepted improvements thus came from experiment agents probing research directions another team had closed or skipped\revision{, and} the shared workshop and the post KEEP cross-team suggestion threads carried the coordination signal that an analyst would \revision{otherwise} have written. \revision{Experiment agents can therefore take over part of the analyst's coordination role, but not all of it. Although each ablation condition is a single run, the gap to the full system (0.9817 vs.\ 0.9777) is larger than twice the per seed noise of \texttt{val\_bpb} ($2\sigma \approx 0.0015$, Appendix~\ref{app:frozen_teams}) and than the spread of three independent full system runs (0.9777--0.9795, Appendix~\ref{app:run_stability}).}

\subsection{Independent Agents (\texttt{abl-independent})}

\begin{figure}[h]
    \centering
    \includegraphics[width=0.85\linewidth]{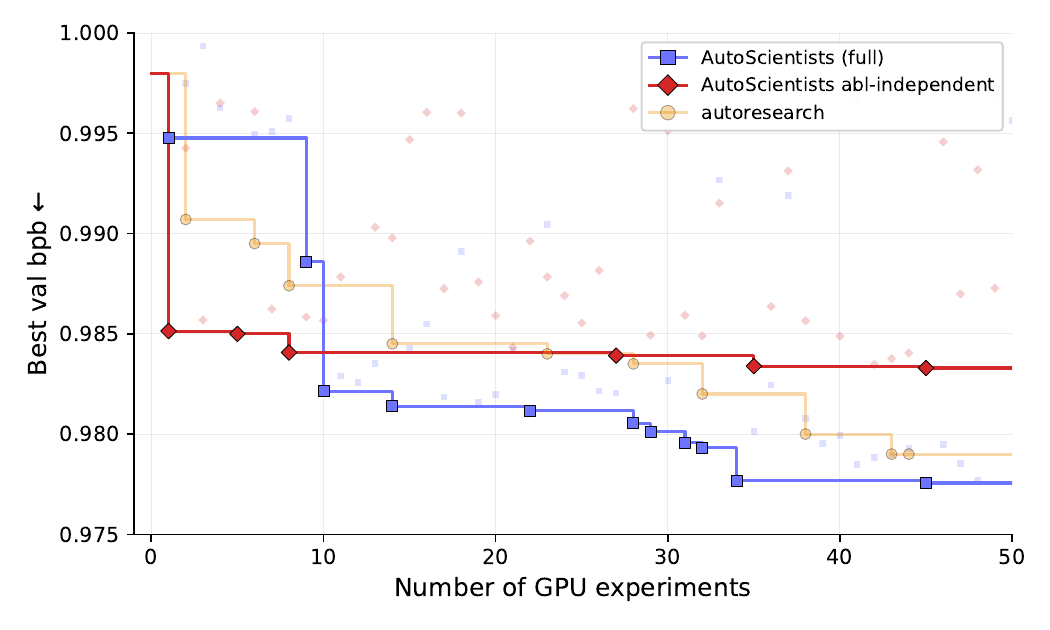}
    \caption{Per experiment trajectory for the \texttt{abl-independent} ablation compared with the full \name system and single-agent autoresearch, all anchored at the pristine nanochat baseline (\texttt{val\_bpb}~$=0.998$). Solid step curves are the running best-so-far validation \texttt{val\_bpb} (population minimum across the six solo loops for \texttt{abl-independent}), opaque markers indicate global best-so-far drops, and transparent markers show per experiment attempts that did not improve the running min. Final values are autoresearch 0.9773 (15 KEEPs / 83 exps), full \name 0.9777 (11 KEEPs / 71 exps), \texttt{abl-independent} 0.9833 (best-of-6 over 50 exps). \revision{The ablation uses a 50-experiment budget, while the full system and Autoresearch curves are longer reference trajectories and are not a controlled terminal comparison.}}
    \label{fig:ablation_independent}
\end{figure}

Removing all cross-agent coordination (no comments, no notifications, no suggestions, near-miss reports, discussion posts, gap analyses, or rankings, and no shared artifacts) and reducing the roster to six experiment agents leaves each agent running an independent autoresearch loop with its own private champion seeded from the pristine upstream and no view of any other agent's state. The resulting system reaches a best-of-population \texttt{val\_bpb} of 0.9833 over 50 unique experiments (Fig.~\ref{fig:ablation_independent}), against 11 KEEPs and 0.9777 in 71 experiments for the full system on the same anchor. Whereas \texttt{abl-no-cross-agent} forbids \revision{communication} between agents but still lets them inherit each other's wins through a shared champion program and dead-ends file, this ablation additionally hides those files. The cost of this further isolation is visible in two ways. First, five of the six agents independently rediscovered the same dominant first \revision{improvement} (the \texttt{TOTAL\_BATCH\_SIZE} $2^{19}\!\to\!2^{18}$ reduction also surfaced in \texttt{abl-no-cross-agent}) within their first two experiments, spending roughly a third of the budget on duplicates that the shared dead-ends ledger would have suppressed. Second, the agents that escaped the pristine plateau followed disjoint search paths and produced incompatible champions\revision{. Without shared artifacts, the population cannot combine these disjoint wins,} and the best-of-population trajectory plateaus roughly $0.006$ above the full system anchor.

\section{\name Output on GPT nanochat}
\label{app:autoscientists_output}

On the GPT nanochat training optimization task, \name descended from the Autoresearch baseline at \texttt{val\_bpb}~$=0.998$ to a final champion at \texttt{val\_bpb}~$=0.97769$ over 75 experiments. Alongside the champion training script, the run produces two structured documents that follow the templates released with the system, a \emph{model card} (Sec.~\ref{app:model_card_app}) covering model details, training procedure, evaluation, and compute, and a \emph{research insights} document (Sec.~\ref{app:research_insights_app}) recording the reasoning trajectory, including which research directions were tried, which were accepted, which were rejected, and what the analyst notes recorded as the mechanism in each case. The per experiment trajectory is the Run~1 trace in Figure~\ref{fig:run1_run2_run3}.

\subsection{Model Card}\label{app:model_card_app}
This card is populated from the champion record. Fields that are not informative at this scale (carbon emission accounting, demo links, downstream use guidance) are omitted. An analogous card for a BioML-Bench task is shown in Figure~\ref{fig:model-card} of the main paper.

\paragraph{Model details.}
\begin{itemize}
\item \textbf{Developed by:} \name multi-agent system (6 experiment and 3 analyst agents), with Claude Sonnet as the per agent backend.
\item \textbf{Model type:} decoder-only transformer language model, 87\,M parameters.
\item \textbf{Language:} English (FineWeb shard).
\item \textbf{Base model:} pristine nanochat baseline released with Autoresearch~\cite{karpathy2026autoresearch} (\texttt{val\_bpb}~$=0.998$).
\end{itemize}

\paragraph{Uses.}
\begin{itemize}
\item \textbf{Direct use:} a research artifact for short budget LM training optimization. The model is below the scale at which broad linguistic competence emerges and is not intended for production language modeling.
\item \textbf{Out-of-scope use:} downstream language tasks, bias sensitive applications, and any deployment requiring distributional coverage beyond the FineWeb training shard.
\end{itemize}

\paragraph{Bias, risks, and limitations.}
The model is trained on \revision{FineWeb} and inherits any biases of that source. The 300\,s training budget produces an underfit model relative to publication scale LMs, so \texttt{val\_bpb} should not be interpreted as a language task performance score.

\paragraph{Training data.}
A FineWeb shard distributed with the Autoresearch reference, tokenized by the upstream BPE-8192 tokenizer. The training and evaluation data specifications are inherited verbatim from the task definition shared by all agents.

\paragraph{Training procedure.}
\begin{itemize}
\item \textbf{Architecture knobs:} \texttt{DEPTH}~$=7$, \texttt{ASPECT\_RATIO}~$=96$ (\texttt{model\_dim}~$=768$), \texttt{HEAD\_DIM}~$=128$ (6 heads), \texttt{WINDOW\_PATTERN}~$=$\,\texttt{SSSL} with short window ratio $1/8$, RoPE, ResFormer alternating value embeddings.
\item \textbf{Optimizer:} hybrid Muon (NorMuon variant, Polar-Express normalization at constant $1.0$ with $4$ Newton--Schulz iterations) for 2D matrix parameters, and AdamW for scalars, embeddings, and gates. Per group learning rates are \texttt{EMBEDDING\_LR}~$=0.6$, \texttt{UNEMBEDDING\_LR}~$=0.004$, \texttt{MATRIX\_LR}~$=0.04$, \texttt{SCALAR\_LR}~$=0.5$. \texttt{WEIGHT\_DECAY}~$=0.2$.
\item \textbf{Schedule:} \texttt{TOTAL\_BATCH\_SIZE}~$=2^{18}$, \texttt{DEVICE\_BATCH\_SIZE}~$=128$, \texttt{WARMUP\_RATIO}~$=0$, \texttt{WARMDOWN\_RATIO}~$=0.5$, \texttt{FINAL\_LR\_FRAC}~$=0$.
\item \textbf{Training regime:} bf16 mixed precision, and \texttt{torch.compile} with full graph capture.
\end{itemize}

\paragraph{Speeds, sizes, times.}
$300$\,s wall-clock training (excluding compilation warmup), $1{,}293$ optimizer steps, $339$\,M tokens processed, $43.5\%$ model FLOPs utilization, and $58$\,GB peak VRAM.

\paragraph{Evaluation.}
\begin{itemize}
\item \textbf{Test data:} pinned held-out FineWeb shard \revision{(about 21\,M evaluation tokens)}, excluded from training.
\item \textbf{Metric:} validation bits-per-byte, $\texttt{val\_bpb} = \text{total\_nats} / (\log 2 \cdot \text{total\_bytes})$, with byte length weighting per token.
\item \textbf{Result:} $\texttt{val\_bpb}~=~0.97769$, reproducible to within the empirical noise floor ($\sigma=1.0\!\times\!10^{-3}$, $n=7$ cross-seed pairs).
\end{itemize}

\paragraph{Compute infrastructure.}
Single NVIDIA H100 80\,GB GPU. The training script is self-contained, and rerunning it requires only the upstream Autoresearch repository plus the script.

\subsection{Research Insights}\label{app:research_insights_app}
The companion document recording the reasoning trajectory of the run.

\paragraph{Run details.}
\begin{itemize}
\item \textbf{Task:} GPT nanochat training optimization.
\item \textbf{Starting metric value:} \texttt{val\_bpb}~$=0.998$.
\item \textbf{Final metric value:} \texttt{val\_bpb}~$=0.97769$.
\item \textbf{Total experiments run:} 75.
\item \textbf{Experiments accepted (KEEPs):} 8 (1 baseline lock plus 7 improvements, $10.7\%$ acceptance).
\item \textbf{Wall-clock duration:} $\approx 6.4$ hours.
\item \textbf{Number of agents:} 9 (6 experiment, 3 analyst).
\item \textbf{Base LLM:} Claude Sonnet (per agent).
\end{itemize}

\paragraph{Findings.}
The seven accepted improvements (Table~\ref{tab:v9_keep_chain}) span three distinct mechanism categories, \emph{throughput} (more optimizer steps within the 300\,s budget), \emph{capacity} (more parameters per token), and \emph{optimizer quality} (better gradient signal per step), rather than concentrating on a single \revision{research direction}. The dominant first \revision{improvement} is a throughput effect, where halving the per step batch from $2^{19}$ to $2^{18}$ approximately doubles the number of optimizer steps at near constant total tokens and alone accounts for $-0.0090$ of the total $-0.017$ descent. The largest single capacity step bundles a learning rate correction, since widening the model from $d{=}512$ to $d{=}768$ raises parameter count from 50\,M to 94\,M and simultaneously brings the $d_\text{model}$-LR-scale factor from $0.78$ to $1.0$. The remaining five wins refine the throughput / quality balance, including $\texttt{ns\_steps}~5\!\to\!4$ in Polar-Express Muon, two halvings of the short attention window ratio, removing the $1.02\!\times$ Polar-Express normalization margin, and a final trade of $8\%$ parameters for $14.4\%$ optimizer steps via depth reduction $8\!\to\!7$. Each refinement is \revision{confirmed on multiple seeds} against the noise floor.

\begin{table}[h]
\centering
\caption{Chain of accepted improvements (KEEPs) produced by \name on the GPT nanochat task. Each row records the proposing experiment agent, its team, the modified \revision{research direction} and change applied, the resulting validation \texttt{val\_bpb}, the $\Delta$ from the contemporaneous champion, and the mechanism category recorded by the analyst on the proposing team. Row~0 is the run local baseline lock, and rows~1--7 form the productive descent to the champion at \texttt{val\_bpb}~$=0.977687$. Mechanism categories are \emph{T} = throughput (more optimizer steps within the budget), \emph{C} = capacity (more parameters per token), \emph{Q} = optimizer quality (better gradient signal per step).}
\label{tab:v9_keep_chain}
\resizebox{\linewidth}{!}{%
\begin{tabular}{cllllrrcl}
\toprule
\textbf{\#} & \textbf{Agent} & \textbf{Team} & \textbf{\revision{Research direction}} & \textbf{Change} & \textbf{\texttt{val\_bpb}} & \textbf{$\Delta$} & \textbf{Cat.} & \textbf{Mechanism (analyst recorded)} \\
\midrule
0 & Experiment Agent 3 & schedule    & ---                          & upstream baseline lock                                  & 0.99476 & ---        & ---       & run local anchor for noise-floor calibration \\
1 & Experiment Agent 1 & throughput  & \texttt{TOTAL\_BATCH\_SIZE}  & $2^{19}\!\to\!2^{18}$                                   & 0.98573 & $-0.00903$ & T         & $\sim\!2\times$ optimizer steps at near constant total tokens \\
2 & Experiment Agent 5 & hidden      & \texttt{ASPECT\_RATIO}       & $64\!\to\!96$ ($d_\text{model}\!:\!512\!\to\!768$)      & 0.98214 & $-0.00359$ & C         & 50M$\to$94M params, $d_\text{model}$ LR scale to 1.0 (removes 22\% LR overcorrection) \\
3 & Experiment Agent 6 & hidden      & \texttt{ns\_steps}           & $5\!\to\!4$                                             & 0.98106 & $-0.00108$ & Q         & 4 Newton--Schulz iterations sufficient for Polar-Express orthogonalization \\
4 & Experiment Agent 6 & hidden      & \texttt{short\_window\_ratio}& $1/2\!\to\!1/4$                                         & 0.98054 & $-0.00052$ & T         & shorter S-layer window $\to$ $+2.8\%$ optimizer steps \\
5 & Experiment Agent 6 & hidden      & \texttt{short\_window\_ratio}& $1/4\!\to\!1/8$                                         & 0.98013 & $-0.00042$ & T         & \revision{continued bracket of the research direction}, $+2.2\%$ steps with second seed confirmation \\
6 & Experiment Agent 2 & throughput  & \texttt{polar\_express\_norm}& $1.02\!\to\!1.0$                                        & 0.97956 & $-0.00056$ & Q         & exact normalization in Polar-Express Muon (removes numerical safety margin) \\
7 & Experiment Agent 4 & schedule    & \texttt{DEPTH}               & $8\!\to\!7$ (at \texttt{ASPECT\_RATIO}~$=96$)           & 0.97769 & $-0.00188$ & T         & $-8\%$ params (94M$\to$87M) for $+14.4\%$ optimizer steps (1130$\to$1293) \\
\bottomrule
\end{tabular}%
}
\end{table}

\paragraph{Generalizable insights.}
\begin{enumerate}
\item Under a fixed wall-clock training budget, optimizer step granularity dominates over total token volume on this task, since the same total tokens processed in $\sim\!2\times$ as many steps yields a $\sim\!1\%$ \texttt{val\_bpb} reduction. Counterevidence would be a fixed budget benchmark in which step doubling at constant total tokens produces no improvement.
\item Capacity \revision{improvements} in this regime are gated by accompanying LR scale corrections, since widening the model alone improved \texttt{val\_bpb} by $-0.0036$, with the analyst notes attributing most of that to the $d_\text{model}$-LR scale becoming exact rather than to raw parameter count. Counterevidence would be a width up KEEP whose internal LR scale factor was already exact.
\item Schedule shape is load-bearing, since pure shape rewrites (e.g.\ replacing the linear warmdown with a constant) regress, whereas magnitude changes within the existing functional form are productive.
\end{enumerate}

\paragraph{Task-specific findings.}
\begin{enumerate}
\item Polar-Express Muon's $1.02\!\times$ numerical stability margin is unnecessary at this scale, since removing it improves \texttt{val\_bpb} by $-0.000564$ at $4$ Newton--Schulz iterations. Whether this transfers to other Muon implementations or other model scales is not tested by this run.
\item The \texttt{SSSL} window pattern with short window ratio $1/8$ is dominant over both pure short (\texttt{SSSS}) and shifted (\texttt{SSLL}) replacements, and the analyst notes interpret this as preserving a local / long-range attention balance the alternatives break.
\end{enumerate}

\paragraph{Dead ends and negative results.}
The run rejects $67$ experiments, \revision{one of which was} flagged by the multi seed gate as a near-miss (\texttt{short\_window\_ratio}~$=1/16$, where the two seeds disagreed). Table~\ref{tab:v9_dead_ends} reports ten representative directions spanning the search space and reveals two different kinds of dead end. The first sets limits on the productive region, \revision{as} $|\Delta|>5\sigma$ regressions on \texttt{rotary\_base}~$=1$k, \texttt{DEPTH}~$=6$, \texttt{HEAD\_DIM}~$=64$, \texttt{ASPECT\_RATIO}~$=112$, and \texttt{ns\_steps}~$=3$ constrain the champion to a narrow ridge in five orthogonal directions. The second characterizes the ridge rather than bounding it, since both \texttt{mlp\_ratio} directions (3 and 5) are worse than the champion's $4$ by similar amounts, all four \texttt{WARMDOWN\_RATIO} probes ($0.25, 0.30, 0.65, 0.70$) regress relative to the champion's $0.5$, and replacing the \texttt{SSSL} window pattern with either \texttt{SSSS} or \texttt{SSLL} loses the local / long-range balance.

\begin{table}[h]
\centering
\caption{Representative dead ends from the \name run on GPT nanochat. Of 75 logged experiments, 67 were rejected. Ten representative research directions drawn from across the search space are shown here, and for each, the row reports the direction tested, the value, the resulting \texttt{val\_bpb}, $\Delta$ from the contemporaneous champion (positive by definition for a DISCARD), and the mechanism the analyst recorded for why the change failed.}
\label{tab:v9_dead_ends}
\resizebox{\linewidth}{!}{%
\begin{tabular}{lllrrl}
\toprule
\textbf{Axis} & \textbf{Direction} & \textbf{Value} & \textbf{\texttt{val\_bpb}} & \textbf{$\Delta$} & \textbf{Reason} \\
\midrule
\texttt{DEPTH}              & decrease & 6 (\texttt{ASPECT\_RATIO}~$=64$) & 1.0226 & $+0.0278$ & capacity collapse from 50M$\to$26M params ($-48\%$) outweighs throughput \\
\texttt{HEAD\_DIM}          & decrease & 64                                & 0.9904 & $+0.0094$ & lower quality attention scores dominate over throughput \revision{improvement} \\
\texttt{ASPECT\_RATIO}      & increase & 112                               & 0.9877 & $+0.0100$ & per step compute outpaces capacity \revision{improvement} at fixed budget \\
\texttt{TOTAL\_BATCH\_SIZE} & decrease & $2^{17}$                          & 0.9913 & $+0.0056$ & gradient variance penalty exceeds step count \revision{improvement} \\
\texttt{WINDOW\_PATTERN}    & replace  & SSSS, SSLL                        & 0.9808, 0.9829 & $+0.0031$, $+0.0019$ & SSSL is load-bearing, and pure short / shifted patterns lose long-range signal \\
\texttt{rotary\_base}       & both     & 1k, 100k                          & 0.9876, 0.9826 & $+0.0055$, $+0.0004$ & frequency band mismatch with 2k-token training context \\
\texttt{ns\_steps}          & decrease & 3                                 & 0.9891 & $+0.0080$ & insufficient orthogonalization quality (opposite direction from KEEP at 4) \\
\texttt{WARMDOWN\_RATIO}    & both     & 0.25, 0.30, 0.65, 0.70            & 0.9856--0.9979 & $+0.0010$ to $+0.0099$ & champion 0.5 sits at an optimum, and both shorter and longer warmdown waste gradient signal \\
\texttt{softcap}            & decrease & 30, 10                            & 0.9994, 0.9951 & $+0.0046$, $+0.0003$ & logit softcap clips informative high confidence logits \\
\texttt{mlp\_ratio}         & both     & 3, 5                              & 0.9801, 0.9801 & $+0.0006$, $+0.0024$ & narrow local optimum at champion's value of 4 \\
\bottomrule
\end{tabular}%
}
\end{table}

\paragraph{Coordination and team dynamics.}
The cold start discussion produced three teams (architecture, schedule, throughput) over $12$ substantive forum posts and a $6/6$ DONE / MORE vote split, with no predefined \revision{research directions}. The accepted improvements distribute across all three teams (4 in architecture, 2 in throughput, and 2 in schedule). Cross-team transfer is observable in the chain, where the short attention window \revision{research direction} was bracketed by the architecture team, the Polar-Express normalization refinement was proposed by the throughput team on the resulting wider model champion, and the final depth reduction was proposed by the schedule team on top of the throughput team's polar norm KEEP. The multi seed gate prevented one near-miss promotion (\texttt{short\_window\_ratio}~$=1/16$), where the two seeds disagreed.

\paragraph{Limitations of these insights.}
The findings come from a single run of the system on a single task with a single agent backend at the $300$\,s budget. The throughput dominance claim is best supported by the $-0.009$ \revision{effect of the first improvement}, while the smaller refinement effects ($\le 0.002$) are individually multi seed confirmed but their compounding behavior under different orderings is untested, since each refinement was applied to the contemporaneous champion rather than to the pristine baseline. Several research directions were not probed, including pretraining objective changes, weight initialization beyond the embedding init scales tested, and data mixture ratios.

\section{Implementation Details of BioML-Bench}\label{app:biomlbench}

\subsection{Setup}
\xhdr{Experiment compute resources} In the BioML-Bench protocol drug discovery, protein engineering, and single cell omics tasks are run on CPU-only machines with an 8-hour limit. In contrast, we run \name and Autoresearch under a unified experimental compute setting of 4 hours on 1 H100 GPU with 16 CPUs and 48\,GB memory for all tasks except biomedical imaging\revision{,} which had a wall-clock budget of 16 hours. \revision{So that every agent receives the same experimental compute, we also rerun MLAgentBench~\cite{huang2023mlagentbench}, AIDE~\cite{jiang2025aide}, STELLA~\cite{jin2025stella}, and Biomni~\cite{huang2025biomni} on the 20 drug discovery, protein engineering, and single-cell omics tasks under this setting (4 hours on 1 H100 GPU with 16 CPUs and 48\,GB memory). For these four agents on the four biomedical imaging tasks, we report the published BioML-Bench results, which were obtained with a 16-hour GPU budget. The Reference baseline is a fixed random estimator.} \revision{We adopt this unified setting, which uses a shorter wall-clock budget,} to better reflect contemporary biomedical ML practice, where GPU-backed environments are commonly available even for non-imaging workloads. Additionally, the agents did not use web search or fetch in any of their experiments. The results are available in Table~\ref{tab:biomlbench-full}.

\xhdr{Validation and test evaluation} For each task, validation sets are defined using domain-specific splitting strategies from the BioML-Bench training set, including scaffold-based holdout splits for drug discovery, patient-level held-out folds for biomedical imaging, and batch or site stratified holdouts for single-cell omics. Final benchmark performance is reported on a held-out BioML-Bench test \revision{set for biomedical imaging, drug discovery, and single-cell omics, whereas protein engineering tasks are scored with the benchmark's cross-validation protocol}.

\subsection{Task Datasets and Evaluation Metrics}

\subsubsection{Biomedical Imaging}

\xhdr{kaggle-histopathologic-cancer-detection}
PatchCamelyon (PCam) contains $96\!\times\!96$ RGB pathology image patches with
a binary metastatic cancer label defined by the center $32\!\times\!32$
region.  Training uses $174{,}464$ labeled patches and the test set \revision{has} $45{,}561$ patches.  The validation set is a random split of the training set.  The metric is ROC-AUC (higher is better).

\xhdr{kaggle-osic-pulmonary-fibrosis-progression}
A clinical time-series task with $158$ idiopathic pulmonary fibrosis
patients in the training set (each with multi week FVC measurements
and a baseline CT scan). There are
$18$ test patients with baseline visit only and CT scans.  The test grader scores
$1{,}908$ \texttt{Patient\_Week} predictions covering weeks $-12$ to
$133$ for the $18$ test patients. A
patient-level cross-validation split is used for model development.
The metric is the modified Laplace log-likelihood
$-\sqrt{2}\,\Delta/\sigma_c - \log(\sqrt{2}\,\sigma_c)$ with
$\Delta=\min(|FVC_{\mathrm{true}}-FVC_{\mathrm{pred}}|,1000)$ and
$\sigma_c=\max(\sigma,70)$, averaged over all test (patient, week) pairs
(higher is better).

\xhdr{kaggle-rsna-miccai-brain-tumor-radiogenomic-classification}
Multi-parametric brain MRI for the RSNA-MICCAI BraTS challenge, where each
patient has four DICOM sequences (FLAIR, T1w, T1wCE, T2w).  Training uses the $526$ patients
with binary MGMT promoter methylation labels and the test set is $59$ patients.  Validation uses a patient-level
 cross-validation split.  The metric
is ROC-AUC of the predicted MGMT methylation probability (higher is
better).

\xhdr{kaggle-uw-madison-gi-tract-image-segmentation}
MR-Linac guided MRI for radiation therapy planning with $50$ cancer patients
imaged across $1$--$5$ treatment days, with stomach / small bowel /
large bowel run-length-encoded segmentation masks.  Provided are
$77{,}328$ slice class rows in the training set and a $17{,}760$-row validation set with held-out patients and held-out treatment dates. The test set contains
$20{,}400$ entries ($6{,}800$ slices $\times$ $3$
classes). The metric is
$0.4\,\mathrm{Dice} + 0.6\,(1-\mathrm{HD}_{3D})$, where Dice is per
(slice, class) (with $0$ for the empty pred / empty truth case) and
$\mathrm{HD}_{3D}$ is the symmetric Hausdorff distance of the
class-OR-merged 3D case day volume, normalized by the unit cube diagonal
$\sqrt{3}$ to lie in $[0,1]$ (higher overall score is better).

\subsubsection{Drug Discovery}

All drug discovery tasks share the same supervised structure with a public
training set and a held-out, label blinded test set, both downloaded
from the Polaris Hub. During development the training set is split into
five Murcko scaffold groups and 5-fold scaffold
cross-validation is used as $\ell_\text{eval}$.

\xhdr{tdcommons-bbb-martins}
Binary classification of blood--brain-barrier (BBB) permeability for
$1{,}624$ training molecules and $406$ test molecules. The metric is
ROC-AUC (higher is better).

\xhdr{tdcommons-caco2-wang}
Regression of Caco-2 cell permeability ($\log P_{\text{app}}$) for
$728$ training molecules and $182$ test molecules. The metric is mean
absolute error (MAE, lower is better).

\xhdr{tdcommons-cyp2d6-substrate-carbonmangels}
Binary classification of CYP2D6 substrate activity for $532$ training
molecules and $135$ test molecules. The metric is PR-AUC (higher
is better) to remain meaningful under imbalance.

\xhdr{tdcommons-herg}
Binary classification of hERG potassium channel inhibition for $523$
training molecules and $132$ test molecules. The metric is ROC-AUC
(higher is better).

\xhdr{tdcommons-lipophilicity-astrazeneca}
Regression of octanol/water lipophilicity ($\log D_{7.4}$) for $3{,}360$
training molecules and $840$ test molecules from the AstraZeneca DMPK
release. The metric is MAE (lower is better).

\xhdr{polaris-adme-fang-hclint-1}
Regression of human liver microsome intrinsic clearance
(\texttt{LOG\_HLM\_CLint}, mL/min/kg) on the
\texttt{adme-fang-hclint-1} benchmark with $2{,}229$
training molecules and $575$ test molecules.  The
metric is Pearson correlation $r$ between predicted and measured log
clearance (higher is better).

\xhdr{polaris-adme-fang-hppb-1}
Regression of human plasma-protein binding (\texttt{LOG\_HPPB},
\% unbound) on the \texttt{adme-fang-hppb-1}. There are $126$ training molecules and $34$ test molecules. The metric is Pearson $r$ (higher is better).

\xhdr{polaris-adme-fang-solu-1}
Regression of aqueous solubility (\texttt{LOG\_SOLUBILITY}) on the
\texttt{adme-fang-solu-1} benchmark with $1{,}578$
training molecules and $400$ test molecules.  The
metric is Pearson $r$ (higher is better).

\xhdr{polaris-pkis2-egfr-wt-c-1}
Binary classification of EGFR wild-type kinase inhibition
(\texttt{CLASS\_EGFR}) on the \texttt{pkis2-egfr-wt-c-1} benchmark with $496$
training compounds and $144$ test compounds and severe class
imbalance.  The metric is PR-AUC (higher is better) to remain
meaningful under imbalance.

\subsubsection{Single Cell Omics}

\xhdr{open-problems-predict-modality}
A bone marrow mononuclear cell (BMMC) CITE-seq dataset in which paired
RNA expression ($\sim$13{,}000 genes)
and surface protein abundance ($\sim$134 proteins) are measured in the
same cells. \revision{Validation uses} held-out site/donor batches. The metric is RMSE between predicted and measured protein values across all (cell, protein) pairs (lower is
better).

\xhdr{open-problems-single-cell-perturbations}
A PBMC perturbation screen with $144$ compounds and $5{,}317$ genes, in
which differential expression profiles
(\texttt{clipped\_sign\_log10\_pval} clipped to $[-4,4]$) are measured per
(compound, cell type) pair.  Training uses
the T-cell, NK-cell and regulatory T-cell rows. Validation is
a 20\% within cell type split of those training rows. Testing predicts the $151$ (compound, cell type) profiles for B cells and Myeloid cells. The metric is Mean Rowwise RMSE (MRRMSE, lower is better).

\xhdr{open-problems-cell-cell-communication-ligand-target}
This task contains a triple-negative breast cancer single-cell RNA-seq dataset accompanied by an OmniPath ligand--receptor prior.  The supervised
labels are sparse, with $81$ labeled (ligand, target cell type)
pairs and $731$ unlabeled test pairs. \revision{Validation uses} a random $80/20$ split of the $81$ labeled pairs.  The metric is the odds ratio of true positives in the top-5\% scored pairs against the held-out binary response, with the
$0.5$-shrinkage formula $\text{score}=1-1/(1+\text{OR}/2)$ for boundedness (higher is better).

\xhdr{open-problems-label-projection}
A diabetic kidney disease single nucleus RNA-seq dataset with $13$ cell
types.  The validation set is a held-out batch. The metric is the weighted
F1-score across cell types (higher is better).

\xhdr{open-problems-spatially-variable-genes}
A SlideSeqV2 mouse cortex spatial transcriptomics dataset covering $210$ genes and a smaller cerebellum dataset with labels is provided. The task is unsupervised and cerebellum labels may be used to verify that an unsupervised statistic correlates with ground truth.  The metric is Kendall's $\tau$ between predicted spatial scores and the
held-out continuous \texttt{spatial\_var\_score} for the $210$
cortex genes (higher is better).

\subsubsection{Protein Engineering}

ProteinGym datasets do not provide a held-out test set, so the out-of-fold
predictions from a prescribed 5-fold cross-validation are themselves the
official evaluation, with three split strategies
(\texttt{fold\_random\_5}, \texttt{fold\_modulo\_5},
\texttt{fold\_contiguous\_5}) for substitution scans and only
\texttt{fold\_random\_5} for indel scans.  The final score is the mean
Spearman correlation between predicted and measured fitness, computed in
raw target space (higher is better). The four substitution mutation DMS datasets are SPIKE\_SARS2\_Starr\_2020\_binding, SBI\_STAAM\_Tsuboyama\_2023\_2JVG, CBX4\_HUMAN\_Tsuboyama\_2023\_2K28, PSAE\_PICP2\_Tsuboyama\_2023\_1PSE, and the two indel datasets are Q8EG35\_SHEON\_Campbell\_2022 and CSN4\_MOUSE\_Tsuboyama\_2023\_1UFM.

\begin{table}[th]
\centering
\caption{Performance on four domains in BioML-Bench. \revision{All agents use the same experimental compute allocation. MLAgentBench, AIDE, STELLA, and Biomni are rerun under our compute setting on drug discovery, protein engineering, and single-cell omics, and on biomedical imaging, and for Reference, we report the published results from \cite{miller2025bioml}.} Detailed results are available in Table~\ref{tab:biomlbench_all_tasks}. Values are mean (SE) across tasks.}
\label{tab:biomlbench-full}
\resizebox{\linewidth}{!}{
\vspace{1mm}
\begin{tabular}{llccccc}
\toprule
\textbf{Domain} & \textbf{Agent} & \textbf{Leaderboard Percentile $(\uparrow)$} & \textbf{Mean Rank $(\downarrow)$} & \textbf{Above Median (\% $\uparrow$)} & \textbf{Any Medal (\% $\uparrow$)} & \textbf{Completion Rate (\% $\uparrow$)} \\
\midrule

\multirow{7}{*}{\shortstack[l]{Biomedical Imaging\\$(n=4)$}}
& Reference        & $3.30\,(2.52)$  & --   & $0.0\,(0.0)$   & $0.0\,(0.0)$   & NA    \\
& MLAgentBench$^\circ$ & $21.73\,(11.41)$ & \revision{4.00} & $12.5\,(12.5)$ & $0.0\,(0.0)$   & 100.0 \\
& AIDE$^\circ$         & $12.11\,(9.43)$  & \revision{4.25} & $6.2\,(6.2)$   & $0.0\,(0.0)$   & 81.2  \\
& STELLA$^\circ$      & $5.91\,(4.98)$   & \revision{4.25} & $6.2\,(6.2)$   & $0.0\,(0.0)$   & 68.8  \\
& Biomni$^\circ$     & $19.04\,(10.83)$ & \revision{4.75} & $12.5\,(12.5)$ & $12.5\,(12.5)$ & 100.0 \\
& Autoresearch$^\circ$ & $39.60\,(21.75)$ & \revision{2.25} & $25.0\,(25.0)$ & $25.0\,(25.0)$ & 100.0 \\
\rowcolor{modelcolor}
& \name $^\circ$ & $\mathbf{45.75}\,(\mathbf{22.18})$ & \revision{$\mathbf{1.50}$} & $\mathbf{50.0}\,(\mathbf{28.9})$ & $\mathbf{50.0}\,(\mathbf{28.9})$ & 100.0 \\

\cmidrule(lr){1-7}

\multirow{7}{*}{\shortstack[l]{Drug Discovery\\$(n=9)$}}
& Reference        & $1.11\,(1.11)$   & -- & $0.0\,(0.0)$   & $0.0\,(0.0)$   & NA    \\
& \revision{MLAgentBench$^\dagger$} & \revision{$37.01\,(9.06)$} & \revision{4.33} & \revision{$22.2\,(14.7)$} & \revision{$22.2\,(14.7)$} & \revision{100.0} \\
& \revision{AIDE$^\dagger$} & \revision{$42.23\,(11.43)$} & \revision{2.78} & \revision{$44.4\,(17.6)$} & \revision{$33.3\,(16.7)$} & \revision{100.0} \\
& \revision{STELLA$^\dagger$} & \revision{$16.12\,(4.94)$} & \revision{5.44} & \revision{$0.0\,(0.0)$} & \revision{$0.0\,(0.0)$} & \revision{100.0} \\
& Biomni$^\dagger$ & $47.91\,(10.77)$ & \revision{3.11} & $44.4\,(17.6)$ & $44.4\,(17.6)$ & 100.0 \\
& Autoresearch$^\dagger$ & $46.16\,(10.59)$ & \revision{2.89} & $33.3\,(16.7)$ & $33.3\,(16.7)$ & 100.0\\
\rowcolor{modelcolor}
& \name $^\dagger$ & $\mathbf{64.52}\,(\mathbf{8.37})$ & \revision{$\mathbf{2.44}$} & $\mathbf{55.6}\,(\mathbf{17.6})$ & $\mathbf{55.6}\,(\mathbf{17.6})$ & 100.0 \\

\cmidrule(lr){1-7}

\multirow{7}{*}{\shortstack[l]{Protein Engineering\\$(n=6)$}}
& Reference        & $0.00\,(0.00)$   & -- & $0.0\,(0.0)$   & $0.0\,(0.0)$   & NA    \\
& \revision{MLAgentBench$^\dagger$} & \revision{$46.97\,(20.11)$} & \revision{5.33} & \revision{$50.0\,(22.4)$} & \revision{$50.0\,(22.4)$} & \revision{100.0} \\
& \revision{AIDE$^\dagger$} & \revision{$71.21\,(11.39)$} & \revision{4.00} & \revision{$66.7\,(21.1)$} & \revision{$66.7\,(21.1)$} & \revision{100.0} \\
& \revision{STELLA$^\dagger$} & \revision{$48.48\,(19.59)$} & \revision{5.67} & \revision{$50.0\,(22.4)$} & \revision{$50.0\,(22.4)$} & \revision{100.0} \\
& Biomni$^\dagger$     & $93.94\,(3.83)$ & 2.50 & $\mathbf{100.0}\,(\mathbf{0.0})$ & $\mathbf{100.0}\,(\mathbf{0.0})$ & 100.0 \\
& Autoresearch$^\dagger$ & $\mathbf{96.97}\,(\mathbf{3.03})$ & 2.00 & $\mathbf{100.0}\,(\mathbf{0.0})$ & $\mathbf{100.0}\,(\mathbf{0.0})$ & 100.0 \\
\rowcolor{modelcolor}
& \name $^\dagger$ & $\mathbf{96.97}\,(\mathbf{3.03})$ & $\mathbf{1.50}$ & $\mathbf{100.0}\,(\mathbf{0.0})$ & $\mathbf{100.0}\,(\mathbf{0.0})$ & 100.0 \\

\cmidrule(lr){1-7}

\multirow{7}{*}{\shortstack[l]{Single Cell Omics\\$(n=5)$}}
& Reference        & $7.34\,(4.52)$   & -- & $0.0\,(0.0)$   & $0.0\,(0.0)$   & NA    \\
& \revision{MLAgentBench$^\dagger$} & \revision{$79.00\,(10.74)$} & \revision{4.20} & \revision{$80.0\,(20.0)$} & \revision{$\mathbf{80.0}\,(\mathbf{20.0})$} & \revision{100.0} \\
& \revision{AIDE$^\dagger$} & \revision{$58.33\,(15.58)$} & \revision{4.40} & \revision{$60.0\,(24.5)$} & \revision{$60.0\,(24.5)$} & \revision{100.0} \\
& \revision{STELLA$^\dagger$} & \revision{$74.00\,(16.14)$} & \revision{5.20} & \revision{$80.0\,(20.0)$} & \revision{$60.0\,(24.5)$} & \revision{100.0} \\
& Biomni$^\dagger$ & $78.00\,(10.20)$ & \revision{3.60} & $80.0\,(20.0)$ & $\mathbf{80.0}\,(\mathbf{20.0})$ & 100.0 \\
& Autoresearch $^\dagger$ & $86.00\,(9.80)$ & \revision{2.00} & $\mathbf{100.0}\,(\mathbf{0.0})$ & $\mathbf{80.0}\,(\mathbf{20.0})$ & 100.0 \\
\rowcolor{modelcolor}
& \name $^\dagger$ & $\mathbf{88.00}\,(\mathbf{9.70})$ & $\mathbf{1.60}$ & $\mathbf{100.0}\,(\mathbf{0.0})$ & $\mathbf{80.0}\,(\mathbf{20.0})$& 100.0 \\

\cmidrule(lr){1-7}

\multirow{7}{*}{\shortstack[l]{\revision{Overall}\\\revision{$(n=24)$}}}
& \revision{Reference}        & \revision{$2.50\,(1.16)$} & -- & \revision{$0.0\,(0.0)$} & \revision{$0.0\,(0.0)$} & \revision{NA} \\
& \revision{MLAgentBench}     & \revision{$45.70\,(7.43)$} & \revision{4.50} & \revision{$39.6\,(10.0)$} & \revision{$37.5\,(10.1)$} & \revision{100.0} \\
& \revision{AIDE}             & \revision{$47.81\,(7.18)$} & \revision{3.67} & \revision{$46.9\,(10.2)$} & \revision{$41.7\,(10.3)$} & \revision{96.9} \\
& \revision{STELLA}           & \revision{$34.57\,(7.86)$} & \revision{5.25} & \revision{$30.2\,(9.4)$} & \revision{$25.0\,(9.0)$} & \revision{94.8} \\
& \revision{Biomni}           & \revision{$60.87\,(7.23)$} & \revision{3.33} & \revision{$60.4\,(10.0)$} & \revision{$60.4\,(10.0)$} & \revision{100.0} \\
& \revision{Autoresearch}     & \revision{$66.07\,(7.38)$} & \revision{2.38} & \revision{$62.5\,(10.1)$} & \revision{$58.3\,(10.3)$} & \revision{100.0} \\
\rowcolor{modelcolor}
& \revision{\name}            & \revision{$\mathbf{74.40}\,(\mathbf{6.20})$} & \revision{$\mathbf{1.88}$} & \revision{$\mathbf{75.0}\,(\mathbf{9.0})$} & \revision{$\mathbf{70.8}\,(\mathbf{9.5})$} & \revision{100.0} \\

\bottomrule
\end{tabular}
}
\begin{minipage}{\linewidth}
\vspace{1mm}
\footnotesize
$^\circ$$^\dagger$ Wall-clock time and compute access of agents are 4h GPU \& CPU for $\dagger$ and 16h GPU \& CPU for $\circ$. \revision{All rows use the same compute protocol within each domain, and imaging results for MLAgentBench, AIDE, STELLA, and Biomni are the published values, obtained with a 16-hour GPU budget.}
\end{minipage}
\end{table}

\begin{table*}[t]
\centering
\scriptsize
\caption{Per task comparison of all six agents on BioML-Bench. For each method we report the task score, leaderboard percentile (LB\%), whether the score is above the median leaderboard entry (Med), and medal status. \revision{LB\% is the percentage of leaderboard entries with a lower score and Med compares the score with the leaderboard median, so scores with LB\% $=50.0$ can fall on either side of the median.} The best score across the six agents is shown in \textbf{bold}. \revision{Imaging entries for MLAgentBench, AIDE, STELLA, and Biomni are the published BioML-Bench results over four runs, where LB\% counts failed runs as 0, Score averages completed runs (-- if none completed), and a decimal in Medal is the fraction of the four runs.}}
\label{tab:biomlbench_all_tasks}
\resizebox{\linewidth}{!}{
\begin{tabular}{llcccccccc>{\columncolor{modelcolor}}c>{\columncolor{modelcolor}}c>{\columncolor{modelcolor}}c>{\columncolor{modelcolor}}c}
\toprule
& & \multicolumn{4}{c}{\textbf{Biomni}} & \multicolumn{4}{c}{\textbf{Autoresearch}} & \multicolumn{4}{c}{\cellcolor{modelcolor}\textbf{\name}} \\
\cmidrule(lr){3-6} \cmidrule(lr){7-10} \cmidrule(lr){11-14}
\textbf{Task} & \textbf{Metric}
& \textbf{Score} & \textbf{LB\%} & \textbf{Med} & \textbf{Medal}
& \textbf{Score} & \textbf{LB\%} & \textbf{Med} & \textbf{Medal}
& \textbf{Score} & \textbf{LB\%} & \textbf{Med} & \textbf{Medal} \\
\midrule

\multicolumn{14}{l}{\textbf{Biomedical imaging}} \\
kaggle-histopathologic-cancer-detection & ROC-AUC
& 0.81818 & 20.8 & N & --
& 0.99832 & 99.3 & Y & Gold
& \textbf{0.99834} & 99.3 & Y & Gold \\
kaggle-osic-pulmonary-fibrosis-progression & Laplace LL$\uparrow$
& -9.42451 & 5.0 & N & --
& -7.51872 & 9.8 & N & --
& \textbf{-7.11904} & 13.9 & N & -- \\
kaggle-rsna-miccai-brain-tumor-radiogenomic-classification & ROC-AUC
& 0.51338 & 49.0 & 0.50 & 0.50
& 0.52353 & 44.3 & N & --
& 0.54353 & 64.8 & Y & Bronze \\
kaggle-uw-madison-gi-tract-image-segmentation & Dice+HD$\uparrow$
& 0.20356 & 1.4 & N & --
& \textbf{0.56327} & 5.0 & N & --
& 0.55114 & 5.0 & N & -- \\
\midrule
\multicolumn{14}{l}{\textbf{Drug discovery}} \\
tdcommons-bbb-martins & AUROC
& 0.91014 & 37.5 & N & --
& 0.90908 & 25.0 & N & --
& \textbf{0.92030} & 75.0 & Y & Bronze \\
tdcommons-caco2-wang & MAE$\downarrow$
& \textbf{0.27560} & 100.0 & Y & Gold
& 0.32257 & 25.0 & N & --
& 0.27663 & 100.0 & Y & Gold \\
tdcommons-cyp2d6-substrate-carbonmangels & AUPRC
& 0.63100 & 22.2 & N & --
& 0.67413 & 22.2 & N & --
& 0.61870 & 22.2 & N & -- \\
tdcommons-herg & AUROC
& 0.85464 & 71.4 & Y & Silver
& 0.79985 & 14.3 & N & --
& \textbf{0.86672} & 85.7 & Y & Silver \\
tdcommons-lipophilicity-astrazeneca & MAE$\downarrow$
& 0.54186 & 0.0 & N & --
& \textbf{0.40003} & 88.9 & Y & Gold
& 0.42191 & 77.8 & Y & Bronze \\
polaris-adme-fang-hclint-1 & Pearson $r$
& 0.71530 & 60.0 & Y & Bronze
& \textbf{0.73308} & 90.0 & Y & Gold
& 0.69122 & 50.0 & N & -- \\
polaris-adme-fang-hppb-1 & Pearson $r$
& 0.84818 & 80.0 & Y & Silver
& 0.83560 & 80.0 & Y & Bronze
& \textbf{0.87292} & 80.0 & Y & Silver \\
polaris-adme-fang-solu-1 & Pearson $r$
& 0.64745 & 20.0 & N & --
& 0.65073 & 20.0 & N & --
& \textbf{0.65778} & 50.0 & N & -- \\
polaris-pkis2-egfr-wt-c-1 & PR-AUC
& 0.74681 & 40.0 & N & --
& 0.77942 & 50.0 & N & --
& 0.76616 & 40.0 & N & -- \\
\midrule
\multicolumn{14}{l}{\textbf{Single Cell Omics}} \\
open-problems-predict-modality & RMSE$\downarrow$
& \textbf{0.59122} & 100.0 & Y & Gold
& 0.64356 & 100.0 & Y & Gold
& 0.69722 & 100.0 & Y & Gold \\
open-problems-single-cell-perturbations & MRRMSE$\downarrow$
& 0.78246 & 80.0 & Y & Silver
& 0.78163 & 80.0 & Y & Silver
& \textbf{0.77241} & 90.0 & Y & Gold \\
open-problems-cell-cell-communication-ligand-target & Odds ratio$\uparrow$
& 0.68281 & 100.0 & Y & Gold
& 0.70835 & 100.0 & Y & Gold
& \textbf{0.92367} & 100.0 & Y & Gold \\
open-problems-label-projection & F1-weighted
& 0.95437 & 60.0 & Y & Bronze
& \textbf{0.97484} & 100.0 & Y & Gold
& 0.96394 & 100.0 & Y & Gold \\
open-problems-spatially-variable-genes & Kendall $\tau$
& 0.64992 & 50.0 & N & --
& 0.68923 & 50.0 & Y & --
& \textbf{0.69631} & 50.0 & Y & -- \\
\midrule
\multicolumn{14}{l}{\textbf{Protein engineering}} \\
proteingym-dms-SPIKE\_SARS2\_Starr\_2020\_binding & Spearman$\uparrow$
& 0.59660 & 81.8 & Y & Silver
& 0.59294 & 81.8 & Y & Silver
& \textbf{0.67011} & 81.8 & Y & Silver \\
proteingym-dms-SBI\_STAAM\_Tsuboyama\_2023\_2JVG & Spearman$\uparrow$
& 0.80046 & 81.8 & Y & Silver
& 0.81837 & 100.0 & Y & Gold
& \textbf{0.83384} & 100.0 & Y & Gold \\
proteingym-dms-CBX4\_HUMAN\_Tsuboyama\_2023\_2K28 & Spearman$\uparrow$
& 0.94496 & 100.0 & Y & Gold
& 0.95137 & 100.0 & Y & Gold
& \textbf{0.95676} & 100.0 & Y & Gold \\
proteingym-dms-PSAE\_PICP2\_Tsuboyama\_2023\_1PSE & Spearman$\uparrow$
& 0.93362 & 100.0 & Y & Gold
& 0.90897 & 100.0 & Y & Gold
& \textbf{0.97606} & 100.0 & Y & Gold \\
proteingym-dms-Q8EG35\_SHEON\_Campbell\_2022\_indels & Spearman$\uparrow$
& 0.80133 & 100.0 & Y & Gold
& \textbf{0.82683} & 100.0 & Y & Gold
& 0.81119 & 100.0 & Y & Gold \\
proteingym-dms-CSN4\_MOUSE\_Tsuboyama\_2023\_1UFM\_indels & Spearman$\uparrow$
& 0.93493 & 100.0 & Y & Gold
& \textbf{0.93965} & 100.0 & Y & Gold
& 0.93186 & 100.0 & Y & Gold \\
\bottomrule
\end{tabular}
}
\vspace{1.5mm}
\resizebox{\linewidth}{!}{
\begin{tabular}{llcccccccccccc}
\toprule
& & \multicolumn{4}{c}{\revision{\textbf{MLAgentBench}}} & \multicolumn{4}{c}{\revision{\textbf{AIDE}}} & \multicolumn{4}{c}{\revision{\textbf{STELLA}}} \\
\cmidrule(lr){3-6} \cmidrule(lr){7-10} \cmidrule(lr){11-14}
\textbf{Task} & \textbf{Metric}
& \textbf{Score} & \textbf{LB\%} & \textbf{Med} & \textbf{Medal}
& \textbf{Score} & \textbf{LB\%} & \textbf{Med} & \textbf{Medal}
& \textbf{Score} & \textbf{LB\%} & \textbf{Med} & \textbf{Medal} \\
\midrule

\multicolumn{14}{l}{\textbf{Biomedical imaging}} \\
kaggle-histopathologic-cancer-detection & ROC-AUC
& \revision{0.87243} & \revision{26.4} & \revision{N} & \revision{--}
& \revision{0.93320} & \revision{39.7} & \revision{0.25} & \revision{--}
& \revision{--} & \revision{0.0} & \revision{N} & \revision{--} \\
kaggle-osic-pulmonary-fibrosis-progression & Laplace LL$\uparrow$
& \revision{-8.32835} & \revision{8.8} & \revision{N} & \revision{--}
& \revision{-8.02789} & \revision{8.7} & \revision{N} & \revision{--}
& \revision{-8.43104} & \revision{2.9} & \revision{N} & \revision{--} \\
kaggle-rsna-miccai-brain-tumor-radiogenomic-classification & ROC-AUC
& \revision{0.53353} & \revision{51.7} & \revision{0.50} & \revision{--}
& \revision{--} & \revision{0.0} & \revision{N} & \revision{--}
& \revision{\textbf{0.56118}} & \revision{20.7} & \revision{0.25} & \revision{--} \\
kaggle-uw-madison-gi-tract-image-segmentation & Dice+HD$\uparrow$
& \revision{0.00000} & \revision{0.0} & \revision{N} & \revision{--}
& \revision{0.00000} & \revision{0.0} & \revision{N} & \revision{--}
& \revision{--} & \revision{0.0} & \revision{N} & \revision{--} \\
\midrule
\multicolumn{14}{l}{\textbf{Drug discovery}} \\
tdcommons-bbb-martins & AUROC
& \revision{0.88575} & \revision{12.5} & \revision{N} & \revision{--}
& \revision{0.90514} & \revision{25.0} & \revision{N} & \revision{--}
& \revision{0.86773} & \revision{12.5} & \revision{N} & \revision{--} \\
tdcommons-caco2-wang & MAE$\downarrow$
& \revision{0.30738} & \revision{50.0} & \revision{N} & \revision{--}
& \revision{0.27662} & \revision{100.0} & \revision{Y} & \revision{Gold}
& \revision{0.32976} & \revision{25.0} & \revision{N} & \revision{--} \\
tdcommons-cyp2d6-substrate-carbonmangels & AUPRC
& \revision{\textbf{0.71899}} & \revision{77.8} & \revision{Y} & \revision{Silver}
& \revision{0.66415} & \revision{22.2} & \revision{N} & \revision{--}
& \revision{0.68610} & \revision{33.3} & \revision{N} & \revision{--} \\
tdcommons-herg & AUROC
& \revision{0.83196} & \revision{42.9} & \revision{N} & \revision{--}
& \revision{0.84433} & \revision{42.9} & \revision{N} & \revision{--}
& \revision{0.78954} & \revision{14.3} & \revision{N} & \revision{--} \\
tdcommons-lipophilicity-astrazeneca & MAE$\downarrow$
& \revision{0.54913} & \revision{0.0} & \revision{N} & \revision{--}
& \revision{0.54182} & \revision{0.0} & \revision{N} & \revision{--}
& \revision{0.67589} & \revision{0.0} & \revision{N} & \revision{--} \\
polaris-adme-fang-hclint-1 & Pearson $r$
& \revision{0.68945} & \revision{50.0} & \revision{N} & \revision{--}
& \revision{0.71598} & \revision{60.0} & \revision{Y} & \revision{Bronze}
& \revision{0.57202} & \revision{20.0} & \revision{N} & \revision{--} \\
polaris-adme-fang-hppb-1 & Pearson $r$
& \revision{0.81197} & \revision{60.0} & \revision{Y} & \revision{Bronze}
& \revision{0.86624} & \revision{80.0} & \revision{Y} & \revision{Silver}
& \revision{0.52120} & \revision{0.0} & \revision{N} & \revision{--} \\
polaris-adme-fang-solu-1 & Pearson $r$
& \revision{0.53018} & \revision{0.0} & \revision{N} & \revision{--}
& \revision{0.61415} & \revision{0.0} & \revision{N} & \revision{--}
& \revision{0.55827} & \revision{0.0} & \revision{N} & \revision{--} \\
polaris-pkis2-egfr-wt-c-1 & PR-AUC
& \revision{0.75470} & \revision{40.0} & \revision{N} & \revision{--}
& \revision{\textbf{0.78995}} & \revision{50.0} & \revision{Y} & \revision{--}
& \revision{0.71802} & \revision{40.0} & \revision{N} & \revision{--} \\
\midrule
\multicolumn{14}{l}{\textbf{Single Cell Omics}} \\
open-problems-predict-modality & RMSE$\downarrow$
& \revision{0.71235} & \revision{100.0} & \revision{Y} & \revision{Gold}
& \revision{0.79191} & \revision{20.0} & \revision{N} & \revision{--}
& \revision{0.71344} & \revision{100.0} & \revision{Y} & \revision{Gold} \\
open-problems-single-cell-perturbations & MRRMSE$\downarrow$
& \revision{0.78233} & \revision{80.0} & \revision{Y} & \revision{Silver}
& \revision{0.77921} & \revision{80.0} & \revision{Y} & \revision{Silver}
& \revision{0.86806} & \revision{60.0} & \revision{Y} & \revision{--} \\
open-problems-cell-cell-communication-ligand-target & Odds ratio$\uparrow$
& \revision{0.62828} & \revision{100.0} & \revision{Y} & \revision{Gold}
& \revision{0.65613} & \revision{100.0} & \revision{Y} & \revision{Gold}
& \revision{0.59923} & \revision{100.0} & \revision{Y} & \revision{Gold} \\
open-problems-label-projection & F1-weighted
& \revision{0.95610} & \revision{73.3} & \revision{Y} & \revision{Bronze}
& \revision{0.95463} & \revision{66.7} & \revision{Y} & \revision{Bronze}
& \revision{0.95825} & \revision{93.3} & \revision{Y} & \revision{Gold} \\
open-problems-spatially-variable-genes & Kendall $\tau$
& \revision{0.61977} & \revision{41.7} & \revision{N} & \revision{--}
& \revision{0.56144} & \revision{25.0} & \revision{N} & \revision{--}
& \revision{-0.00904} & \revision{16.7} & \revision{N} & \revision{--} \\
\midrule
\multicolumn{14}{l}{\textbf{Protein engineering}} \\
proteingym-dms-SPIKE\_SARS2\_Starr\_2020\_binding & Spearman$\uparrow$
& \revision{0.48816} & \revision{72.7} & \revision{Y} & \revision{Bronze}
& \revision{0.58868} & \revision{81.8} & \revision{Y} & \revision{Silver}
& \revision{0.45693} & \revision{72.7} & \revision{Y} & \revision{Bronze} \\
proteingym-dms-SBI\_STAAM\_Tsuboyama\_2023\_2JVG & Spearman$\uparrow$
& \revision{0.39911} & \revision{9.1} & \revision{N} & \revision{--}
& \revision{0.56342} & \revision{45.5} & \revision{N} & \revision{--}
& \revision{0.41457} & \revision{18.2} & \revision{N} & \revision{--} \\
proteingym-dms-CBX4\_HUMAN\_Tsuboyama\_2023\_2K28 & Spearman$\uparrow$
& \revision{0.90282} & \revision{100.0} & \revision{Y} & \revision{Gold}
& \revision{0.91489} & \revision{100.0} & \revision{Y} & \revision{Gold}
& \revision{0.86738} & \revision{100.0} & \revision{Y} & \revision{Gold} \\
proteingym-dms-PSAE\_PICP2\_Tsuboyama\_2023\_1PSE & Spearman$\uparrow$
& \revision{0.85182} & \revision{100.0} & \revision{Y} & \revision{Gold}
& \revision{0.89160} & \revision{100.0} & \revision{Y} & \revision{Gold}
& \revision{0.80951} & \revision{100.0} & \revision{Y} & \revision{Gold} \\
proteingym-dms-Q8EG35\_SHEON\_Campbell\_2022\_indels & Spearman$\uparrow$
& \revision{0.71507} & \revision{0.0} & \revision{N} & \revision{--}
& \revision{0.75969} & \revision{33.3} & \revision{N} & \revision{--}
& \revision{0.75079} & \revision{0.0} & \revision{N} & \revision{--} \\
proteingym-dms-CSN4\_MOUSE\_Tsuboyama\_2023\_1UFM\_indels & Spearman$\uparrow$
& \revision{0.75814} & \revision{0.0} & \revision{N} & \revision{--}
& \revision{0.87693} & \revision{66.7} & \revision{Y} & \revision{Bronze}
& \revision{0.40814} & \revision{0.0} & \revision{N} & \revision{--} \\
\bottomrule
\end{tabular}
}
\end{table*}

\subsection{Performance Across Wall-Clock Time}
\begin{figure}[t]
\centering
\includegraphics[width=\linewidth]{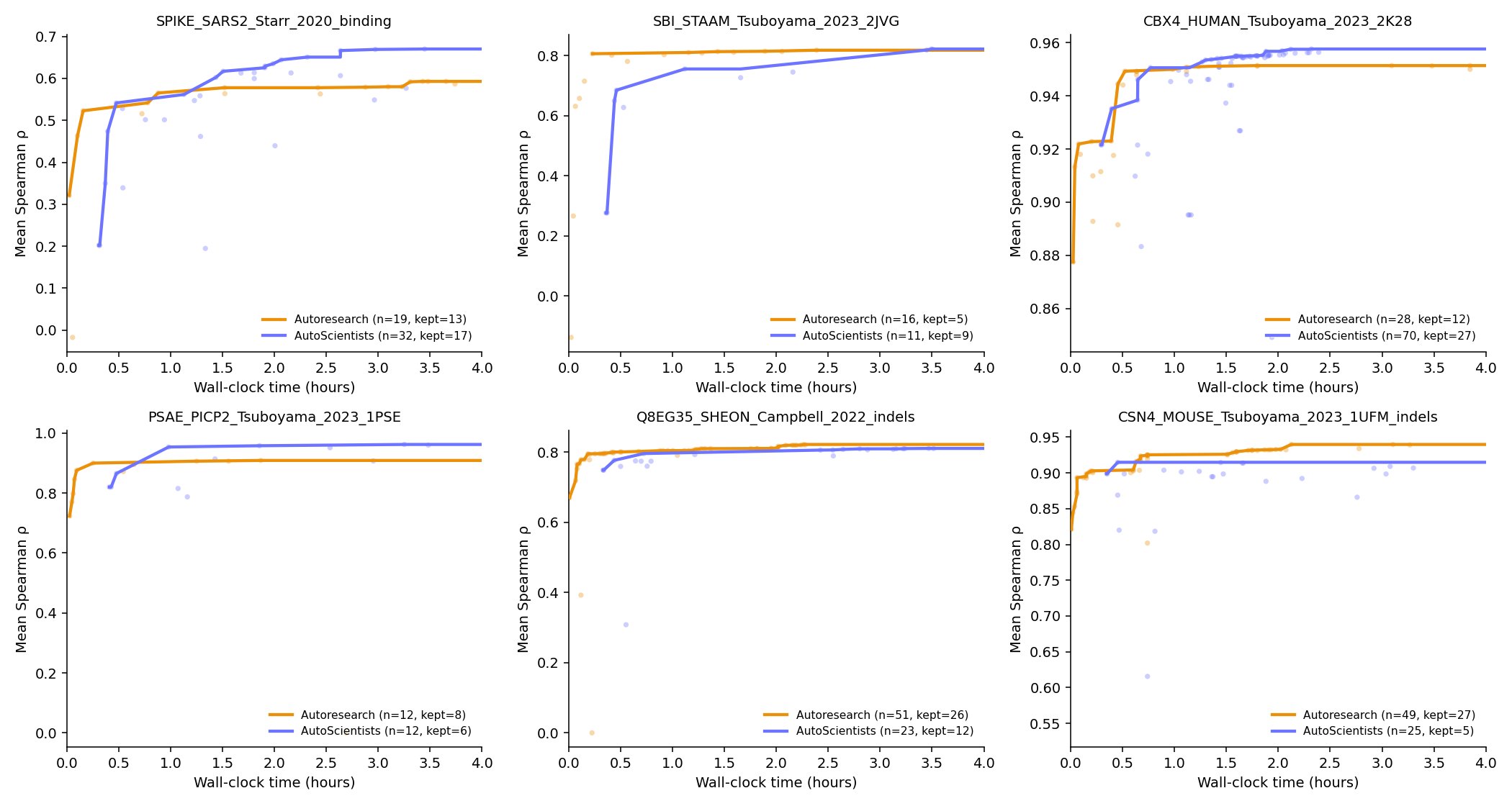}
\caption{Champion mean Spearman $\rho$ over wall-clock time for \name \revision{vs.} Autoresearch on six BioML-Bench Protein Engineering tasks. The solid line tracks the current champion, and the points show experiments that were run but did not outperform the current champion.
}
\label{fig:proteingym-efficiency}
\end{figure}

Here we plot mean Spearman $\rho$ over wall-clock time for \name \revision{vs.} Autoresearch on six BioML-Bench Protein Engineering tasks. We draw this comparison for the Protein Engineering subset of BioML-Bench because each task reports the same mean Spearman $\rho$ averaged across the same set of CV splits for $\ell_\text{eval}$ and for the final leaderboard performance. \revision{Tracking $\ell_\text{eval}$ over wall-clock time therefore captures progress on the final leaderboard. In contrast, the other BioML-Bench tasks score} a held-out test submission, which evaluates how well the agent searched the model space and how well its final approach generalizes to unseen data. For the Protein Engineering tasks, we show how quickly the champion of each approach improves \revision{within the 4h budget} (Fig.~\ref{fig:proteingym-efficiency}).

\subsection{Behavior of \name on BioML-Bench Tasks}
\label{app:biomlbench_approaches}

Here we analyze the approaches taken by \name. Each task's pipeline is
classified into seven non-exclusive method categories,
\textbf{(1) Gradient-boosted trees} (XGBoost, LightGBM, CatBoost,
ExtraTrees, RandomForest);
\textbf{(2) Foundation model frozen features} (using a pretrained
protein/molecule/image model only as a fixed feature extractor);
\textbf{(3) Foundation model fine-tuning} (LoRA on ESM-2, full
ImageNet CNN fine-tuning, end-to-end ChemBERTa);
\textbf{(4) Custom neural net trained from scratch} (MLP, ResMLP,
Chemprop D-MPNN, U-Net without pretraining);
\textbf{(5) Kernel and instance methods} (RBF-SVM, Tanimoto-SVM,
Gaussian process with Tanimoto kernel, SVR, $k$-NN);
\textbf{(6) Linear/regularized linear models} (Ridge, RidgeCV, Lasso,
LogisticRegression — usually as a meta-learner);
\textbf{(7) Hand-crafted heuristics} (no learned model).

\xhdr{Biomedical Imaging (4 tasks)}
Three of four imaging tasks center on ImageNet pretrained CNNs fine-tuned
end-to-end.
\texttt{histopathologic-cancer} fine-tunes a dual stream EfficientNet-B3~\cite{tan2019efficientnet}
with a 3072$\to$512$\to$128$\to$1 fusion head, MixUp, and 4-rotation
TTA,
\texttt{rsna-brain-tumor} fine-tunes EfficientNet-B0~\cite{tan2019efficientnet} augmented with a
slice attention pool and a two-layer Transformer encoder, and
\texttt{uw-madison-gi} fine-tunes a \texttt{segmentation\_models\_pytorch}
U-Net with an EfficientNet-B4~\cite{tan2019efficientnet} encoder for multi organ segmentation.
Foundation model fine-tuning therefore accounts for 3/4 of imaging tasks.
The fourth task, \texttt{osic-pulmonary-fibrosis}, is
distinct, since it uses pre-extracted CNN deep features as a fixed embedding
(FM-frozen), fits Ridge regressions for FVC slope and curvature
(Linear/Ridge), and retrieves nearest neighbors in PCA-50 CT
feature space (Kernel/$k$-NN).
No imaging task uses gradient boosting, a custom NN from scratch, or
a heuristic pipeline.

\xhdr{Drug Discovery (9 tasks)}
Drug discovery pipelines are dominated by gradient-boosted trees on
hand-crafted RDKit fingerprint stacks (6/9). The three tasks that omit
boosting are \texttt{lipophilicity} (ChemBERTa-2 fine-tuned
end-to-end), \texttt{hclint} (a 10-seed residual MLP ensemble on Morgan
+ MACCS + AtomPair + Mordred features), and \texttt{solu} (Chemprop v2
D-MPNN, 5-fold scaffold CV).
\revision{Chemistry-specific D-MPNNs (Chemprop) are the primary or co-primary models in two tasks, \texttt{solu} and \texttt{pkis2-egfr}, and together with the residual MLP that \texttt{hclint} uses on fingerprint features they account for the 3/9 NN-scratch tasks.}
Only \texttt{hppb} uses a foundation model in frozen feature mode. The \name approach for \texttt{hppb}
appends ChemBERTa-2 embeddings (PCA reduced to 32 dimensions) to
a 10-model fingerprint stack before a Ridge meta-learner.
\texttt{lipophilicity} is the sole task that fine-tunes a foundation
model (ChemBERTa-2 via \texttt{RobertaForMaskedLM}, CLS-token head).
Kernel methods appear in three tasks, where
\texttt{cyp2d6} adds a calibrated RBF-SVM to its CatBoost+LGB+XGB+ET
stack, \texttt{hppb} includes three SVR variants (RBF $C\in\{1,5\}$
and linear), and \texttt{pkis2-egfr} uses a Tanimoto-SVM and a
Tanimoto-GP alongside four boosting models and a Chemprop MPNN.
Linear/Ridge models appear as stack meta-learners in four tasks
(\texttt{hERG}, \texttt{cyp2d6}, \texttt{BBB}, \texttt{hppb}).

\xhdr{Single Cell Omics (5 tasks)}
Two of five tasks are hand-crafted, training-free pipelines.
\texttt{spatially-variable-genes} builds a 24-signal spatial statistics
ensemble (Moran's I, Geary's C, SPARK-X, Getis-Ord $G^*$, variogram,
and others) with weight optimization via differential evolution plus
coordinate descent, while \texttt{cell-cell-communication} scores
ligand-receptor pairs using proportion and specificity heuristics with
a fine grid search over weights.
Two tasks use linear models as their core predictor.
\texttt{single-cell-perturbations} fits per gene Ridge regressors on
four interpretable features (T cell, NK cell, Treg expression levels,
and Tanimoto drug similarity), tuned via leave-one-out CV, while
\texttt{label-projection} applies a multinomial LogisticRegression probe
to PCA reduced HVG features after batch correction.
The remaining task, \texttt{predict-modality}, trains a residual MLP
(TruncatedSVD input $\to$ 1024 $\to$ 512 $\to$ 256 with skip
connections, BatchNorm, and Dropout) from scratch for RNA-to-protein
prediction.
No single-cell task uses gradient boosting, a pretrained foundation
model, or a kernel method.

\xhdr{Protein Engineering (6 tasks)}
Five of six ProteinGym pipelines rely on frozen ESM-2 embeddings
(\texttt{SPIKE\_SARS2}, \texttt{SBI\_STAAM}, \texttt{Q8EG35},
\texttt{CBX4\_HUMAN}, \texttt{CSN4\_MOUSE}), and only \texttt{PSAE\_PICP2}
is built only from fine-tuned models.
Two tasks use ESM-2 LoRA fine-tuning as a component, where
\texttt{CSN4\_MOUSE} blends a LoRA model alongside frozen embedding
variants, while \texttt{PSAE\_PICP2} is a pure seven-way LoRA ensemble
(ESM-2-35M and ESM-2-8M).
The dominant downstream regressor is Ridge or stacked ridge variants
(3/6 tasks, \texttt{SPIKE\_SARS2}, \texttt{SBI\_STAAM}, \texttt{Q8EG35}),
followed by gradient boosting as an ensemble member (2/6 tasks,
\texttt{SPIKE\_SARS2} includes a LightGBM submodel, and
\texttt{Q8EG35} includes XGBoost and LGB submodels).
Kernel methods appear only in \texttt{Q8EG35}, whose 15-model ensemble
includes three SVR variants on ESM-2 features.
No ProteinGym pipeline trains a neural net from scratch. Every MLP in
the ensembles sits on top of precomputed ESM-2 embeddings, placing
them in the FM-frozen category.

\begin{figure}[t]
\centering
\includegraphics[width=0.95\linewidth]{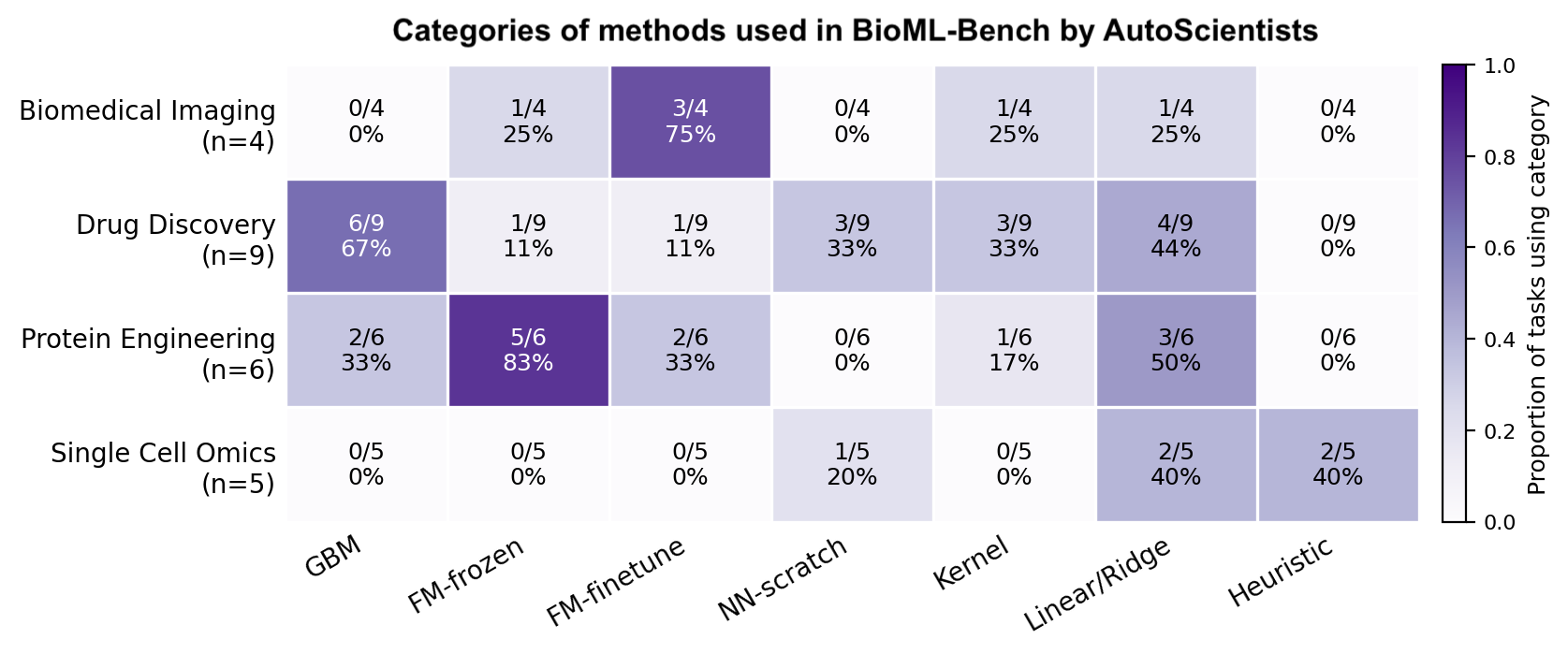}
\caption{Proportion of tasks within each BioML-Bench task type that uses each
method category. The cell value is the fraction of tasks in that row's family
in which at least one component of the \name approach falls in that
column's category and categories are non-exclusive (most tasks use several). GBM = gradient-boosted trees; FM-frozen / FM-finetune = pretrained foundation model used as frozen features / fine-tuned; NN-scratch = neural net trained from scratch; \revision{Kernel = kernel and instance methods (SVM, SVR, GP, $k$-NN);} Linear/Ridge = Ridge, RidgeCV, Lasso; Heuristic = no learned model.}
\label{fig:method-heatmap}
\end{figure}

\xhdr{Overall trends}
Across 24 tasks, Linear/Ridge models are the most frequently occurring
category (10/24), but almost always as meta-learners or simple downstream
probes on top of richer representations rather than standalone predictors.
Gradient-boosted trees on hand-crafted descriptors are the most common
\emph{primary} modeling strategy (8/24), concentrated in drug discovery
(6/9).
Pretrained foundation models are used either as frozen feature extractors
(7/24, predominantly in protein engineering) or fine-tuned end-to-end
(6/24, predominantly in biomedical imaging and one small molecule task).
Kernel methods and custom NNs from scratch each appear in roughly a fifth
of tasks (5/24 and 4/24, respectively) and nearly always as components
within ensembles.
Hand-crafted, training-free pipelines win on the two Single Cell Omics tasks
where supervised learning is infeasible due to absence of meaningful
labeled training signal.
Classical molecular fingerprints (Morgan/MACCS/AtomPair/Torsion/RDKit
descriptors) are present in every chemistry pipeline that does not
fine-tune a foundation model.
Fig.~\ref{fig:method-heatmap} shows the proportion of tasks within each
BioML-Bench task family that uses each of the seven method categories.

\begin{table*}[t]
\centering
\scriptsize
\caption{
Per task LLM token usage and estimated cost (\$ USD) on BioML-Bench. Token counts are reported in millions of tokens (MTok). Costs are estimated using Claude Sonnet 4.6 May 2026 pricing with cache writes charged at the 1-hour cache write price, and exclude GPU compute.
}
\label{tab:llm_usage_by_task}
\resizebox{\textwidth}{!}{
\begin{tabular}{lrrrrr|rrrrr}
\toprule
& \multicolumn{5}{c|}{Autoresearch} & \multicolumn{5}{c}{\name} \\
\cmidrule(lr){2-6} \cmidrule(lr){7-11}
Task
& Input & Output & CacheCreate & CacheRead & Cost (\$)
& Input & Output & CacheCreate & CacheRead & Cost (\$) \\
\midrule
{\detokenize{kaggle-histopathologic-cancer-detection}} & 0.001 & 0.4 & 0.8 & 100.0 & 40.1 & 0.005 & 0.8 & 19.6 & 232.4 & 199.6 \\
{\detokenize{kaggle-osic-pulmonary-fibrosis-progression}} & 0.006 & 4.8 & 5.9 & 535.7 & 267.9 & 0.018 & 5.7 & 32.8 & 738.3 & 503.9 \\
{\detokenize{kaggle-rsna-miccai-brain-tumor-radiogenomic-classification}} & 0.002 & 0.8 & 1.7 & 103.4 & 52.9 & 0.010 & 2.9 & 27.9 & 491.4 & 358.0 \\
{\detokenize{kaggle-uw-madison-gi-tract-image-segmentation}} & 0.002 & 1.1 & 1.7 & 146.8 & 70.3 & 0.008 & 0.9 & 10.7 & 156.1 & 124.5 \\
{\detokenize{open-problems-cell-cell-communication-ligand-target}} & 0.000 & 0.4 & 0.6 & 26.2 & 17.6 & 0.005 & 1.6 & 8.5 & 191.3 & 133.1 \\
{\detokenize{open-problems-label-projection}} & 0.000 & 0.4 & 0.4 & 30.9 & 16.9 & 0.002 & 0.6 & 4.7 & 79.2 & 60.5 \\
{\detokenize{open-problems-predict-modality}} & 0.000 & 0.1 & 0.1 & 10.2 & 5.5 & 0.003 & 0.7 & 5.9 & 165.5 & 95.0 \\
{\detokenize{open-problems-single-cell-perturbations}} & 0.000 & 0.3 & 0.4 & 23.1 & 13.9 & 0.007 & 2.1 & 13.2 & 320.0 & 206.3 \\
{\detokenize{open-problems-spatially-variable-genes}} & 0.003 & 1.8 & 3.5 & 219.8 & 114.0 & 0.008 & 2.6 & 13.4 & 317.2 & 214.2 \\
{\detokenize{polaris-adme-fang-hclint-1}} & 0.000 & 0.1 & 0.2 & 18.2 & 8.4 & 0.007 & 0.8 & 7.7 & 165.6 & 108.0 \\
{\detokenize{polaris-adme-fang-hppb-1}} & 0.000 & 0.2 & 0.3 & 36.7 & 16.4 & 0.007 & 1.7 & 11.4 & 419.5 & 220.4 \\
{\detokenize{polaris-adme-fang-solu-1}} & 0.000 & 0.2 & 0.2 & 28.2 & 12.3 & 0.003 & 0.7 & 5.4 & 96.9 & 71.7 \\
{\detokenize{polaris-pkis2-egfr-wt-c-1}} & 0.001 & 0.8 & 1.1 & 130.9 & 58.0 & 0.007 & 1.9 & 14.3 & 331.9 & 213.2 \\
{\detokenize{proteingym-dms-CBX4_HUMAN_Tsuboyama_2023_2K28}} & 0.001 & 0.4 & 0.5 & 44.2 & 22.8 & 0.006 & 1.5 & 9.5 & 286.1 & 166.2 \\
{\detokenize{proteingym-dms-CSN4_MOUSE_Tsuboyama_2023_1UFM_indels}} & 0.001 & 0.3 & 0.4 & 34.4 & 17.2 & 0.003 & 0.7 & 5.7 & 106.4 & 77.5 \\
{\detokenize{proteingym-dms-PSAE_PICP2_Tsuboyama_2023_1PSE}} & 0.000 & 0.1 & 0.2 & 9.7 & 6.0 & 0.006 & 1.4 & 12.1 & 286.1 & 179.4 \\
{\detokenize{proteingym-dms-Q8EG35_SHEON_Campbell_2022_indels}} & 0.001 & 1.6 & 1.7 & 121.8 & 70.0 & 0.005 & 1.3 & 10.2 & 206.2 & 143.1 \\
{\detokenize{proteingym-dms-SBI_STAAM_Tsuboyama_2023_2JVG}} & 0.000 & 0.3 & 0.3 & 19.4 & 13.0 & 0.003 & 0.8 & 6.7 & 149.3 & 96.4 \\
{\detokenize{proteingym-dms-SPIKE_SARS2_Starr_2020_binding}} & 0.000 & 0.6 & 0.6 & 25.8 & 20.6 & 0.005 & 1.4 & 8.9 & 174.0 & 126.1 \\
{\detokenize{tdcommons-bbb-martins}} & 0.000 & 0.1 & 0.2 & 21.1 & 9.3 & 0.009 & 1.6 & 12.2 & 356.2 & 203.7 \\
{\detokenize{tdcommons-caco2-wang}} & 0.000 & 0.2 & 0.2 & 17.7 & 8.9 & 0.006 & 1.3 & 7.6 & 398.8 & 184.2 \\
{\detokenize{tdcommons-cyp2d6-substrate-carbonmangels}} & 0.000 & 0.1 & 0.2 & 22.4 & 9.7 & 0.014 & 1.5 & 9.3 & 303.1 & 168.5 \\
{\detokenize{tdcommons-herg}} & 0.000 & 0.2 & 0.3 & 33.2 & 15.2 & 0.002 & 0.7 & 4.5 & 69.1 & 57.4 \\
{\detokenize{tdcommons-lipophilicity-astrazeneca}} & 0.001 & 0.4 & 0.6 & 47.4 & 23.2 & 0.003 & 0.5 & 3.9 & 143.7 & 73.9 \\
\midrule
Total & 0.024 & 15.7 & 21.9 & 1807.5 & 910.1 & 0.153 & 35.5 & 266.1 & 6184.5 & 3984.7 \\
\bottomrule
\end{tabular}
}
\end{table*}

\section{\name-Kermut for ProteinGym \revision{SPIKE-SARS2} binding DMS}
\label{app:proteingym}
Our \name-Kermut champion ProteinGym predictor extends the Kermut Gaussian process model
\citep{kermut} into a three-GP ensemble whose components share a
Kermut style structure kernel but differ in their sequence side feature sets
and kernel families.  All hyperparameters (kernel forms, priors,
ensemble composition, region-aware noise schedule, and 1000-step optimization
recipe) were chosen on the \revision{SPIKE-SARS2} binding task
during system development and then frozen. The same configuration is applied
unchanged to every other ProteinGym single substitution protein. \revision{This champion comes from the first of three independent runs from the Kermut seed, and all three runs and a single-agent Autoresearch control are reported in Appendix~\ref{app:repeated_runs}.}

\subsection{Setup}
\xhdr{Inputs and per fold preprocessing}
For each variant, \name-Kermut reads (i) a 1280-dimensional ESM-2 \texttt{t33\_650M\_UR50D}
mean pool embedding, (ii) the wild-type marginal masked log-probabilities at
the mutated position from a ProteinMPNN conditional probability map, used both
to form $(\log p_{\mathrm{mut}}, \log p_{\mathrm{wt}}, \log p_{\mathrm{mut}} -
\log p_{\mathrm{wt}})$ scalars and as the basis of the structure kernel,
(iii) the ESM-2 zero-shot fitness score, and (iv) fifteen additional
zero-shot predictors covering inverse folding, structure-aware, MSA-based and
homology-based methods (\textsc{MIF}~\cite{yang2023masked}, VenusREM~\cite{tan2025venusrem}, \textsc{ProSST}-128/2048/4096~\cite{li2024prosst},
\textsc{RSALOR}~\cite{tsishyn2025residue}, \textsc{ESCOTT}~\cite{tekpinar2025prescott}, xTrimoPGLM-1B-CLM~\cite{chen2025xtrimopglm}, SaProt-650M-AF2~\cite{su2024saprot},
ESM-IF1~\cite{hsu2022learning}, Unirep evotuned~\cite{alley2019unified}, \textsc{S3F}-MSA~\cite{zhang2024multi}, MSA-Transformer ensemble~\cite{rao2021msa},
VespaG~\cite{vespag}, SiteRM~\cite{prillo2024ultrafast}).  The four lightweight C$\alpha$ contact map features appended are
the number of residues within 8\,\AA, the mean residue distance, the local
density within 10\,\AA, and an inverse contact RSA proxy.  Within each
cross-validation fold, contact features are $z$-scored using only training
statistics, missing extra zero-shot values are filled with the per fold
training set median, and the resulting 23-dimensional scalar block is
standardized with training fold mean and standard deviation.  GP regression
targets are quantile normalized to a standard normal on the training fold
via van~der~Waerden scores, where rank $r_i$ (1-indexed, average tied) is mapped
to $\Phi^{-1}\!\bigl((r_i-0.5)/N\bigr)$, keeping quantiles inside
$(0,1)$ and avoiding $\pm\infty$, and test time MSE is computed against
$z$-scored targets to match the ProteinGym/Kermut benchmark convention,
with predictions clipped to $[-10,10]$ to guard against GP extrapolation
blow-up on out-of-distribution folds.

\xhdr{Structure kernel}
\name-Kermut adopts the Kermut structure kernel.  For each variant indexed by (mutated residue position, mutant
amino acid) \name-Kermut precomputes three pairwise residue matrices on the training
fold, (i) a Hellinger distance between the full 20-amino-acid ProteinMPNN
conditional probability distributions at the mutated positions,
(ii) an L1 distance in log space, $|\log p(\mathrm{aa}_i\mid \mathrm{ctx}_i) -
\log p(\mathrm{aa}_j\mid \mathrm{ctx}_j)|$, i.e.\ the absolute difference of the
scalar log-probabilities at the respective mutant amino acids, and
(iii) the Euclidean distance between their C$\alpha$ coordinates.  The structure kernel is the radial product
$k_{\mathrm{struct}}(i,j) = \exp(-\ell_h h_{ij} - \ell_p p_{ij} - \ell_d d_{ij})$
with positive constrained per component lengthscales, wrapped in a
\texttt{ScaleKernel} whose output scale carries a $\mathrm{LogNormal}(0,0.5)$
prior.

\xhdr{Three-GP ensemble}
Each ensemble member combines the structure kernel with a sequence kernel via
a learnable mixture
$k = \pi\,k_{\mathrm{struct}} + (1-\pi)\,k_{\mathrm{seq}}$, where
$\pi=\sigma(\mathrm{raw}\_\pi)$ is reparameterized through a sigmoid.
The mean function is a single input \texttt{LinearMean} over the ESM-2
zero-shot score, providing a strong unsupervised prior on top of which the
GP models residuals.  The three members differ only in $k_{\mathrm{seq}}$
and in the sequence side feature set.
\textbf{(GP1)} a \textsc{Mat\'ern}-3/2 ARD kernel on the
$1280+23 = 1303$-dimensional concatenation of ESM-2 embedding and
standardized scalars,
\textbf{(GP2)} a \textsc{Mat\'ern}-5/2 ARD kernel on the 1280-d ESM-2
embedding alone, providing a smoother sequence kernel that does not see the
auxiliary scalars, and
\textbf{(GP3)} a linear kernel on a compact 18-d feature vector formed by
dropping the five trailing extra zero-shot predictors (ProSST-4096,
\textsc{S3F}-MSA~\cite{zhang2024multi}, MSA-Transformer ensemble~\cite{rao2021msa}, VespaG, and SiteRM) from the
scalar block. The three log-probability scalars, the ESM-2 zero-shot score,
the first ten extra zero-shot predictors, and all four contact map features
are retained ($23 - 5 = 18$).  Predictions
from members that converge are averaged, and if a member fails for numerical
reasons (e.g.\ Cholesky errors on a degenerate fold) it is silently dropped
and the remaining members are averaged.

\xhdr{Likelihood and region-aware noise}
\name-Kermut uses a \texttt{FixedNoiseGaussianLikelihood} with learnable additional
homoscedastic noise carrying a $\mathrm{HalfCauchy}(0.3)$ prior. The fixed
per sample noise variances are derived from a region-aware schedule that
down-weights training points whose CV fold index is far from the held-out
fold. With $b_i = |\mathrm{fold}_i - \mathrm{test\_fold}|$, a sample weight
$w_i = 1 + 0.5\,e^{-b_i}$ is converted to fixed noise
$\sigma^{2}_i = \mathrm{clip}(\,0.05\,(w_i^{-1} - \min_j w_j^{-1}),\,10^{-4},\infty)$.
This places near zero fixed noise on the most reliable points (those
adjacent to the test fold) and inflates noise on points distant from the
test region, encouraging the GP to fit local structure rather than
distant region trends.

\xhdr{Optimization}
Each GP's parameters (sequence kernel ARD lengthscales, structure kernel
component lengthscales, structure outputscale, $\pi$, mean parameters, and
additional likelihood noise) are jointly trained for 1000 steps of AdamW with
learning rate $10^{-1}$ decayed cosine style to $10^{-3}$, maximizing the
exact GP marginal likelihood.  The same number of steps and the same
optimizer are used for every protein and every CV fold, with random seeds
fixed to $2024$.

\xhdr{Cross-validation and reporting}
\name-Kermut trains and evaluates on the 5-fold CV columns provided by the ProteinGym
single substitutions benchmark (\texttt{fold\_random\_5},
\texttt{fold\_modulo\_5}, or \texttt{fold\_contiguous\_5}), refitting all
per fold preprocessing, per fold ensemble members, and per fold target
normalization independently for each held-out fold.

\subsection{Ablations of \name-Kermut} \label{app:kermut-ablations}

We ablate three key design choices that \name-Kermut introduces to the original Kermut. All runs use the same five-fold cross-validation on the \revision{SPIKE-SARS2} binding task dataset. The ablations are as follows. (1) \textbf{No ensemble} keeps only GP1 (full ESM-2 + scalar features, Mat\'ern-3/2 + ARD). GP2 (ESM-2 only, Mat\'ern-5/2) and GP3 (compact 18-d, linear) are removed. (2) \textbf{No quantile norm targets} trains on simple per fold $z$-scored targets instead of mapping training targets to a standard normal via rank quantiles. (3) \textbf{No extra zero-shot scores} drops the 15 auxiliary
zero-shot predictors (MIF, VenusREM, ProSST-\{128,2048,4096\}, RSALOR,
ESCOTT, xTrimoPGLM-1B-CLM, SaProt\_650M\_AF2, ESM-IF1, Unirep\_evotune,
S3F\_MSA, MSA\_Transformer\_ensemble, VespaG, SiteRM), and only ESM-2 650M
zero-shot remains in the scalar block.

\begin{table}[h]
\centering
\caption{Ablations on \name-Kermut on ProteinGym \revision{performance} on \texttt{SPIKE\_SARS2\_Starr\_2020\_binding}. Per split Spearman\,$\uparrow$ and MSE\,$\downarrow$ are reported. The best value is shown in \textbf{bold} and the second best is shown in \textit{italics}.}
\label{tab:ablation-spike}
\footnotesize
\setlength{\tabcolsep}{4pt}
\renewcommand{\arraystretch}{1.05}
\begin{tabular}{l cc cc cc cc}
\toprule
& \multicolumn{2}{c}{\textbf{Contiguous}}
& \multicolumn{2}{c}{\textbf{Modulo}}
& \multicolumn{2}{c}{\textbf{Random}}
& \multicolumn{2}{c}{\textbf{Average}}\\
\cmidrule(lr){2-3}\cmidrule(lr){4-5}\cmidrule(lr){6-7}\cmidrule(lr){8-9}
\textbf{Variant}
  & $\rho$ & MSE & $\rho$ & MSE & $\rho$ & MSE & $\rho$ & MSE\\
\midrule
Kermut
  & 0.6950          & 0.5830
  & 0.7060          & 0.5220
  & 0.8410          & \textit{0.2070}
  & 0.7473          & \textit{0.4373}\\
\textbf{\name}
  & \textbf{0.7842} & \textit{0.5229}
  & \textbf{0.8174} & \textit{0.5004}
  & \textit{0.9204} & 0.3347
  & \textbf{0.8407} & 0.4527\\
\midrule
\multicolumn{9}{l}{\textbf{Ablations}}\\
No ensemble
  & 0.7455          & 0.6092
  & \textit{0.8030} & 0.5460
  & \textbf{0.9207} & 0.3527
  & \textit{0.8231} & 0.5026\\
No quantile norm
  & \textit{0.7638} & \textbf{0.4595}
  & 0.7782          & \textbf{0.4263}
  & 0.8791          & \textbf{0.1645}
  & 0.8070          & \textbf{0.3501}\\
No extra ZS
  & 0.7591          & 0.5802
  & 0.7821          & 0.5718
  & 0.9109          & 0.3590
  & 0.8174          & 0.5037\\
\bottomrule
\end{tabular}
\end{table}

The ablations in Table~\ref{tab:ablation-spike} show \revision{that these three components each contribute} to \name-Kermut's ranking performance. \revision{Averaged over the three splits, removing quantile normalized targets causes the largest drop in Spearman correlation (0.8407 to 0.8070, $-0.0337$), removing the extra zero-shot scores the second largest (to 0.8174, $-0.0233$), and removing the ensemble the smallest (to 0.8231, $-0.0176$). The ensemble ablation causes the largest drop only on the contiguous split, because the three GPs carry complementary inductive biases and their diversity is most valuable on this harder split, where sequence extrapolation matters most.} Conversely, removing \revision{quantile normalization} improves MSE, since z-scored targets compress prediction scale. The \revision{Spearman drop indicates} that mapping skewed DMS scores to a standard normal before fitting the GP \revision{improves the learned ranking.} The 15 auxiliary zero-shot predictors contribute a consistent \revision{improvement} in both $\rho$ and MSE across all splits, with no single split driving the effect, suggesting they supply diverse evolutionary signal that the ESM-2 embedding alone does not capture rather than overfitting to a particular data regime.

\end{document}